\documentclass[journal]{IEEEtran}
\IEEEoverridecommandlockouts
\usepackage{cite}
\usepackage{amsmath,amssymb,amsfonts}
\usepackage{algorithmic}
\usepackage{graphicx}
\usepackage{textcomp}
\usepackage{xcolor}
\usepackage{tikz}
\usepackage{makecell}
\usetikzlibrary{positioning, shapes.geometric}
\usepackage{utfsym}
\usepackage{stfloats}
\usepackage{soul,color, framed}
\soulregister{\cite}7 
\soulregister{\citep}7 
\soulregister{\citet}7 
\soulregister{\ref}7 
\soulregister{\pageref}7
\usepackage{colortbl}
\definecolor{mygray}{gray}{.9}
\usepackage{makecell}
\usepackage{tablefootnote}
\usepackage{url}

\usepackage{algorithm}
\usepackage{array}
\usepackage{multirow}
\usepackage{fancyhdr}
\usepackage{ragged2e}
\usepackage{caption}
\usepackage{hyperref}
\usepackage{setspace}
\usepackage{tikz}
\usepackage{bm}
\usepackage{booktabs}
\usepackage{siunitx}
\usepackage{pdflscape}
\usepackage{subcaption}
\usepackage{lipsum}
\newcommand{\vect}[1]{\bm{#1}}
\newcommand{\methodname}{{\tt{FedFM}}}

\usepackage{xcolor}
\usepackage{xpatch}

\makeatletter
\ExplSyntaxOn
\cs_new:Npn \bibColoredItems #1#2
  {
    \clist_map_inline:nn {#2} { \cs_new:cpn {bib@colored@##1} {#1} } 
  }
\ExplSyntaxOff

\newcommand\bib@setcolor[1]{%
  \ifcsname bib@colored@#1\endcsname
    \expanded{\noexpand\color{\csname bib@colored@#1\endcsname}}%
  \else
    \normalcolor
  \fi
}

\IfPackageLoadedTF{hyperref}{\@tempswatrue}{\@tempswafalse}
\if@tempswa
  \xpatchcmd\@bibitem {\H@item}{\bib@setcolor{#1}\H@item}{}{\PatchFailed}
  \xpatchcmd\@lbibitem{\H@item}{\bib@setcolor{#2}\H@item}{}{\PatchFailed}
\else
  \xpatchcmd\@bibitem {\item}  {\bib@setcolor{#1}\item}  {}{\PatchFailed}
  \xpatchcmd\@lbibitem{\item}  {\bib@setcolor{#2}\item}  {}{\PatchFailed}
\fi
\makeatother

\definecolor{revisioncolor}{HTML}{E30022} 

\def\BibTeX{{\rm B\kern-.05em{\sc i\kern-.025em b}\kern-.08em
    T\kern-.1667em\lower.7ex\hbox{E}\kern-.125emX}}

\begin{document}

\title{Advances and Open Challenges in Federated Foundation Models}
\author{Chao Ren,~\IEEEmembership{Member,~IEEE,} 
      Han Yu,~\IEEEmembership{Senior Member,~IEEE,}
      Hongyi Peng, 
      Xiaoli Tang, 
      Bo Zhao, 
      Liping Yi, 
      
      Alysa Ziying Tan,~\IEEEmembership{Member,~IEEE,} 
      Yulan Gao,~\IEEEmembership{Member,~IEEE,}
      Anran Li,~\IEEEmembership{Member,~IEEE,} 

      Xiaoxiao Li,~\IEEEmembership{Member,~IEEE,} 
      Zengxiang Li, 
      and Qiang Yang,~\IEEEmembership{Fellow,~IEEE}
  
\thanks{The work is supported, in part, by the Internal talent award (TRACS) with Wallenberg-NTU Presidential Postdoctoral Fellowship, Wallenberg AI, Autonomous Systems and Software Program (WASP) and Nanyang Technological University, Sweden and Singapore; the National Research Foundation Singapore and DSO National Laboratories under the AI Singapore Programme (No: AISG2-RP-2020-019); and the RIE 2020 Advanced Manufacturing and Engineering (AME) Programmatic Fund (No. A20G8b0102), Singapore.}
\thanks{Chao Ren is currently a Wallenberg-NTU Presidential Postdoctoral Fellow with the School of Electrical Engineering and Computer Science, KTH Royal Institute of Technology, Sweden; and the College of Computing and Data Science, Nanyang Technological University, Singapore (e-mail: renc0003@e.ntu.edu.sg).}
\thanks{Han Yu is an Associate Professor at the College of Computing and Data Science, Nanyang Technological University, Singapore (\textit{Corresponding Author}, e-mail: han.yu@ntu.edu.sg).} %
\thanks{Hongyi Peng, Xiaoli Tang, Bo Zhao, and Alysa Ziying Tan are PhD Candidates at the College of Computing and Data Science, Nanyang Technological University, Singapore (e-mail: \{hongyi001, xiaoli001, bo008, s190109\}@ntu.edu.sg).}
\thanks{Liping Yi is a PhD Candidate at the College of Computer at Nankai University, China (e-mail: yiliping@nbjl.nankai.edu.cn).}
\thanks{Yulan Gao is a Postdoctoral Fellow at the School of Electrical Engineering and Computer Science, KTH Royal Institute of Technology, Sweden (e-mail: yulang@kth.se).}  
\thanks{Anran Li is a Postdoctoral Associate at the School of Medicine, Yale University, USA (e-mail: anran.li@yale.edu).}  
\thanks{Zengxiang Li is the Executive Vice President of the Digital Research Institute of ENN Group, Langfang, China (e-mail: lizengxiang@enn.cn).}%
\thanks{Xiaoxiao Li is an Assistant Professor at the Department of Electrical and Computer Engineering, The University of British Columbia, Vancouver, BC, Canada (e-mail: xiaoxiao.li@ece.ubc.ca).}%
\thanks{Qiang Yang is Professor Emeritus at the Department of Computer Science and Engineering,
Hong Kong University of Science and Technology, Hong Kong, and the Chief AI Officer of WeBank, Shenzhen, China (e-mail: qyang@cse.ust.hk).}%

} %

\markboth{Journal of \LaTeX\ Class Files,~Vol.~X, No.~X, XXX,~2024}%
{Shell \MakeLowercase{\textit{et al.}}: A Sample Article Using IEEEtran.cls for IEEE Journals}

\maketitle

\begin{abstract}
The integration of Foundation Models (FMs) with Federated Learning (FL) presents a transformative paradigm in Artificial Intelligence (AI). This integration offers enhanced capabilities, while addressing concerns of privacy, data decentralization and computational efficiency.
This paper provides a comprehensive survey of the emerging field of \underline{Fed}erated \underline{F}oundation \underline{M}odels (\methodname{}), elucidating their synergistic relationship and exploring novel methodologies, challenges, and future directions that the FL research field needs to focus on in order to thrive in the age of FMs. 
A systematic multi-tiered taxonomy is proposed, categorizing existing \methodname{} approaches for model training, aggregation, trustworthiness, and incentivization. 
Key challenges, including how to enable FL to deal with high complexity of computational demands, privacy considerations, contribution evaluation, and communication efficiency, are thoroughly discussed. 
Moreover, this paper explores the intricate challenges of communication, scalability and security inherent in training/fine-tuning FMs via FL. It highlights the potential of quantum computing to revolutionize the processes of training, inference, optimization and security.
This survey also introduces the implementation requirement of \methodname{} and some practical \methodname{} applications. It highlights lessons learned with a clear understanding of our findings for \methodname{}.
Finally, this survey not only provides insights into the current state and challenges of \methodname{}, but also offers a blueprint for future research directions, emphasizing the need for developing trustworthy solutions.
It serves as a foundational guide for researchers and practitioners interested in contributing to this interdisciplinary and rapidly advancing field.

\end{abstract}


\section{Introduction}
\subsection{Motivation}
\label{sec:Motivation}

The Artificial Intelligence (AI) landscape is undergoing significant advancement propelled by the emergence of Foundation Models (FMs) \cite{yuan2023power,bommasani2021opportunities}. FMs are large-scale machine learning (ML) models that serve as a foundational backbone for a wide range of learning tasks across diverse modalities\cite{zhou2023comprehensive}.
They are typically pre-trained on extensive datasets using self-supervised learning methods, which equip them with general and adaptable knowledge for various tasks. While technically representing an evolutionary rather than revolutionary development in AI \cite{bommasani2021opportunities}, FMs have significantly reshaped model training and deployment practices.
This practice arises from the high costs associated with FM training and the relatively limited scale of available datasets, making the traditional training-from-scratch paradigm impractical.

Recently, FMs with their diverse designs and learning processes have achieved remarkable success in tackling complex tasks that were once considered too difficult or even impossible. Particularly in Natural Language Processing (NLP), FMs such as those in the GPT series \cite{radford2019language, brown2020language, OpenAI_GPT4_2023}, LLaMa \cite{llama} and PaLM \cite{chowdhery2022palm} have demonstrated extraordinary capabilities. In computer vision, Segment Anything \cite{segmentanyting} has emerged as a standout performer. In the realm of Generative AI models, Stable Diffusion \cite{stablediffusion} has garnered significant attention. Notably, DALL-E \cite{dalle} and CLIP \cite{clip} have demonstrated outstanding performance in multi-modal tasks, bridging the understanding of textual and visual data. 

However, the success of FMs comes at a high cost. The enhanced capabilities require a substantial amount of high-quality training data and computational resources. In addition, centralized training of FMs raises concerns about data privacy and security. These challenges underscore the potential value of a collaborative FM training paradigm that enables the distribution of both data resources and computational loads, while preserving data privacy.

Federated Learning (FL) \cite{kairouz2021advances} presents a promising solution to these challenges. It is a collaborative learning approach that enhances privacy, leverages distributed datasets, and distributes computational loads. Training and fine-tuning FMs via FL has several advantages.
FL can broaden the data horizons for FMs by aggregating diverse data sources from both the public and private sectors \cite{chen2023federated}, including research institutions and industries. It can also distribute computational loads among multiple participants, leading to more efficient resource utilization and mitigating the risk of AI monopolies by major technology  companies. On the other hand, 
the robustness of FMs strengthens the effectiveness of FL, particularly in managing non-iid (non-independent and identically distributed) data. Moreover, the flexibility of FMs in adapting to various downstream tasks facilitates easier personalization of FL models. These benefits highlight the promising potential of integrating FMs with FL.

\captionsetup[table]{labelformat=simple, labelsep=newline, textfont=sc, justification=centering}

\begin{table*}[htbp]
\caption{Comparison between \methodname{} and Traditional FMs}
\label{tab:comparison}
\centering
\resizebox{\textwidth}{!}{
\begin{tabular}{l|cc|cc}
\toprule
{\textsc{Factors}} & \multicolumn{2}{c|}{\textsc{\methodname{}}} & \multicolumn{2}{c}{\textsc{Traditional FMs}} \\
\midrule
Data Efficiency & More Diverse Data Across Devices  & \usym{1F5F8}   & Higher Data Demand for Same Performance & \usym{2717}  \\ 
Data Privacy & Privacy Preserving Mechanisms  & \usym{1F5F8} & Centralized Data Collection & \usym{2717}  \\ 
Model Performance & Diverse and Adaptive Improvement & \usym{1F5F8} & Lack Diversity and Personalization & \usym{2717} \\ 
System Operation & Distributed Coordination & \usym{1F5F8} & Central Management & \usym{2717} \\ 
Scalability & Scalable to Multiple Participants & \usym{1F5F8} & Unscalable with much Larger Datasets  &  \usym{2717} \\ \hline
Deployment  & Challenging & \usym{2717} & Easier to Setup & \usym{1F5F8}\\
Consistency & Not Particularly Controllable & \usym{2717} &  Controllable for Consistent Updating & \usym{1F5F8}\\
Latency & Distributed Communication and Computation & \usym{2717} & Lower under Centralized Environment & \usym{1F5F8} \\
\bottomrule 
\end{tabular}
}
\vspace{4pt}
\end{table*}

\begin{table*}[htbp]
\setlength\tabcolsep{0.5pt}
\caption{Comparison of Existing Reviews and Position Papers with This Survey Paper}
\label{tab:existing review}
\centering
\resizebox{\textwidth}{!}{
\begin{tabular}{c|c|cccccccc}
\toprule
{\textsc{Years}} & {\textsc{References}} & 
{\makecell[c]{\textsc{\methodname{}} \\ \textsc{Taxonomy}} } & 
{\makecell[c]{\textsc{\methodname{}} \\ \textsc{Efficiency}}  } & 
{\makecell[c]{\textsc{\methodname{}} \\ \textsc{Trustworthiness}}  } &
{\makecell[c]{\textsc{\methodname{}} \\ \textsc{Incentivization}} } & 
{\makecell[c]{\textsc{\methodname{}} \\ \textsc{Evaluation}} } &
{\makecell[c]{\textsc{\methodname{}} \\ \textsc{Framework}} } & {\makecell[c]{\textsc{\methodname{}} \\ \textsc{Challenges}} } &
{\makecell[c]{\textsc{\methodname{}} \\ \textsc{Future} \\ \textsc{Directions}}  }\\

\midrule
\multicolumn{1}{c|}{\multirow{4}{*}{\begin{tabular}[c]{@{}l@{}}2023\end{tabular}}} &
Chen \textit{et al}. \cite{chen2023federated} & \usym{2717}   & \usym{1F5F8} & \usym{2717} & \usym{2717}  & \usym{2717} & \usym{2717} & \usym{1F5F8} & \usym{2717} \\ 
\multicolumn{1}{c|}{} &
Zhuang \textit{et al}. \cite{zhuang2023foundation} & \usym{2717}   & \usym{1F5F8} & \usym{1F5F8} & \usym{1F5F8}  & \usym{2717} & \usym{2717} & \usym{1F5F8} & \usym{1F5F8} \\ 
\multicolumn{1}{c|}{} &
Yu \textit{et al}. \cite{yu2023federated} & \usym{2717}   & \usym{1F5F8} & \usym{1F5F8} & \usym{2717}  & \usym{2717} & \usym{2717} & \usym{1F5F8} & \usym{2717} \\ 
\multicolumn{1}{c|}{} &
Kang \textit{et al}. \cite{kang2023grounding} & \usym{1F5F8}   & \usym{1F5F8} & \usym{1F5F8} & \usym{2717}  & \usym{2717} & \usym{2717} & \usym{1F5F8} & \usym{1F5F8} \\ \midrule

\multicolumn{1}{c|}{\multirow{3}{*}{\begin{tabular}[c]{@{}l@{}}2024\end{tabular}}} &
Herbert \textit{et al}. \cite{woisetschlager2024survey} &  \usym{1F5F8}  & \usym{1F5F8}   & \usym{2717}  & \usym{2717}  & \usym{2717}  & \usym{2717}   & \usym{2717}  &  \usym{1F5F8} \\ 
\multicolumn{1}{c|}{} &
Li \textit{et al}. \cite{li2024position} & \usym{1F5F8}   & \usym{2717} & \usym{1F5F8} & \usym{2717}  & \usym{2717} & \usym{2717} & \usym{1F5F8} & \usym{2717}  \\ 
\multicolumn{1}{c|}{} &
Li \textit{et al}. \cite{li2024synergizing} & \usym{1F5F8}   & \usym{1F5F8} & \usym{1F5F8} & \usym{2717}  & \usym{2717} & \usym{1F5F8} & \usym{1F5F8} & \usym{1F5F8}  \\ \midrule

\multicolumn{2}{c|}{\textbf{This survey paper}} & 
 \usym{1F5F8}   & \usym{1F5F8} & \usym{1F5F8} & \usym{1F5F8}  & \usym{1F5F8} & \usym{1F5F8} & \usym{1F5F8} & \usym{1F5F8} \\ 

\bottomrule 
\end{tabular}
}
\end{table*}

\subsection{Related Works and Contributions}
\label{sec:Contribution}

Recent efforts to integrate FMs into the FL training process have introduced a range of novel techniques and designs. The latest work in \cite{sani2024future} show that generative pre-trained large language models (LLMs) training process is highly relevant to the classical challenges of federated statistical and hardware heterogeneity. Also, convergence is robust to partial participation, opening the avenue for computational-efficient collaborative training. We refer to these emerging approaches as \underline{Fed}erated \underline{F}oundation \underline{M}odels (\methodname{}). 
The motivation for \methodname{} is to address the limitations of traditional FMs and improve upon them by leveraging the potential of FL, offering several advantages over traditional FMs. The key factors driving the development of \methodname{} are summarized in Table \ref{tab:comparison} (with \usym{1F5F8} indicating advantages, and \usym{2717} indicating limitations).
Table \ref{tab:comparison} contrasts \methodname{} and traditional FMs across a diverse spectrum of topics ranging from data efficiency, data privacy, model performance, system operation, scalability, deployment, consistency and latency, to the broader array of trustworthy AI. 

As the research field of \methodname{} is still emerging, there are currently only two position papers \cite{chen2023federated,li2024position} and five short review papers \cite{zhuang2023foundation,yu2023federated,kang2023grounding,woisetschlager2024survey,li2024synergizing} on this topic. Table \ref{tab:existing review} compares these existing position papers and review paper with our survey paper in terms of the coverage of important aspects of \methodname{} (with \usym{1F5F8} indicating mentioned, and \usym{2717} indicating no discussion). While the aforementioned survey efforts provide a reasonable overview of some aspects of \methodname{}, they tend to be brief and often lack comprehensiveness. Moreover, they are limited to considering selected aspects of \methodname{}, such as training/aggregation efficiency and trustworthiness, which does not provide a comprehensive overview of this interdisciplinary field for readers. Additionally, they neglect key issues of \methodname{} contribution evaluation and frameworks. To the best of our knowledge, this survey provides the most comprehensive coverage of the \methodname{} topic.

The primary contribution of this survey paper is the proposal of a systematic multi-tiered taxonomy for \methodname{}. It categorizes existing approaches and identifies key challenges and opportunities in this domain, encompassing innovative strategies for \methodname{} training and aggregation, strategies for achieving trustworthiness, and designs for \methodname{} incentive mechanism prioritizing efficiency, privacy, novel contribution evaluation, and active selection. Additionally, we explore the potential of quantum computing to enhance the effectiveness, efficiency and security of \methodname{}. Our survey not only reviews existing works, but also comprehensively discusses the motivations behind these approaches, often inspired by developments and insights regarding FMs. Finally, we outline promising future research directions, aiming to inspire advancements in the creation of efficient, secure and scalable \methodname{} approaches. 

\begin{figure*}[t]
    \centering
    \includegraphics[width=1.0\linewidth]{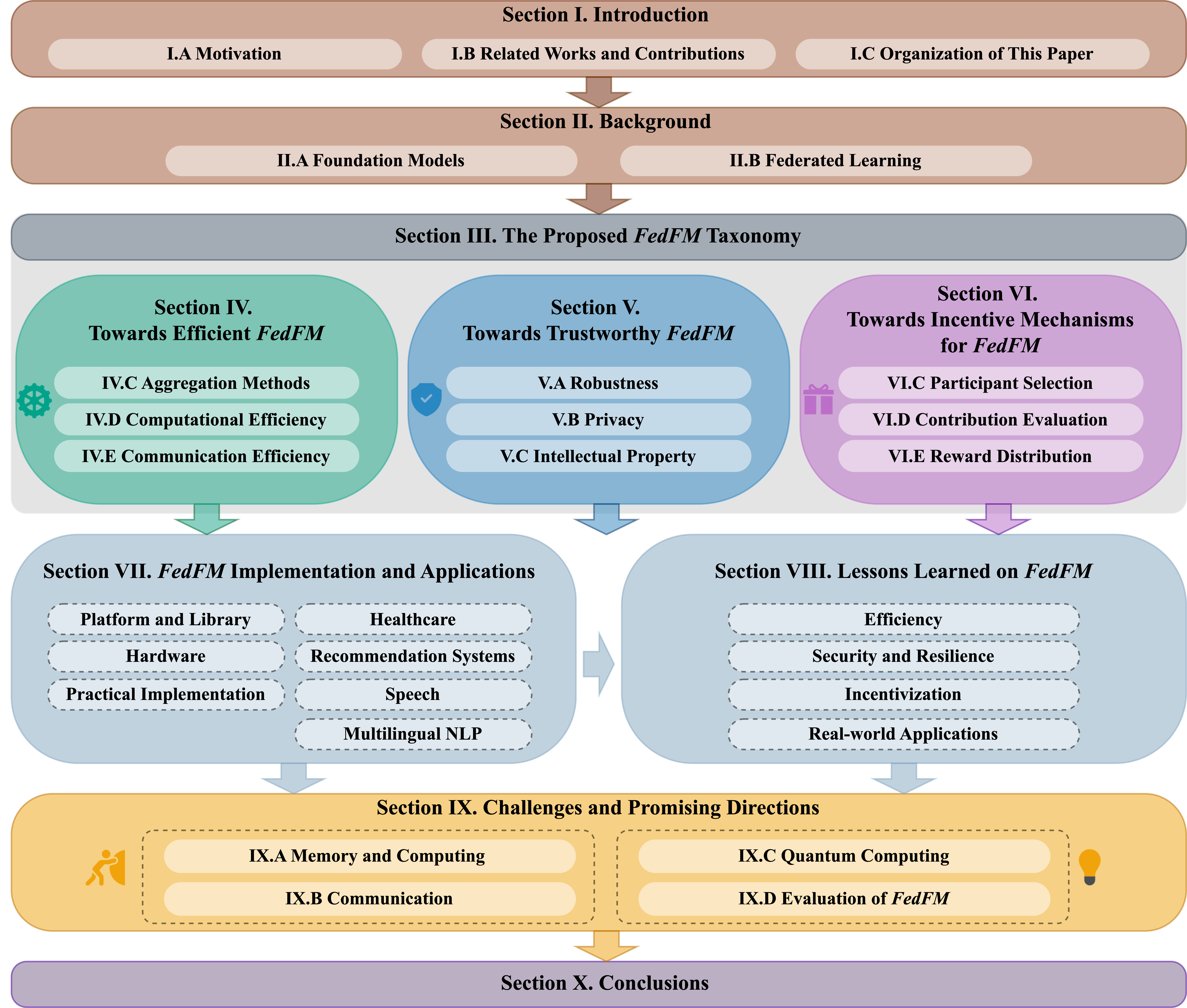}
    \caption{Organization of the \methodname{} survey.}
    \label{fig:organization}
    \vspace{-1pt}
\end{figure*}

\subsection{Organization of This Survey Paper}
\label{sec:Organization}
The survey paper is organized as follows. Section \ref{sec:terminology} introduces FMs and FL, concepts relevant to our discussion. Section \ref{sec:taxonomy} describes the \methodname{} taxonomy, which encompasses several key aspects: training and aggregation methods, ensuring trustworthiness, creating effective incentive mechanisms, and managing evaluation. 
Section \ref{sec: training and aggregation} focuses on aggregation, communication and computational efficiency, highlighting the nuanced considerations required for the effective deployment of \methodname{}.
Section \ref{sec:trustworthy} addresses the critical areas of robustness, privacy and intellectual property for \methodname{}.
Section \ref{sec:incentive} highlights the complexity of developing effective incentive mechanisms for \methodname{}. 

Then, Section \ref{sec:implementation} describes the practical implementation and applications of \methodname{} in terms of platforms and libraries and hardware consideration. Section \ref{sec:lesson} summarizes the key insights of three categories of \methodname{} and connects these lessons to Section \ref{sec:future}. Section \ref{sec:future} discusses open research challenges of \methodname{} in terms of memory, computing and communication, while also delving into potential future directions involving quantum computing techniques for \methodname{} and effective \methodname{} evaluation. Finally, Section \ref{sec:conclusion} concludes this \methodname{} survey paper. 

\section{Background}
\label{sec:terminology}
This section aims to provide an overview of FMs and FL, outlining the fundamental concepts and historical development trajectories that have propelled their current prominence in the field. There are more recent surveys on FMs and FL fields, respectively.

{Recent surveys on FMs have highlighted the advancements and applications of these models in various AI tasks. Bommasani et al. \cite{bommasani2021opportunities} provide a detailed analysis of the opportunities and risks associated with FMs, discussing their impact on different AI fields. Another survey by Myers et al. \cite{myers2024foundation} explores the design, training techniques, and deployment challenges of FMs, offering insights into their scalability and performance. Furthermore, Brown et al. \cite{brown2020language} review the progress and future trends in FMs, with a particular focus on language models like GPT-3/GPT-4 and their implications for AI research.

Recent surveys have extensively discussed relevant schemes for efficiency, privacy, and other challenges in FL. For instance, Li et al. \cite{li2020federated} provide a comprehensive review of FL, covering various optimization techniques, privacy-preserving methods, and system challenges. Kairouz et al. \cite{kairouz2021advances} delve into the advancements in FL, emphasizing theoretical foundations, communication efficiency, and security measures. In addition, Yang et al. \cite{yang2019federated} discuss the applications of FL across different domains, highlighting the practical considerations and future directions for FL research.}

\subsection{Foundation Models}
\label{sec:FMs}
The term FMs was coined in \cite{bommasani2021opportunities} to denote the role of such models as a foundational base, from which numerous task-specific models can be built through adaptation. FMs emphasize the stability of their architecture during training and the consistency of parameters throughout the adaptation phase. This term also signifies a profound shift in AI research and deployment, which began in the field of NLP, where the synergy of large-scale models and transfer learning \cite{TransferLearning2010} resulted in the development of powerful models.

The need for FMs arises from several critical challenges and driving forces in the field of AI:
\begin{itemize}
    \item \textbf{High Costs of Training}: Training large-scale models from scratch is resource-intensive, requiring substantial computational power and vast amounts of data. FMs address this by being pre-trained on large datasets, enabling them to learn generalized knowledge that can be fine-tuned for specific tasks, reducing the need for extensive resources for each new task. For example, GPT-2 \cite{radford2019language} consists of 1.3 billion parameters with a memory footprint of approximately 2.6 GB, while GPT-3 \cite{brown2020language} consists of 175 billion parameters with a memory footprint of around 350 GB in half-precision format. GPT-4 \cite{OpenAI_GPT4_2023} has reached the scale of over one trillion parameters.
    \item \textbf{Data Scarcity}: Many AI tasks suffer from a lack of sufficient high-quality labeled data. FMs, pre-trained on diverse and extensive datasets, can be adapted to new tasks with limited data, leveraging their pre-existing knowledge to achieve high performance even with smaller datasets. GPT-3, for instance, is trained on a dataset with 300 billion tokens \cite{brown2020language}, equivalent to hundreds of gigabytes of data. Research suggests a potential depletion of high-quality language data by 2026 \cite{villalobos2022will}, raising concerns that data availability might eventually impede the advancement of FMs.
    \item \textbf{Scalability}: Traditional models often struggle with scalability when dealing with diverse tasks and large datasets. FMs, with their scalable architectures, can handle a wide range of tasks across different domains, making them versatile and efficient for various applications \cite{ibrahim2024simple}.
    \item \textbf{Transfer knowledge Ability}: The ability to transfer knowledge from one task to another is a significant advantage of FMs. By pre-training on broad data and fine-tuning for specific tasks, FMs facilitate transfer learning, allowing models to be quickly adapted to new tasks without starting from scratch \cite{subramanian2024towards}.
    \item \textbf{Efficiency and Performance}: FMs have demonstrated remarkable success in improving efficiency and performance across numerous AI applications. In NLP, for instance, models like GPT series \cite{radford2019language, brown2020language, OpenAI_GPT4_2023}, LLaMa\cite{llama}, and PaLM have set new benchmarks for language understanding and generation. In computer vision, models like Segment Anything and generative models like Stable Diffusion and DALL-E have achieved impressive results. For instance, the LLaMA model requires 2,048 NVIDIA A100 GPUs for a duration of 21 days for training\cite{llama}, while the vision transformer in CLIP requires 8 core TPUv3 for approximately 30 days for training \cite{dosovitskiy2020image}.
    \item \textbf{Unified Architectures}: FMs provide a unified architecture that can be applied to multiple tasks, simplifying the model development process and fostering consistency in performance \cite{pan2024unifying}. This unified approach contrasts with the need to develop and train separate models for each task, streamlining the workflow and enhancing efficiency.
\end{itemize}

Besides, Table \ref{tab:comparison} also contrasts \methodname{} and traditional FMs across a diverse spectrum of topics ranging from data efficiency, data privacy, model performance, system operation, scalability, deployment, consistency and latency. All of these factors have collectively propelled the evolution and widespread adoption of trustworthy AI.


\subsection{Federated Learning}
\label{sec:FL}
FL aims to collaboratively train models across multiple decentralized data owners holding potentially sensitive local data in a privacy-preserving manner \cite{kairouz2021advances}. 
The essence of FL lies in its ability to learn a shared model by aggregating locally computed updates, rather than directly accessing or sharing the raw data. This ML paradigm not only enhances privacy and security, but also enables the utilization of distributedly owned data for model training, making it particularly suitable for mission critical applications, such as healthcare \cite{nguyen2022federated} and finance \cite{long2020federated}, where data privacy is of paramount concern. The motivation behind FL is driven by several key factors:
\begin{itemize}
    \item \textbf{Data Privacy and Security}: In many domains, especially healthcare and finance, data are highly sensitive and cannot be shared directly due to privacy concerns and regulatory constraints \cite{li2020federated}. FL addresses this by ensuring that raw data remain locally and only model updates are shared, significantly reducing the risk of privacy breaches.
    \item \textbf{Utilization of Distributed Data}: Organizations often have valuable data that is distributed across various locations. Traditional centralized training methods cannot effectively leverage this distributed data without transferring it to a central server, which is often impractical \cite{yang2019federated}. FL enables the use of this distributed data for training robust models without requiring data centralization.
    \item \textbf{Reduction of Communication Overhead}: Transferring large datasets to a central server for training can be resource-intensive and time-consuming \cite{kairouz2021advances}. FL reduces this overhead by only transmitting model updates, which are typically much smaller in size compared to the raw data.
    \item \textbf{Scalability and Collaboration}: FL facilitates collaboration among multiple entities, such as hospitals or financial institutions, allowing them to build more accurate and generalizable models by leveraging diverse datasets. This collaborative approach enhances the scalability of AI solutions across different domains\cite{yang2019federated}.
    \item \textbf{Enhanced Model Robustness}: By training on data from multiple sources, FL can produce models that are more robust and generalizable \cite{li2020federated}. This is particularly important for applications where the model needs to perform well across different environments and conditions.
\end{itemize}

The typical FL process involves several key steps. Initially, a global model is distributed to all participating FL clients from the FL server. Each client then trains the ML model on its local data to derive an updated local model. These model updates are subsequently sent back to the FL server, and then are aggregated to update the global model. Such cycle is repeated until convergences or specific performance criteria are met.
A widely used framework in FL is Federated Averaging (FedAvg) \cite{mcmahan2017communication}, renowned for its ability to aggregate model updates with minimal communication overhead. This is particularly crucial in FL settings, where a potentially large number of participants may have limited communication bandwidth \cite{yang2019federated}.

To achieve meaningful real-world impact, FL must address several intricate challenges, including non-independently and identically distributed (non-IID) data across clients, system heterogeneity, and scalability \cite{kairouz2021advances}. Non-IID data can introduce biases into models by favoring the data distribution of specific participants \cite{tan2022towards}. System heterogeneity, characterized by variations in computation and communication capabilities among client participants, can result in uneven contributions to the model training process. Additionally, scalability concerns arise as the number of participants increases, necessitating the development of efficient algorithms and infrastructure to manage the aggregation of updates and distribution of the global model.
The development of FL continues to evolve, spurred by advancements in both theoretical research and practical applications across various industries. FL is anticipated to play a pivotal role in building AI solutions that prioritize user privacy and data sovereignty.

\section{The Proposed \methodname{} Taxonomy}
\label{sec:taxonomy}

\begin{figure*}[!htbp]
    \centering
    \includegraphics[width=0.80\textwidth]{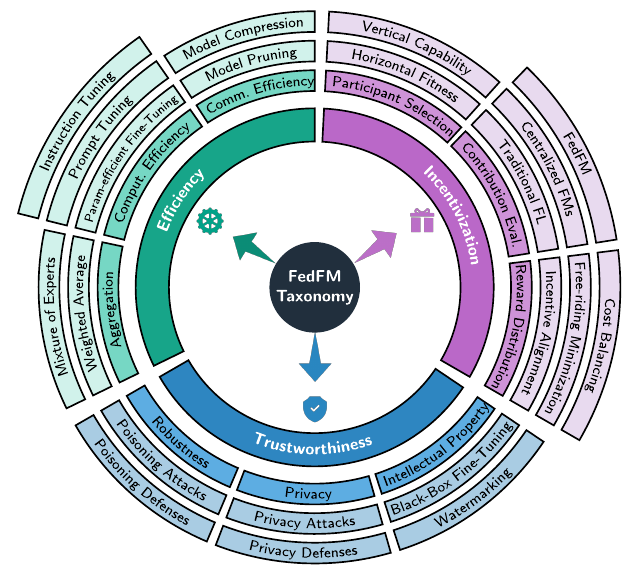}
    \caption{The proposed taxonomy of \methodname{}.}
    \label{fig:taxonomy}
\end{figure*}

In the rapidly evolving domain of FL, the integration of FMs into federated settings, termed \methodname{}, represents a significant leap forward. This integration aims to leverage the capabilities of FMs and the collaborative training process of FL to enable \methodname{} to access privately owned decentralized data. The development of \methodname{} requires rethinking of several key aspects of the current FL paradigm: 1) achieving efficiency in federated training and aggregation, 2) ensuring trustworthiness, and 3) building effective incentive mechanisms. Each part plays a crucial role in the successful implementation and operation of \methodname{}. Based on the above considerations, we propose a novel and multi-tiered \methodname{} taxonomy, as illustrated in Fig. \ref{fig:taxonomy}. 

\subsection{Efficient Federated Training and Aggregation: The Foundation of \methodname{}}
Research on FL training and aggregation methods for \methodname{} seeks to address the unique challenges posed by the size and complexity of FMs. The goal is to adapt traditional FL training methods to work efficiently with FMs, minimizing computational efficiency and communication overheads. This includes finding ways to effectively aggregate parameter updates involving FMs from potentially a large number of FL clients without overwhelming the communication network.

Existing federated training and aggregation research that holds potential for \methodname{} generally focus on three main areas: 1) advanced aggregation methods, 2) enhancing computational efficiency, and 3) enhancing communication efficiency. We summarize key trends in each of these areas:
\begin{itemize}
    \item \textbf{\methodname{} Aggregation}: Under this topic, we explore weighted averaging-based methods and promising new techniques. Weighted averaging is a widely adopted federated aggregation approach for local model updates in \methodname{}. Despite its simplicity, such methods are favored due to their efficiency and the computational challenges of more complex aggregation techniques given the vast scale of FMs. Nevertheless, innovative aggregation strategies are emerging which can be more effective for \methodname{}. Simple methods, such as model soups, which have shown promising results in enhancing model accuracy and robustness by averaging weights of models fine-tuned with different hyper-parameters. In addition, Mixture of Experts (MoE)-based models can be useful for developing sophisticated aggregation strategies that improve the performance of \methodname{}.
    \item \textbf{Computationally Efficient \methodname{}}: Research in this area focuses on adapting pre-trained FMs to specific tasks with minimal adjustments to the model parameters. Techniques like Parameter-Efficient Fine-Tuning (PEFT), Prompt Tuning (PT), and Instruction Tuning (IT) are possible ways to enhance \methodname{} computational efficiency. These methods allow for significant reductions in the computational and storage demands by only fine-tuning a small subset of the model parameters.
    \item \textbf{Communication Efficient \methodname{}}: Research in this areas focuses on strategies to enhance efficiency in transmitting FM updates between FL clients and the FL server. It generally involves two primary approaches: 1) model pruning, which selectively transmit important FM parameters; and 2) model compression, which aims reduce the size of the model being exchanged. These strategies are crucial for managing the increased communication overhead introduced by the large sizes of FMs and enhancing the scalability of \methodname{}.
\end{itemize}

This part of the review underscores the evolving landscape on \methodname{} research, detailing the challenges, strategies, and innovations in this interdisciplinary field. The focus on scaling, communication, and computational efficiency reflects the nuanced considerations required to effectively deploy \methodname{}.

\subsection{Trustworthiness: A Crucial Pillar of \methodname{} Integrity}
Research on trustworthy \methodname{} encompasses strategies to ensure that the system is robust against attacks and preserves the privacy of participants' data. This involves developing mechanisms to protect against poisoning attacks which aim to corrupt the model and privacy-preserving techniques. The focus is on maintaining the integrity of the learning process and safeguarding participant data from breaches.

Trustworthy \methodname{} address the critical areas of robustness, privacy preservation, and intellectual property in the context of \methodname{}, exploring the challenges and proposing solutions to ensure the integrity and security of these systems. We summarize key trends in each of these areas:
\begin{itemize}
    \item \textbf{Robustness}: Research in this area discusses how poisoning attacks, both untargeted and targeted, aim to compromise the integrity of global models in FL systems. Untargeted attacks disrupt the training process to prevent convergence, while targeted attacks manipulate the model output subtly. The complexity and heterogeneity of \methodname{} training tasks make these attacks particularly challenging to execute and defend against. Byzantine-robust aggregation rules are also being proposed to counteract poisoning attacks, including geometrical outlier detection, top performance selection, and other hybrid schemes. However, the effectiveness of these defenses has been questioned in \methodname{} settings, calling for novel approaches tailored to its complexities.
    \item \textbf{Privacy Preservation}: Research on privacy attacks explore the threat landscape of FL, including membership inference and data reconstruction attacks. Membership inference attacks aim to determine whether specific data samples were used in training, while data reconstruction attacks seek to recreate the original training data. The large scales of FMs and the nature of \methodname{} introduce new vulnerabilities and challenges in defending against these privacy attacks. Research on privacy defenses focuses on developing strategies to counteract privacy attacks, focusing on a balance between preserving knowledge integrity and ensuring privacy. Techniques like differential privacy, confidence masking, and model compression are highlighted as means to protect client data privacy in \methodname{} without significantly compromising the quality of local model updates.
    \item \textbf{Intellectual Property Protection}: Research in this area emphasizes the importance of intellectual property protection for safeguarding the ownership of \methodname{} against unauthorized usage, such as model theft. Two primary strategies are being explored in the context of \methodname{}: black-box fine-tuning and watermarking. Black-box fine-tuning allows for model adaptation while keeping the core intellectual property intact, whereas watermarking embeds identifiable markers within the model to assert ownership. The integration of these methods ensures that \methodname{} can be securely and efficiently utilized across distributed networks. Nevertheless, the effectiveness of these strategies in the diverse and complex environments typical of \methodname{} has prompted ongoing discussions and investigations into more robust and tailored approaches.
\end{itemize}

This part of the review on \methodname{} research underscores the intricate balance required to achieve trustworthiness, emphasizing the need for innovative solutions to address the dual challenges of robustness against poisoning attacks, privacy preservation, and intellectual property. The complexity of FMs and the FL environment necessitates a re-evaluation of traditional defense mechanisms and development of new strategies to ensure the integrity and privacy of such advanced learning systems.

\subsection{Incentive Mechanisms: Fostering Participation, Collaboration, and Adaptability of \methodname{}}
The participation of data owners is vital for the success of \methodname{}. Incentive mechanisms are often leveraged to motivate data owners to contribute their local data and computational resources. These mechanisms often draw upon economic and game theories to fairly compensate participants for their contributions. The challenge lies in creating a system that balances the need to encourage participation with the practicalities of managing the distribution of rewards.
Research in \methodname{} incentive mechanism design delves into the crucial aspects of FL participant selection, contribution evaluation, and reward distribution. We summarize key trends in each of these areas:
\begin{itemize}
    \item \textbf{Participant Selection}: Research in this area develops strategies for selecting participants to join the FL process, emphasizing the importance of both traditional approaches (e.g., Contract Theory, Game theory, auctions) and emerging model-centric approaches. Contract Theory-based mechanisms are employed in scenarios with information asymmetry, focusing on computation and communication resource optimization before determining rewards. Game theoretic approaches aim to optimize resources, while offering incentives through server-client negotiations. Auction-based schemes focus on attracting high-quality participants by addressing latency and resource burdens efficiently. Model-centric approaches select FL participants based on the characteristics of participants' models rather than just the resources they hold, introducing the concepts of horizontal task-specific training fitness and vertical training capability.
    \item \textbf{Contribution Evaluation}: Research in this area addresses the challenges of evaluating participant contributions in \methodname{}, particularly dealing with the impracticality of Shapley Value-based methods due to high latency and the compositional gap inherent in FMs. The complexity of FMs and the sheer volume of participants amplify the latency in training and inference, making timely feedback challenging. FMs face difficulties in generating correct answers to compositional problems, indicating that direct contribution evaluation for FMs built by \methodname{} is beyond current capabilities.
    \item \textbf{Reward Distribution}: Research in this area focuses on the intricate balance required in designing reward distribution mechanisms that align with the objectives of \methodname{} (e.g., deterring free-riding, managing the costs associated with rewards to ensure project sustainability). Incentive alignment ensures that rewards are structured to motivate FL participants and align with the \methodname{} goals. Minimizing free-riding requires mechanisms to ensure participants cannot benefit from \methodname{} without making positive contributions. Cost management strikes a delicate balance between incentivizing FL participants and maintaining the \methodname{} financial viability.
\end{itemize}

\methodname{} incentive mechanism design emphasizes the complexity of effectively motivating data owner participation. It suggests that addressing the interconnected challenges of participant selection, contribution evaluation, and reward distribution is crucial for the success and sustainability of \methodname{}.

\subsection{Summary of the Proposed \methodname{} Taxonomy}
The proposed \methodname{} taxonomy covers the fundamental aspects of \methodname{}, highlighting the need for robust training and aggregation methods, trustworthiness in the form of Byzantine-robustness and privacy preservation, and novel incentive mechanisms to encourage active and meaningful participation. These areas are where the involvement of FMs necessitates significant revisions to the current FL techniques. In subsequent sections, we conduct comprehensive literature review on each of these topics, and put forth a vision for promising future research directions.

\section{Towards Efficient \methodname{}}
\label{sec: training and aggregation}
\begin{figure*}[!t]
\centering
\includegraphics[width=\linewidth]{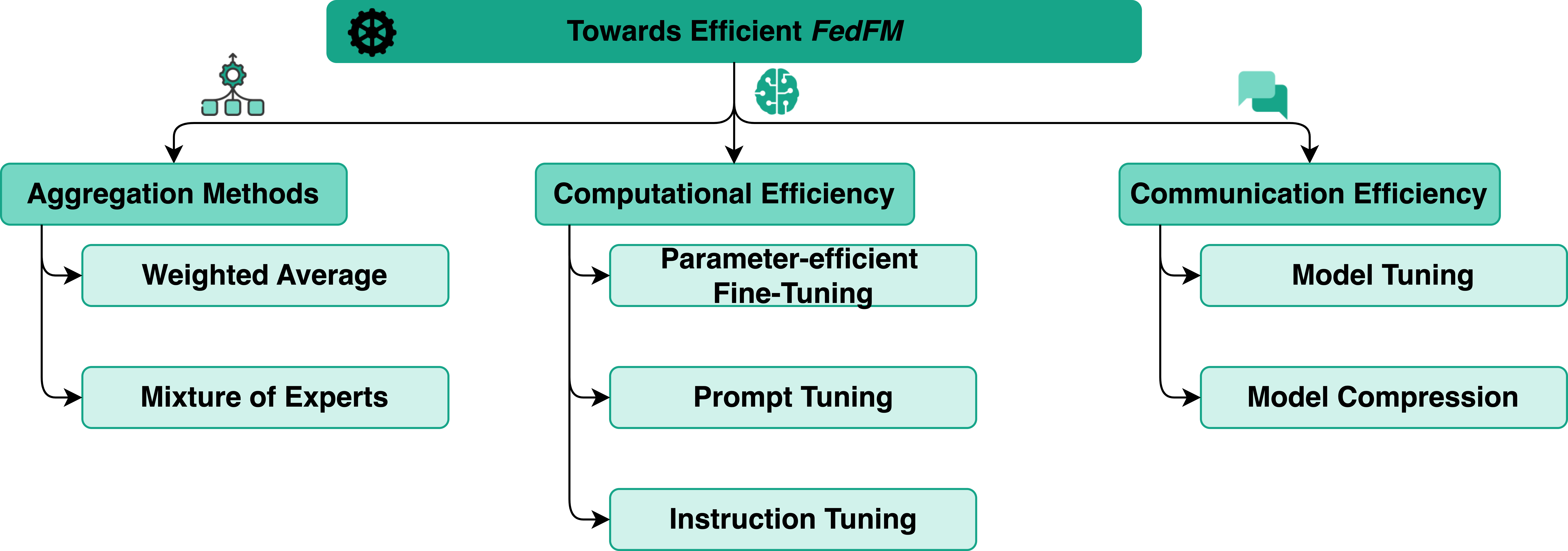}
\caption{Taxonomy of enabling techniques for efficient \methodname{}. The diagram categorizes existing methods into three main domains: 1) Aggregation, 2) Computational Efficiency, and 3) Communication Efficiency. Each domain is further divided into specific strategies that address the challenges of  \methodname{}.}
\label{fig:efficient}
\end{figure*}

In this section, we review and discuss current literature on the topic of efficient collaborative model training/fine-tuning and aggregation for \methodname{}, focusing on emerging settings and scenarios that arise as a result of combining FL with FMs. We further break down each dimension into sub-areas for more detailed analysis (Fig. \ref{fig:efficient}).

\begin{algorithm}[b!]
\caption{Opportunities for efficient training and aggregation of FMs under the existing FL procedure}
\label{alg:FL}
\begin{algorithmic}[1]
\STATE \textbf{Input:} $N$ clients, local epochs $E$, learning rate $\eta$, initial global model weights $\vect{\theta}^1$, total number of communication round $R$
\FOR{each round $t = 1, 2, \ldots, R$}
    \STATE Server selects a subset of $m$ clients $S_t$
    \FOR{each client $i \in S_t$ \textbf{in parallel}}
        \STATE $\vect{\theta}_i^{t+1} \leftarrow \text{ClientUpdate}(i, \vect{\theta}^t)$
    \ENDFOR
    \STATE $\vect{\theta}^{t+1} \leftarrow \text{Aggregate}(\{ \vect{\theta}_i^{t+1}\}_{i=1}^{m})$ 
\ENDFOR 
\STATE \hrulefill
\STATE \textbf{procedure} 
\text{ClientUpdate}($i, \vect{\theta}^t$): 
    \STATE $\mathcal{B} \leftarrow$ (split $\mathcal{D}_c$ into batch of size $B$)
    \FOR{each local epoch $i$ from 1 to $E$}
        \FOR{batch $b \in \mathcal{B}$}
            \STATE $\vect{\theta}_i^t \leftarrow \vect{\theta}^{t} - \eta \nabla \mathcal{L}(\vect{\theta}^t; b)$
            \STATE \textcolor{black}{\textit{// \textbf{Q2)} How to improve the computational efficiency so that FMs can be efficiently hosted and updated on FL clients?}} 
        \ENDFOR
    \ENDFOR
    \STATE \textbf{return} $\vect{\theta}_i^t $ to the server
    \STATE \textcolor{black}{\textit{// \textbf{Q3)} How to improve communication efficiency when $\vect{\theta}_i^t$ is large?}}
\STATE \hrulefill
\STATE \textbf{procedure} \text{Aggregate}($\{ \vect{\theta}_i^{t+1}\}_{i=1}^{m}$)
\STATE {FedAvg \cite{mcmahan2017communication}:} $\vect{\theta}^{t+1} \leftarrow \frac{1}{m} \sum_{i=1}^{m} \vect{\theta}_i^{t+1}$
\STATE \OR
\STATE \textcolor{black}{\textit{// \textbf{Q1)} Aggregation methods for FMs to enhance performance?}}
\end{algorithmic}
\end{algorithm}

\subsection{\methodname{} Training and Aggregation}
To grasp the challenges involved in integrating FMs into the FL process, a revision of the general FL procedure is necessary. FL typically operates under the control and decision-making of the central server with the primary purpose of collaboratively training a Deep Learning (DL) model $\vect{\theta}$ across $N$ clients over multiple rounds, as outlined in Algorithm \ref{alg:FL}. The integration of FMs, with their considerable sizes, introduces challenges at various stages of the FL process. These challenges can be distilled into three critical research questions (highlighted in yellow in Algorithm \ref{alg:FL}):
\begin{itemize}
    \item[\textbf{Q1)}] \textbf{How to aggregate FMs to enhance performance?} Given that one of the primary goals of FL is to improve the global model performance, the challenge lies in designing aggregation approaches that are suited for the scale and complexity of FMs. Despite the large body of existing FL model aggregation techniques, few are tailored for FMs.
    \item[\textbf{Q2)}] \textbf{How to improve FL computational efficiency so that FMs can be efficiently hosted and updated by FL clients?} In traditional FL, especially in cross-device scenarios where client devices often have limited memory and compute power, finding ways to efficiently host and update models is already a significant challenge. The substantial increase in the size of FMs exacerbates this challenge, highlighting the need for techniques with significantly improved computational efficiency.
    \item[\textbf{Q3)}] \textbf{How to improve communication efficiency?} FL often involves transmitting model updates between clients and the server. The enormous size of FMs creates significant communication overhead that overwhelms the network. Thus, developing more efficient communication strategies is essential for the scalability of \methodname{}.
\end{itemize}

\begin{figure*}[t]
    \centering
    \includegraphics[width=1.0\linewidth]{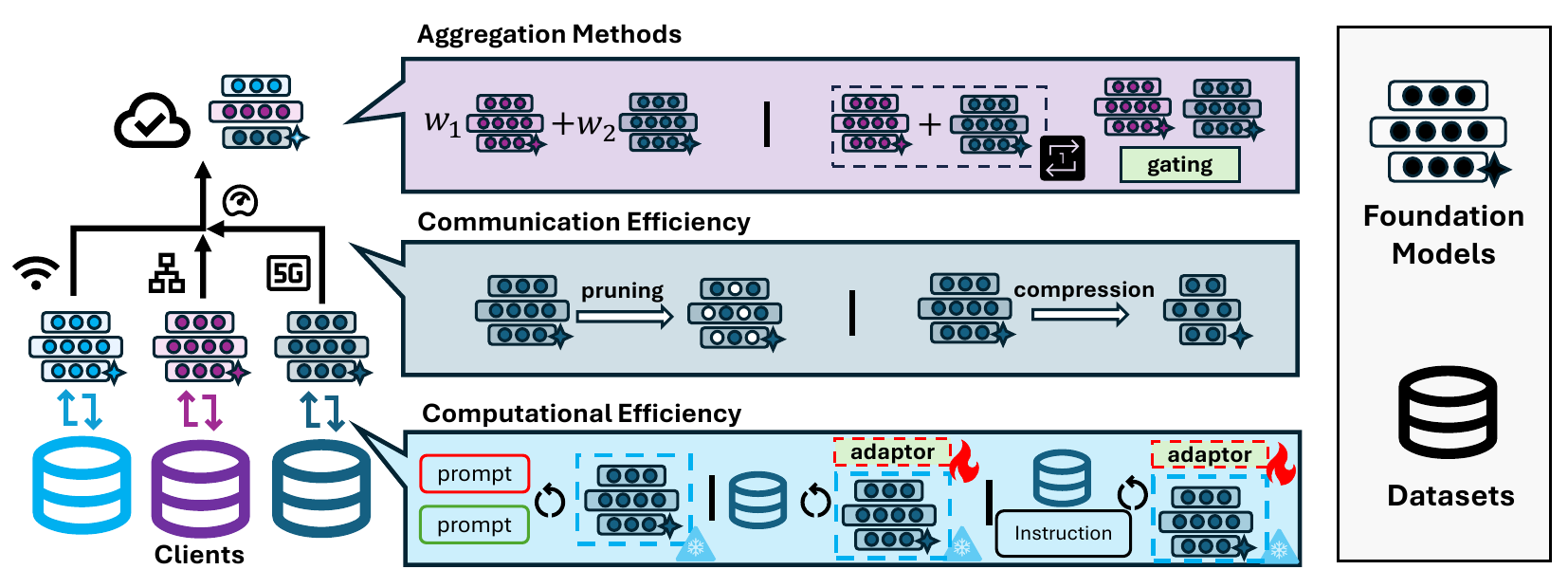}
\caption{Illustration of efficient \methodname{}.}
    \label{fig:fig-efficient FedFM}
\end{figure*}

Therefore, our survey examines innovative methods and strategies useful for \methodname{} from three dimensions: 1) aggregation methods for enhanced effectiveness, 2) techniques for boosting computational efficiency, and 3) techniques for improving communication efficiency. We further break down each domain into specific areas as shown in Fig. \ref{fig:fig-efficient FedFM}.

For aggregation methods, we examine the traditional \textit{weighted average} approach, which is the prevailing approach in current \methodname{} works. In addition, we explore a selection of promising aggregation techniques, which can be particularly effective for \methodname{}, offering insights into their potential applicability and benefits.

For computational efficiency, we divide existing approaches into three categories.  
\begin{itemize}
    \item \textit{Parameter-efficient Fine-tuning}: which aims to adapt a pre-trained FM to a specific task by adjusting only a minimal subset of its parameters.
    \item \textit{Prompt Tuning}: which seeks to enhance FM performance without direct model training, but with carefully crafted textual prompts instead.
    \item \textit{Instruction Tuning}: which fine-tunes FMs by making the model learn to follow and execute a sequence of instructions instead of traditional input-output pairs.
\end{itemize}

For communication efficiency, we divide existing approaches into two categories.
\begin{itemize}
    \item \textit{Model Pruning}: which focuses on selectively transmitting critical parameters between clients and the server, ensuring only the most vital information is shared.
    \item \textit{Model Compression}: which aims to decrease the total number of the model parameters being transmitted.
\end{itemize}

\subsection{Scaling Efforts for \methodname{}}
\label{sec:scale}
It is useful to recognize that traditional FL typically involves models with fewer than 10 million parameters.
For instance, the largest base model aggregated by FedAvg~\cite{mcmahan2017communication} is typically a stacked LSTM with approximately 5 million parameters. The seminal paper of FedProx~\cite{li2020federated} is tested with a stacked LSTM with roughly 2 million parameters. The largest base model tested with SCAFFOLD~\cite{scaffold} is a 2-layer CNN with an estimated parameter count under 5 million. With transformer-based models, FL techniques are starting to be tested on larger models. However, to significantly scale up this respect of FL not only requires new techniques, but also new experiment designs for improved evaluation. 
Although recent studies have started to evaluate new FL methods with larger base models, they are still markedly smaller than FMs which are achieving breakthrough performance under centralized training settings.

\captionsetup[table]{labelformat=simple, labelsep=newline, textfont=sc, justification=centering}

\begin{table*}[!t]
\caption{Sizes of FMs and Aggregation under FL Setting}
\label{tab: model_size and aggregation}
\centering
\begin{tabular}{l | c c c | p{9cm}}
\toprule
{\textsc{Methods}} & {\textsc{Year} $\downarrow$} & {\textsc{Size (M)}}  & {\textsc{Aggregation}} & {\textsc{Model Description}} \\
\midrule
\makecell[l]{FedPAQ \\ \cite{reisizadeh2020fedpaq}} & 2020 & 0.2 & FedSGD & \makecell[l]{Implement quantization techniques to reduce communication \\ overhead in FL by compressing model updates}\\
\rowcolor{mygray} \makecell[l]{HeteroFL \\  \cite{diao2020heterofl}} & 2020 & 11 & FedAvg & \makecell[l]{Support heterogeneous model architectures across different \\clients, optimizing the training process based on individual \\ device capabilities}\\
\makecell[l]{LotteryFL\\  \cite{li2020lotteryfl}} & 2020 & 138 & FedAvg & \makecell[l]{Apply the lottery ticket hypothesis to identify and retain \\ efficient sub-networks to improve training efficiency}\\
\rowcolor{mygray} \makecell[l]{FjORD\\  \cite{horvath2021fjord}} & 2021 & 11& FedAvg & \makecell[l]{Utilize ordered dropout to maintain a structured hierarchy \\ of knowledge within the model, facilitating efficient \\ extraction of sub-models}\\
\makecell[l]{H-FL\\  \cite{yang2021h-fl}} & 2021 & 138 & FedAvg &\makecell[l]{Address statistical heterogeneity among clients by \\ employing lossy SVD and bias correction mechanisms}\\
\rowcolor{mygray} \makecell[l]{PruneFL \\ \cite{jiang2022pruneFL}} & 2022 & 132 & FedAvg & \makecell[l]{Introduce a two-stage pruning process to reduce model size \\ while preserving performance, tailored for FL environments}\\
\makecell[l]{FedPM \\ \cite{isik2022fedpm}} & 2022 & 12 & FedAvg & \makecell[l]{Use a binary mask approach inspired by the lottery ticket \\ hypothesis to prune models, enhancing efficiency}\\
\rowcolor{mygray} \makecell[l]{FedTiny \\ \cite{huang2023fedtiny}} & 2022 & 132 & FedAvg & \makecell[l]{Enhance adaptability by using batch normalization statistics \\ for initializing models, suitable for diverse data distributions \\ across clients}\\
\makecell[l]{SoteriaFL \\ \cite{li2022soteriafl}} & 2022 & 0.05 & FedSGD & \makecell[l]{Combine differential privacy and communication compression \\ to protect data while reducing the overhead in FL}\\
\rowcolor{mygray} \makecell[l]{FedPrompt \\ \cite{zhao2023fedprompt}} & 2022 & 223 & FedAvg & \makecell[l]{Focus on prompt tuning techniques for fine-tuning pre-trained \\ models in federated environments, optimizing performance on \\ specific tasks}\\
\makecell[l]{FedBERT \\ \cite{tian2022fedbert}}& 2022 & 117 & FedAvg & \makecell[l]{Adapt the BERT model for FL, enabling pre-training and \\fine-tuning across distributed datasets}\\
\rowcolor{mygray} \makecell[l]{FedCLIP \\ \cite{lu2023fedclip}} & 2023 & 85 & FedAvg & \makecell[l]{Adapt the CLIP for federated settings by leveraging adapter-\\based fine-tuning to handle multi-modal tasks effectively}\\
\makecell[l]{FedPEFT \\ \cite{fedpeft2022}} & 2023 & 85 & FedAvg & \makecell[l]{Introduce parameter-efficient fine-tuning methods that modify \\only few parameters, reducing the burden on client devices}\\
\rowcolor{mygray} \makecell[l]{FedPETuning \\ \cite{zhang2023fedpetuning}} & 2023 & 125 & FedAvg & \makecell[l]{Conduct benchmark analyses of adapter-based fine-tuning \\techniques, assessing their efficiency in FL}\\
\makecell[l]{SLoRA \\ \cite{babakniya2023slora}} & 2023 & 67 & FedAvg & \makecell[l]{Optimize LoRA for non-IID federated settings, using \\parameter moderation to handle diverse data distributions}\\
\rowcolor{mygray} \makecell[l]{FedOBD \\ \cite{chen2022fedobd}} & 2023 & 17 & FedAvg & \makecell[l]{Segment FMs into semantic blocks for selective transmission, \\reducing communication costs while maintaining performance}\\
\makecell[l]{FedIT \\ \cite{zhang2023fedit}} & 2023 & 7,000 & FedAvg & \makecell[l]{Apply instruction tuning to the LLaMA model, allowing it to \\handle a variety of tasks simultaneously in a FL setting}\\
\rowcolor{mygray} \makecell[l]{FwdLLM\\\cite{xu2023fwdllm}} & 2024 & 7,000 & FedSGD & \makecell[l]{Combines Backpropagation-free training with \\ parameter-efficient methods to adapt FMs for mobile devices \\with limited resources}\\
\bottomrule 
\end{tabular}
\vspace{-3pt}


\end{table*}

We list the methods reviewed alongside with the largest base models involved in their evaluations in Table \ref{tab: model_size and aggregation} and Table \ref{tab:model_size-2}. The substantial difference for model sizes between existing FL research and centralized FMs might cast doubts on the feasibility and practicality of the FL techniques for \methodname{}, calling for further experimental studies involving large-scale FMs. Such an effort depends on the availability of frameworks that can streamline implementation. Thus, in Section \ref{sec:framework}, we analyze the support for FMs in existing FL frameworks.

\begin{table*}[!htp]
\caption{Sizes of FMs under Centralized Learning Setting}
\label{tab:model_size-2}
\centering
\begin{tabular}{l | c c | p{11.5cm}}
\toprule
{\textsc{Methods}} & {\textsc{Year} $\downarrow$} & {\textsc{Size (M)}}  &  {\textsc{Model Description}} \\
\midrule
\makecell[l]{BERT \\ \cite{devlin2018bert}} & 2018 & 340 & \makecell[l]{A transformer-based model pre-trained on large text corpora, widely used for\\ various NLP tasks}\\
\rowcolor{mygray} \makecell[l]{GPT-1 \\ \cite{radford2018improving}} & 2018 & 117 & \makecell[l]{The first Generative Pre-trained Transformer (GPT) model, designed for \\language modeling tasks, setting the foundation for subsequent versions}\\
\makecell[l]{GPT-2 \\ \cite{radford2019language}} & 2019 & 1,500 & \makecell[l]{An improved version of GPT with significantly more parameters, enhancing \\performance on a range of NLP tasks}\\
\rowcolor{mygray} \makecell[l]{GPT-3 \\ \cite{brown2020language}} & 2020 & 175,000 & \makecell[l]{A large-scale language model with 175 billion parameters, capable of \\ performing a wide variety of language generation and comprehension tasks}\\
\makecell[l]{ViT \\ \cite{dosovitskiy2020ViT}} & 2020 & 632 & \makecell[l]{The Vision Transformer model by applying transformer architecture to image \\classification tasks, demonstrating high performance on standard benchmarks}\\
\rowcolor{mygray} \makecell[l]{T5 \\ \cite{raffel2020exploring}} & 2020 & 11,000 & \makecell[l]{The Text-to-Text Transfer Transformer, designed to convert all NLP tasks into \\ the text-to-text format, achieving state-of-the-art performance}\\
\makecell[l]{DALL-E \\ \cite{ramesh2021zero}} & 2021 & 12,000 & \makecell[l]{An image generation model that creates images from textual descriptions, \\showcasing the capabilities of generative models}\\
\rowcolor{mygray} \makecell[l]{CLIP \\ \cite{radford2021learning}} & 2021 & 1,000 & \makecell[l]{A multi-modal model that can understand and generate images and text, \\trained to connect textual descriptions with images}\\
\makecell[l]{Gato \\ \cite{reed2022generalist}} & 2022 & 1,200 & \makecell[l]{A generalist agent by DeepMind capable of performing various tasks across \\different modalities using a single neural network}\\
\rowcolor{mygray} \makecell[l]{PaLM\\ \cite{chowdhery2023palm}} & 2023 & 540,000 & \makecell[l]{A Pathways Language Model designed for scaling up to very large models, \\demonstrating strong performance across a wide range of NLP tasks}\\
\makecell[l]{LLaMA\\ \cite{touvron2023llama}} & 2023 & 69,000 & \makecell[l]{A large-scale language model designed for multi-task learning, integrating \\capabilities across different domains to enhance versatility}\\
\rowcolor{mygray} \makecell[l]{GPT-3.5\\ \cite{GPT-3.5}} & 2023 & 20,000 & \makecell[l]{GPT-3.5 Turbo models can understand and generate natural language or code \\and have been optimized for chat but work well for non-chat tasks as well}\\

\makecell[l]{GPT-4\\ \cite{OpenAI_GPT4_2023}} & 2023 & 1,800,000 & \makecell[l]{GPT-4 is large multimodal model (accepting text\textbackslash image inputs and outputting\\ text) that can solve difficult issues with greater accuracy than GPT-3.5}\\

\rowcolor{mygray} \makecell[l]{Chinchilla \\ \cite{hoffmann2022training}} & 2023 & 70,000 & \makecell[l]{An efficient language model designed by DeepMind, focusing on optimized \\ training efficiency and resource utilization}\\
\makecell[l]{Flamingo\\ \cite{alayrac2022flamingo}} & 2023 & 80,000 & \makecell[l]{A model designed for visual understanding tasks, integrating both vision and \\language inputs for improved performance}\\
\rowcolor{mygray} \makecell[l]{LLAVA \\ \cite{liu2024visual}} & 2024 & 15,000 & \makecell[l]{A vision-language model aimed at handling multi-modal tasks, combining\\ visual and textual data to improve task performance}\\
\makecell[l]{LLaMA 3\\ \cite{LLaMA-3}} & 2024 & 70,000 & \makecell[l]{The models are pre-trained on 15 trillion tokens of text gathered from publicly\\ available sources with the instruct models fine-tuned on instruction datasets}\\
\rowcolor{mygray}\makecell[l]{GPT-4o\\ \cite{GPT-4o}} & 2024 & Not disclosed & \makecell[l]{GPT-4o (“o” for “omni”) has the same high intelligence as GPT-4 Turbo but is \\much more efficient—it generates text 2x faster and is half cheaper}\\
\makecell[l]{LLaMA 3.1\\ \cite{LLaMA-3.1}} & 2024 & 405,000 & \makecell[l]{ An auto-regressive LLM that uses an optimized transformer architecture. The \\ tuned versions use supervised fine-tuning and reinforcement learning with \\human feedback to align with human preferences for helpfulness and safety}\\

\bottomrule 
\end{tabular}

\vspace{-1pt}

\end{table*}




\subsection{Aggregation Methods for \methodname{}}
\label{sec:aggregation}
\subsubsection{Weighted Average}
\
\par
\indent
As shown in Algorithm \ref{alg:FL}, the local training process of traditional FL yields a set of local models $\{\vect{\theta}_1^t, \vect{\theta}_2^t,\ldots, \vect{\theta}_m^t\}$ at communication round $t$. These models are transmitted to the server where they undergo an aggregation process to produce global model $\vect{\theta^{t+1}}\leftarrow \text{Aggregate}(\{\vect{\theta}_1^t, \vect{\theta}_2^t,\ldots, \vect{\theta}_m^t\})$. The aggregation step is pivotal in FL, as the overall FL model performance relies on it \cite{liu2023recent}. Consequently, the development of aggregation methods that balance efficiency with efficacy remains a central theme in FL research. The predominant approach involves computing a weighted average of the parameters from local models (i.e., the weighted average method). Prominent examples of this approach include FedAvg/FedSGD \cite{mcmahan2017communication}, FedProx \cite{li2020federated}, and FedMA \cite{wang2020fedma}.

\begin{figure}[t]
    \centering
    \includegraphics[width=1\linewidth]{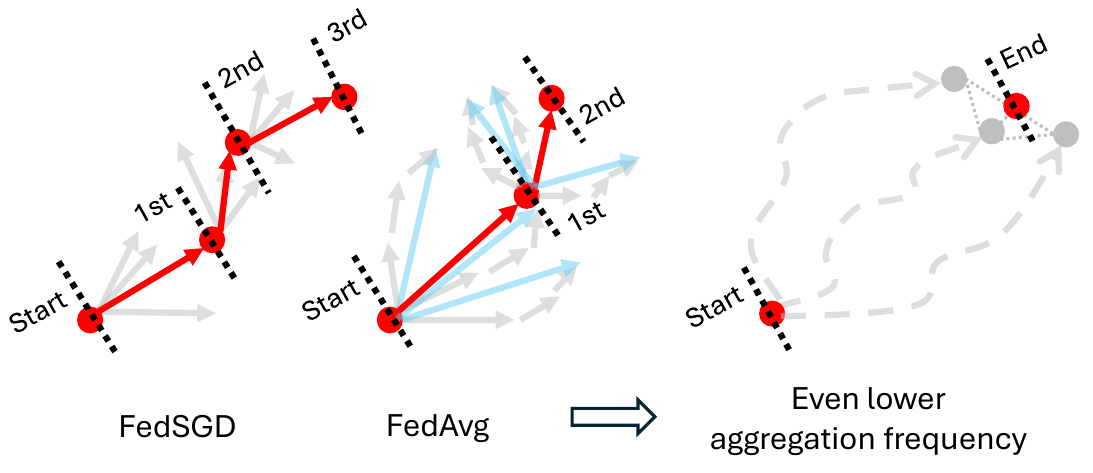}
\caption{Illustration of weight averaging strategy of aggregation methods in \methodname{}.}
    \label{fig:gradient}
    \vspace{-12pt}
\end{figure}

To the best of our knowledge, currently, there is no FL aggregation technique specifically tailored for \methodname{}. The prevailing practice has gravitated towards adopting the basic vanilla versions of FedAvg or FedSGD, with the latter being equivalent to FedAvg when the number of local epochs ($E$) is set to one, as shown in Fig. \ref{fig:gradient}. Table \ref{tab: model_size and aggregation} provides a summary of the aggregation methods adopted by existing \methodname{} works.




The preference for simple weighted averaging by existing \methodname{} methods has been driven by three primary factors. 
\begin{itemize}
    \item The scale of FMs results in a significant computational overhead for the aggregation process at the server-side, rendering more complex methods such as FedMA \cite{wang2020fedma} less appealing, particularly considering the computational challenge of running Hungarian algorithms on billions of parameters. FedAvg-based methods are not only efficient but also well-suited for parallelization, making them effective for aggregating a large number of parameters. 
    \item Recent benchmarking studies \cite{ye2024openfedllm} indicate that empirically, FedAvg-based techniques often outperform more complex methods across a variety of datasets and FMs. 
    \item Perhaps most crucially, many existing methods incorporate the weighted averaging concept inherent in FedAvg (e.g., by aggregating only a subset of parameters). These methods shall be discussed in detail in Section \ref{sec:computation}.  
    Some studies have demonstrated a preference for FedSGD over FedAvg \cite{reisizadeh2020fedpaq, li2022soteriafl, xu2023fwdllm} due to its superior performance, partially attributable to the asymmetric nature of computational and network costs \cite{xu2023fwdllm}.
\end{itemize}

\subsubsection{Mixture of Experts}
\
\par
\indent
In traditional FL, there is commonly a trade-off between the complexity, frequency and overhead of aggregation methods and the performance of the resulting global model. Increased complexity and frequency, along with increased overhead, nudge FL closer to distributed learning, where aggregation occurs after every local iteration, leading to low efficiency. Conversely, reducing these factors can enhance efficiency but often results in reduced model performance. With bandwidth being a limiting factor, the significantly larger sizes of FMs tip the balance in favour of simpler aggregation techniques. 

Recently, promising studies have emerged, suggesting that simpler aggregation methods can be effective. The concept demonstrated in \cite{wortsman2022model} shows that averaging the weights of several models, each fine-tuned with varying hyper-parameters, can enhance both accuracy and robustness. This finding has notably been observed with large pre-trained models. This phenomenon suggests that aggregating independently trained FMs just once after training might still yield performance improvements. Building on this insight, \cite{pmlr-v202-rame23a} advocates for the utilization of FMs that have been fine-tuned across a spectrum of tasks, by averaging all the fine-tuned weights to produce the final model. This final model shows enhanced generalization to out-of-distribution data, indicating that averaging FMs, particularly those fine-tuned on disparate local tasks, can be a promising aggregation technique for \methodname{}.

\begin{figure}[t]
    \centering
    \includegraphics[width=1\linewidth]{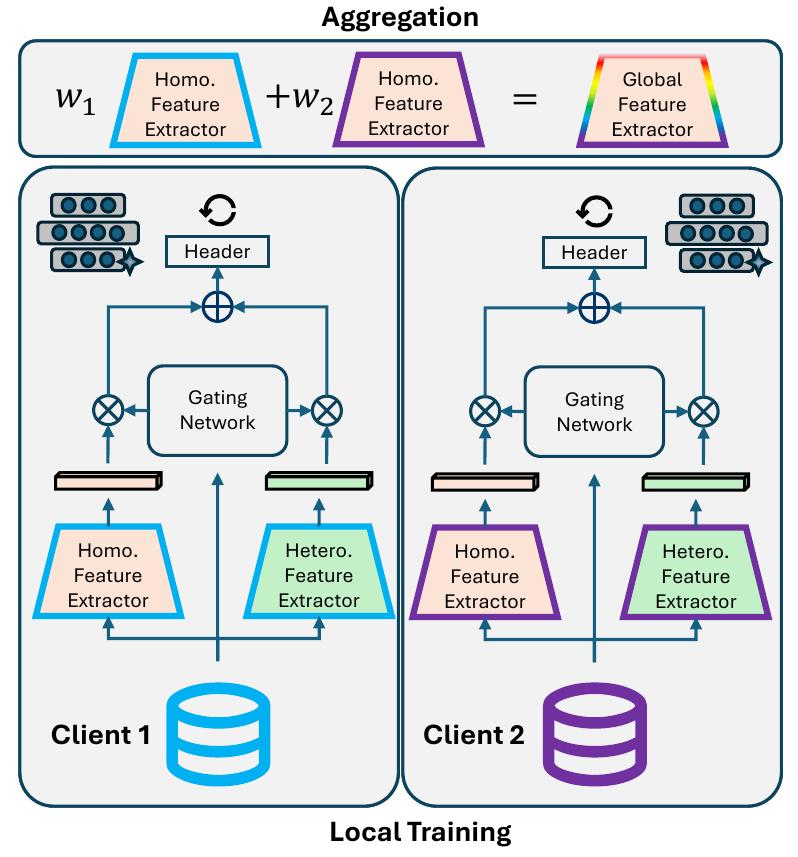}
\caption{MoE strategies of \methodname{} aggregation methods.}
\vspace{-6pt}
    \label{fig:moe}
\end{figure}

On the other hand, the potential for more sophisticated aggregation strategies remains. 
{The emergence of large language models based on Mixture of Experts (MoE)\cite{MoE}},
exemplified by GLaM \cite{du2022glam} and ST-MoE \cite{zoph2022st}, has demonstrated exceptional capabilities across various tasks. These MoE architectures incorporate a myriad of smaller, specialized sub-models orchestrated by a routing function {(gating network)} that aligns with the principle of dedicated local models in FL, as shown in Fig. \ref{fig:moe}. The efficacy of MoE models is significantly influenced by their learned routing mechanisms \cite{xue2024openmoe}. This concept resembles the aggregation strategies in FL, where local model outputs are synthesized to form a global model. The routing mechanism in MoE could inspire a novel class of aggregation strategies in FL, where dynamic data-driven approaches could allocate model contributions in a similar manner. Exploring aggregation strategies that mirror MoE's routing scheme and adapting it to the distributed nature of FL, could inspire new \methodname{} model aggregation methods. Such exploration might involve designing algorithms that not only learn from local data, but also intelligently decide how to combine local updates to form a robust global model, potentially redefining the \methodname{} aggregation approach as a learned component rather than a fixed rule.

MoE-FL \cite{MoE-FL} uses MoE to aggregate a robust global model on the server. It assumes trust between the server and clients where each client shares partial data with the server.
Although the produced robust aggregation model can filter outliers (poisoned/corrupted/outdated client models), sharing partial local data may be prohibited in most FL applications with privacy requirements.
To address the data heterogeneity problem in FL, FedMix \cite{FedMix} allows each client to utilize MoE to adaptive select which client expert models are relevant to it.
Fed-MoEs \cite{Fed-MoEs} obtain all client local models pre-trained on local data. These pre-trained models, the aggregated global model, and the private client model form the experts of MoE on each client. The client uses MoE to dynamically combine these experts for domain adaptation, addressing the domain shift problem. However, it introduces significant communication overhead due to the need to transmit multiple models to each client.

PFL-MoE \cite{PFL-MoE} combines MoE with personalized FL (PFL) to improve model personalization, while maintaining model generalization. It constructs an MoE for each client, involving the gating network and two expert models - the global model and the local personalized model. Each client uses the gating network to generate two weights for the outputs of the two expert models, and mixes the weighted outputs as the final output.
pFedMoE \cite{pFedMoE} first introduced MoE into model-heterogeneous personalized FL (MHPFL) to enable collaboration among heterogeneous client models with efficient communication and computation. Each client is constructed with an MoE formed by a gating network, the shared homogeneous feature extractor as the global expert, and the heterogeneous client model as the local expert. For each local data sample, the gating network produces a pair of personalized weights to represent the two experts. The client performs weighted averaging on them to form a complete representation containing both generalized and personalized features.

\subsection{Computationally Efficient \methodname{}}
\label{sec:computation}
\begin{figure*}[t]
    \centering
    \includegraphics[width=0.7\linewidth]{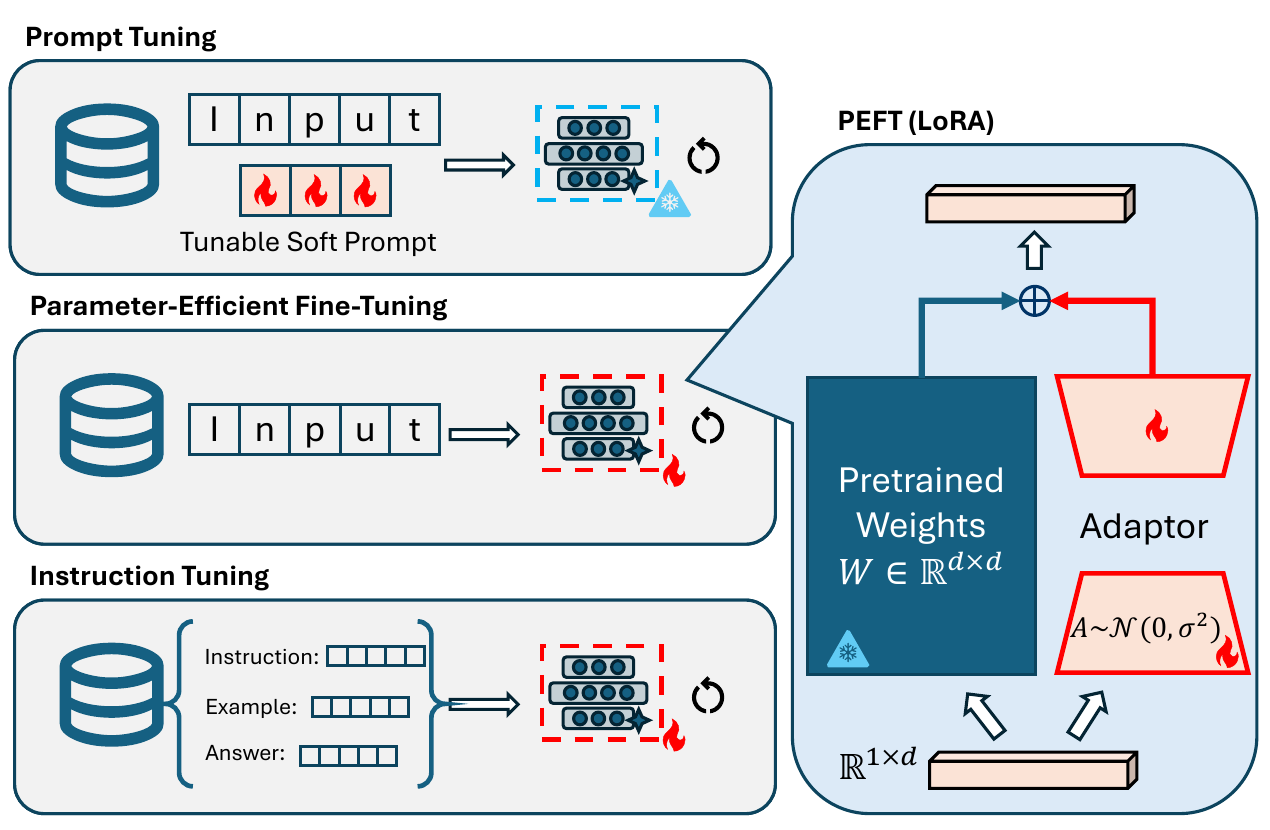}
\caption{Methods for enhancing \methodname{} computational efficiency.}
\vspace{-6pt}
    \label{fig:computation}
\end{figure*}

This section delves into recent advances in training FMs via FL techniques, with a focus on enhancing computational efficiency. Specifically, we examine a set of methods including: \textit{1) Parameter-Efficient Fine-Tuning (PEFT)}, \textit{2) Prompt Tuning}, and \textit{3) Instruction Tuning}. These three types of methods have been shown to be feasible, as shown in Fig. \ref{fig:computation}.

Traditionally, FL has predominantly involved full model training, a process where all parameters of a model are trained from the ground up. This approach includes strategies for training entire \methodname{} framework. For instance, \cite{tian2022fedbert} explores federated pre-training of models within the BERT family, which can include up to 117 million parameters. As a foundational architecture for many FMs, BERT represents a critical component of this research domain.

Training FMs from scratch in an FL context has demonstrated its potential. However, this approach tends to be less reliable and less effective for models exceeding 100 million parameters. The substantial computation and communication requirements associated with training such larger models significantly constrain the feasibility of traditional full model FL training techniques for FMs. Consequently, the shift from pre-training towards more computationally efficient adaptation methods (e.g., PEFT, prompt tuning, instruction tuning) is promising for overcoming these limitations.

\subsubsection{Parameter-Efficient Fine-Tuning}
\
\par
\indent
PEFT refines FMs by optimizing a minimal subset of model parameters, strategically positioned throughout the model. This approach significantly reduces computational and storage requirements, demonstrating that even large models can be efficiently adapted by tweaking a handful of parameters \cite{ding2023parameter}.

A pioneering PEFT technique, adapter tuning \cite{houlsby2019adaptor}, modifies pre-trained models with minimal parameter adjustments. It integrates specialized adapters with a bottleneck design between model layers. BitFit \cite{zaken2021bitfit} selectively updates only the bias terms within FMs, leaving other components unchanged. 
{Low-Rank Adaptation (LoRA) \cite{hu2021lora}} updates attention weights through low-rank matrices, thereby minimizing the number of parameters required for training. Typically, the proportion of trainable parameters in such methods is around 1\% of the entire FM.
For instance, conventional fine-tuning of a model like GPT-3 necessitates adjusting approximately 175 billion parameters---a task that is untenable for most industry and academic settings. By employing LoRA and focusing solely on low-rank matrices within each transformer layer, the training process involves just 37.7 million parameters. 

Under FL settings, Sun et al. \cite{fedpeft2022} introduced FedPEFT, an innovative approach for fine-tuning FMs. This method freezes the majority of the model weights, focusing on adjusting a minimal subset of parameters tailored to specific downstream tasks. FedPEFT evaluates the efficiency of four distinct PEFT strategies including, Head-tuning, Bias, Adapter and Prompt-in the context of the Vision Transformers (ViT-B) with 85 million parameters. Their findings demonstrate that these targeted fine-tuning methods achieve comparable results to centralized model fine-tuning with non-IID data, while remarkably reducing communication overhead by over 99\%. FedCLIP \cite{clip} also experimented with an Adaptor-based PEFT method using adapter aggregation even based on FedAvg, showing the better performance than standard FedProx and FedAvg. 

{FedPETuning \cite{zhang2023fedpetuning} conducted an extensive benchmark analysis of adapter-based fine-tuning techniques under FL settings. Their findings indicate that both approaches, integrating additional adapters and utilizing LoRA, can achieve comparable accuracy. However, LoRA-based fine-tuning incurs only one-third of the communication overhead required by methods involving extra adapters. 
SLoRA \cite{babakniya2023slora} introduced an optimization of LoRA tailored for non-IID FL settings. It employs a ratio to moderate the influence of updates from individual clients, aiming to counteract the drift caused by heterogeneous data distributions. Although it reportedly matches the performance of full model fine-tuning, it requires a considerable warm-up period to ensure that all clients share a common initialization of LoRA. 
pFedLoRA \cite{pFedLoRA} first introduced LoRA into model-heterogeneous FL. It consists of a lightweight linear low-rank adapter shared by clients with heterogeneous local models for effective knowledge exchange, achieving efficient communication and computation.
SA-FedLoRA \cite{SA-FedLoRA} applied LoRA to fine-tune a large-scale pre-trained model in FL. It freezes the pre-trained model and trains a LoRA adapter with reduced ranks as communication rounds increase, improving model performance while incurring low communication and computational overheads.
FFA-LoRA \cite{FFA-LoRA} utilizes LoRA to fine-tune LLMs. It freezes the pre-trained LLMs and the randomly initializes the non-zero matrix of LoRA. It only fine-tunes the zero-initialized matrix of LoRA to address data heterogeneity, enhance privacy preservation under differential privacy (DP), and improve computational efficiency.
}

{Cross-device FL systems often include devices with diverse resource capacities, leading to differences in model training efficiency \cite{chen2024feddat}. To mitigate this issue, several methods have been proposed to tailor model architectures for resource-heterogeneous FL systems. LoRA-based FedPEFT, for example, offers notable flexibility and adaptability in fine-tuning frozen FMs without overloading client devices. \cite{su2023fedra} suggested assigning LoRA adapters to different numbers of layers for clients based on a randomly generated mask matrix. Alternatively, selecting diverse LoRA ranks according to the system capabilities of clients is another strategy. FlexLoRA \cite{bai2024federated} dynamically adjusts local LoRA ranks. It reconstructs the full-sized LoRA module for server-side aggregation, followed by SVD-based parameter redistribution. However, \cite{cho2023heterogeneous} observed that this reconstruction-redistribution approach results in performance degradation compared to homogeneous LoRA. Consequently, they proposed HETLORA \cite{cho2023heterogeneous} which employs zero-padding to standardize module sizes before aggregation, and then truncates the global LoRA modules to fit the specific rank requirements of subsequent clients.
}

FwdLLM \cite{xu2023fwdllm} is designed to adapt FMs for use on mobile devices with stricter resource constraints. To address this issue, it integrates Backpropagation (BP)-free training with parameter-efficient training approaches. Rather than relying on traditional BP to calculate precise gradients, the BP-free method introduces minor and self-generating perturbations to these ML model parameters. It then evaluates how these perturbations affect model predictions compared to the original unperturbed model. If a perturbation results in predictions that are closer to the actual labels, it is considered to be guiding the model towards the global optimum. Remarkably, FwdLLM demonstrates substantial performance enhancements, achieving convergence three orders of magnitude faster and reducing the memory footprint by close to 15 times.

\subsubsection{Prompt Tuning}
\
\par
\indent
Prompt tuning is a highly effective strategy for adapting pre-trained models to specific downstream tasks. This technique involves appending natural language texts to either the beginning or the end of inputs or outputs. It aims to guide the pre-trained models towards executing particular functions \cite{min2023recentPt, yuan2021bartscore}. Among the various fine-tuning methods, prompt tuning stands out for a variety of benefits, which make them particularly suitable for FL for which computational efficiency and task effectiveness are paramount.

A key significant advantage of prompt tuning is that it often does not require modifications to the FM parameters, thereby markedly lowering computational overhead. Furthermore, in scenarios involving limited training data, skillfully crafted prompts can effectively substitute for hundreds of labeled examples \cite{scao2021many}, demonstrating its potential to enhance model performance with minimal resources. Prompt tuning specifically focuses on optimizing the likelihood of obtaining a desired output by augmenting the original input with trainable embedded prompts. It is important to note that under FL settings, the desired outputs can differ across FL clients. 

FedPrompt \cite{zhao2023fedprompt} introduced an approach that focuses on efficient transmission and aggregation of prompts generated across FL clients. It aims to enhance the performance of a pre-trained model on specific downstream tasks by centrally combining the insights gained from locally generated prompts. Their results highlight the challenges posed by data heterogeneity and the inconsistency of desired outputs across FL clients. It has been observed that such heterogeneity in data distributions and objectives leads to performance decline of 5-10\% compared to models trained in a centralized manner.


\subsubsection{Instruction Tuning}
\
\par
\indent
Instruction tuning is an effective method for improving the functionality and manageability of LLMs \cite{zhang2023instruction}. It involves additional training of FMs using pairs of instructions and the corresponding outputs, where an instruction specifies a task for a model and the output represents the expected result in accordance with the given instruction. The appeal of employing instruction tuning for \methodname{} stems not only from its computational efficiency, which facilitates the quick adaptation of FMs to particular domains. In this way, there is no need for extensive re-training or modifications to the model structure. In addition, it also effectively narrows the divergence between FMs' inherent next-word prediction goals and the actual user intent, aligning with the aim of personalized FL \cite{tan2022towards}. This approach also leads to more consistent and predictable model behaviors. Through instruction-based fine-tuning, models are more closely guided in output generation, ensuring alignment with user expectations and enhancing overall controllability.

FedIT \cite{zhang2023fedit} explores the application of instruction tuning to the LLaMA-7B model under FL settings, catering to a variety of client-specific tasks simultaneously. Similar to methods discussed previously, FedIT also applies LoRA for fine-tuning FMs and leverages FedAvg for aggregating LoRA parameters. A distinctive aspect of FedIT is its utilization of structured instruction-output pairs for fine-tuning FMs. The findings from their study indicate that federated instruction tuning, offers substantial benefits over centralized training approaches. This demonstrates its potential to enhance model performance effectively, even in the presence of task diversity.

\subsection{Communication Efficient \methodname{}}
\label{sec:communication}
The complexity of FMs directly translates to inflated communication overhead between FL clients and the FL server in \methodname{}, hindering efficient collaboration. Although as discussed in Section \ref{sec:computation}, it is not always necessary to transmit an entire FM, fine-tuning billion-parameter FMs by adapter still requires the transmission of millions of parameters. This highlights the need for communication efficient \methodname{} research. In this section, we highlight two primary strategies for enhancing efficiency: \textit{1) model pruning} and \textit{2) model compression}, the readers can refer to Fig. \ref{fig:fig-efficient FedFM}. 

\subsubsection{Model Pruning}
\
\par
\indent
Model pruning aims to identify and retain only the essential parts of  the model for a specific purpose. 
An early model pruning approach is HeteroFL \cite{diao2020heterofl}, which aims to accommodate heterogeneous model architectures within FL by adaptively distributing sub-networks suited to the capacities of individual clients. It selects and aggregates subsets of the global model, pruning of a large model into a variety of smaller more manageable models in effect.
FjORD \cite{horvath2021fjord} implements ordered dropout, a technique that organizes knowledge within a deep neural network in a structured and hierarchical manner. This allows for the extraction of compact sub-models without retraining. Ordered dropout enhances computational efficiency by dropping model components in sequence rather than at random, aligning with the optimization capabilities of contemporary linear algebra libraries. The approach is complemented by a self-distillation process to refine the model further.

Building on these foundations, PruneFL \cite{jiang2022pruneFL} proposed a two-stage model pruning process specifically designed for FL environments. Initially, a ``warm-up'' phase involves selecting a single capable and trusted client to prune the model using its local data, thereby starting the FL process with a streamlined model. In the subsequent ``adaptive pruning'' phase, the server periodically adjusts the model by removing or reintroducing parameters over multiple iterations.
FedPM \cite{isik2022fedpm} adopts a strategy inspired by the lottery ticket hypothesis for model pruning. Rather than utilizing a pruned model as an initial point, it initializes the random binary mask guided by the common seed for all clients. At the end of each round of FL training, clients return their binary masks to the server. The server then constructs a global model by computing a weighted average of these masks.
FedTiny \cite{huang2023fedtiny} follows a similar approach to PruneFL, but with a novel initialization step. It utilizes batch normalization values from clients' data as the basis for selecting a shared initialization, thereby enhancing adaptability to diverse client data distributions.

\subsubsection{Model Compression}
\
\par
\indent
Another approach for improving the communication efficiency of \methodname{} is through model compression. Unlike pruning which typically removes or masks unnecessary parameters without changing model structures, compression involves transforming the model structure or using quantization and coding techniques to represent the model more compactly.\footnote{Some research regards pruning as a model compression method \cite{cheng2017surveycompression}.}

Deep models often operate with full precision (32-bit), but in practice such high degree for computation may be not always necessary. FedPAQ \cite{reisizadeh2020fedpaq} leverages this fact to reduce communication overhead. Note that dynamic quantization only benefits communication. Depending on the infrastructure, the model updates might need to cast back to full precision, thereby incurring additional computational burden. 
SoteriaFL \cite{li2022soteriafl} strikes a balance among privacy, convergence accuracy and communication efficiency. For model compression, it utilizes a straightforward algorithm, CDP-SGD, which effectively integrates communication compression with DP-SGD for enhanced efficiency.

H-FL \cite{yang2021h-fl} addresses the challenges posed by the statistical heterogeneity of client data, which often leads to performance degradation in FL models. A notable aspect of its design is the utilization of lossy singular value decomposition (SVD) \cite{klema1980singular} applied to the feature matrix for model compression. While this technique enables efficient data representation, it compromises model accuracy. To counteract this, H-FL incorporates a bias correction mechanism that is activated prior to each FL training round on the client side. It is designed to adjust the gradients of features affected by the lossy compression. This corrector comprises of multiple fully connected layers, which are sequentially arranged and whose parameters are dynamically updated based on the SVD outcomes of the features extracted from the initial, shallow model layers. 

FedOBD \cite{chen2022fedobd} segments a large model into semantic blocks. It enables FL server and clients to selectively exchange quantized blocks. This strategy assesses the importance of blocks over individual parameters, enabling the selective omission of less critical blocks to significantly reduce communication overhead, while preserving model performance. In addition, FedOBD incorporates the advanced Adaptive Deterministic Quantization for Neural Networks (NNADQ) to further improve communication efficiency. Results show that it can achieve a two-fold reduction in communication costs compared to FedPAQ, highlighting its potential in improving communication efficiency for \methodname{}.

\section{Towards Trustworthy \methodname{}}
\label{sec:trustworthy}

\begin{figure*}[!t]
\centering
\includegraphics[width=\linewidth]{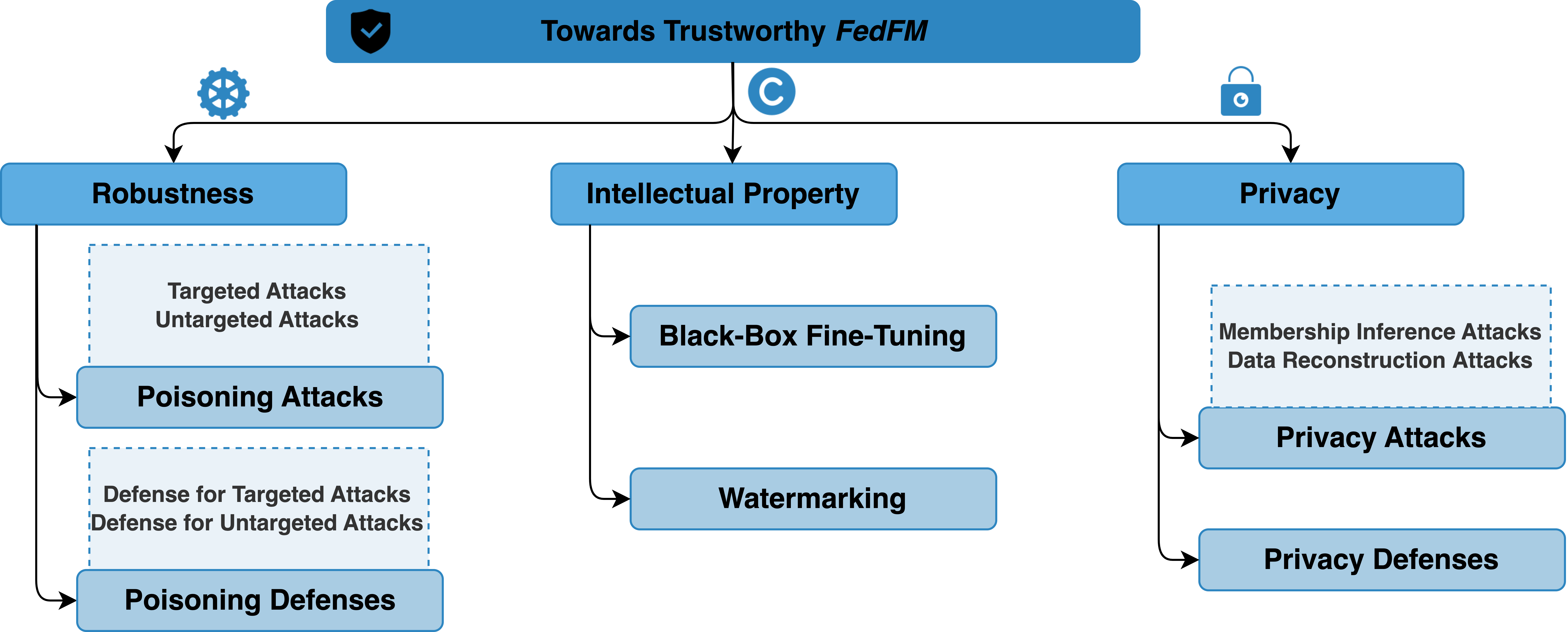}
\caption{Trustworthy \methodname{} research works can be divided into three domains: 1) Robustness, 2) Privacy, and 3) Intellectual Property Protection, each of which is further divided into specific strategies that address the challenges of \methodname{}.}
\vspace{-6pt}
\label{fig:trustworthy}
\end{figure*}

Attacks and defenses on traditional FL processes have been extensively studied \cite{Lyu-et-al:2022TNNLS}. In this section, we review selected studies on robustness and privacy issues in FL, and point out the need for intellectual property protection in \methodname{}. We further break down each domain into specific areas as shown in Fig. \ref{fig:trustworthy}.

\subsection{Robustness of \methodname{}}
\label{sec:Byzantine-Robustness}
There are numerous studies on robust FL, among which the robustness against poisoning attacks launched by Byzantine clients is the main focus. Diverse poisoning attacks that exploit vulnerabilities in FL and various Byzantine-robust FL schemes have been proposed.

\subsubsection{Poisoning Attacks}
\
\par
\begin{figure*}[t]
    \centering
    \includegraphics[width=0.95\linewidth]{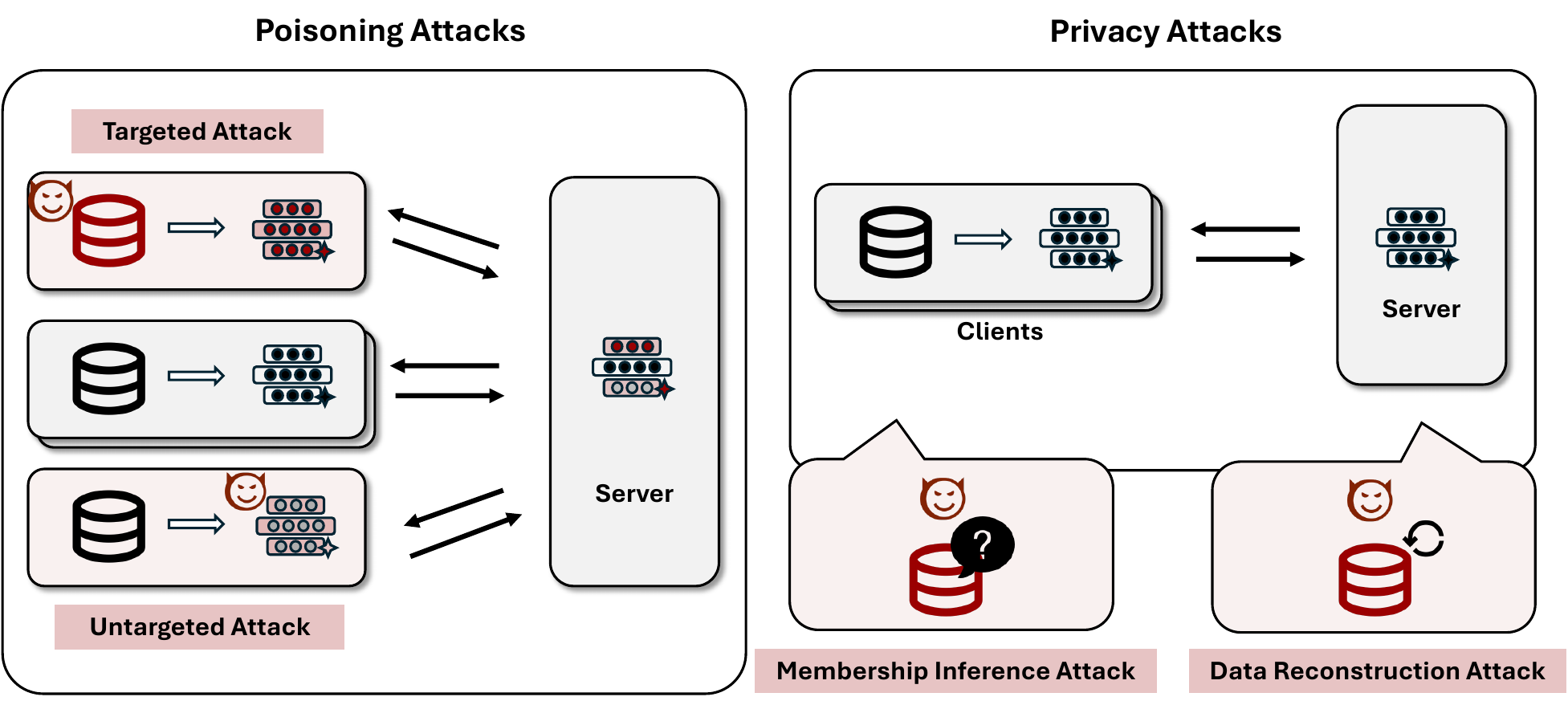}
\caption{Illustration of poisoning and privacy attacks on \methodname{}.}
    \label{fig:trust_attacks}
    \vspace{-6pt}
\end{figure*}

\indent
FL poisoning attacks aim to compromise either the global model or the training process. They can be divided into two categories: 1) targeted attacks and 2) untargeted attacks, based on whether the attacks aim to manipulate model outputs or disrupt global model convergence (Fig. \ref{fig:trust_attacks}).

\textbf{Targeted Attacks}: They are usually utilized by attackers to manipulate the specific global model outputs, while maintaining its benign performance. During the training phase, the attackers inject crafted poisoned samples into the training dataset, which can perturb the decision boundaries of the model within a small sub-space, resulting in miss-classifications. These perturbations usually depend on the victim model's overfitting the trigger patterns on benign inputs. Various trigger generation schemes \cite{BadNets:IdentifyingVulnerabilitiesintheMachineLearningModel, DBLP:journals/compsec/HeSXHTZ24, DBLP:journals/www/HouHYKT23} have been proposed to make attacks stealthier. Targeted attacks on FL \cite{DBLP:conf/aistats/BagdasaryanVHES20, DBLP:conf/aaai/LyuHWLWL023} have been proposed to elevate the survivability and utility of the attacks.

\textbf{Untargeted Attacks}: They are often launched by attackers to prevent the FL model from achieving convergence. Specifically, the attackers could craft and submit local models that introduce significantly high variances if aggregated into the global model, perturbing and even blocking it from being optimized towards the global optimum. An example is the Gaussian attack, in which poisoned model parameters are randomly sampled from a Gaussian distribution. Various advanced poisoning sample generation algorithms \cite{PoisoningAttacksagainstSupportVectorMachines, TargetedBackdoorAttacksonDeepLearning, PoisoningAttackstoGraph-BasedRecommenderSystems, ManipulatingMachineLearning:Poisoning, PoisonFrogs!TargetedClean-LabelPoisoning} usually aim to craft samples in training datasets to make the models achieve maximum validation losses.

Under \methodname{} settings, the reliance on contextual inputs (i.e., in-context learning) makes the prompts given by the trainer play a significant role in FM training. This can be a potential vulnerability to be exploited by targeted attacks. Various prompt-based targeted poisoning attacks against language models \cite{DBLP:conf/acl/MeiLWZM23,DBLP:conf/emnlp/ZhaoWLZF23,DBLP:conf/ijcai/DuZLLW22} have been proposed, which could also threaten the integrity of \methodname{}.

\subsubsection{Poisoning Defenses}
\
\par
\indent
\textbf{Defense for Targeted Attacks}: Defending against targeted attacks is more challenging under \methodname{} settings. Specifically, due to the high complexity of FM training tasks, the poisoned input that triggers the FM misbehavior is more stealthy. Existing backdoor detection approaches \cite{DBLP:conf/sp/WangYSLVZZ19,DBLP:conf/cvpr/0002T0SXL0M023} mainly rely on generating cross-category sample transfer shortcuts via optimization, given the model and total number of categories. However, the ML tasks for FMs introduce a enormously high level of complexity (e.g., millions of image classification categories), making existing defense mechanism against targeted FL attacks intractable.

\textbf{Defense for Untargeted Attacks}: However, achieving the objective of such attacks can be challenging in \methodname{}, due to the large sizes and high local heterogeneity of FM training tasks. Firstly, to achieve validation loss maximization, an attacker needs to repeatedly train a local shadow model to form a differentiable poisoning sample optimization objective. This is infeasible in \methodname{} considering the huge costs introduced by FMs. Secondly, the local ML tasks can be heterogeneous, making it hard for such attacks to compromise global FM convergence or performance.

To defend against FL poisoning attacks, the FL server usually adopts Byzantine-robust aggregation rules to ensure poisoned local model updates are excluded from model aggregation. Such aggregation rules can be divided into three categories: 1) geometrical outlier detection, 2) top performance selection, and 3) other hybrid schemes.

\underline{Geometrical Outlier Detection}: These schemes discard local model updates which are regarded as geometric outliers, and only aggregates the remaining ones to form the global model. Some approaches \cite{Machine_Learning_with_Adversaries} calculate the Euclidian distances among local model updates to determine the divergences among them, and remove those with the highest degree of heterogeneity from the majority. Others \cite{Byzantine-RobustDistributedLearning:Towards} analyze the parameter-wise geometrical divergence, removing the smallest and largest values of the same parameter of all model updates, and only aggregating the remainders to produce the final corresponding parameter in the global model.

However, these schemes are not compatible with \methodname{}. Specifically, due to the significant multi-modality of local models and high levels of heterogeneity among local data distributions, there might be significant natural geometric divergence among benign local model updates, sometimes even making them geometrically incomparable. Besides, the huge scales of FMs also make the calculation of divergence computationally intensive. Hence, for \methodname{}, existing defenses against FL poisoning attacks from Byzantine clients might not be effective.

\underline{Top Performance Selection}: These schemes require a clean validation dataset to be stored by the central FL server to evaluate each local model update from the clients, selecting and aggregating the ones with top performance. Some schemes \cite{DBLP:conf/icml/XieKG19, DBLP:conf/icml/XieKG20} choose to aggregate the local model updates that contribute the largest validation loss reduction. Others \cite{LocalModelPoisoningAttacksto}
choose to aggregate the local model updates that contribute the largest accuracy improvement.

However, these schemes might not be feasible under \methodname{} settings. Specifically, due to the significant heterogeneity of local training tasks, benign local model updates might also show sub-optimal perform on the validation dataset. It can be challenging to construct a validation dataset due to the currently unclear evaluation metrics for FM performance. The huge scales of FMs and the corresponding datasets also make effective evaluation of local model updates costly.

\underline{Hybrid Schemes}: Research works have emerged \cite{DBLP:conf/acsac/HaoLXC021, DBLP:conf/ndss/CaoF0G21, prakash2020mitigating} in an attempt to form hybrids schemes to take advantages of the aforementioned by combining them together. These works often update a benchmark model with a central clean dataset, and lower the weights assigned to local model updates that significantly deviate from the benchmark (or even discard them completely) during aggregation. Although these methods have achieved superior robustness, they have also inherited the limitations from the aforementioned categories of approaches which make them incompatible with the \methodname{} settings. 

\subsection{Privacy of \methodname{}}
\label{sec:privacy}
Privacy preservation has always been a central focus of FL research. A wide range of privacy attacks have been studied and potential defenses have been proposed. 

\subsubsection{Privacy Attacks}
\
\par
\indent
Privacy attacks in FL are generally designed to access the victim client's private information given based on auxiliary information essential for FL model training (e.g., the victim's local model updates or gradients). They can be divided into two categories: 1) membership inference attacks, which attempt to infer the involvement of a specific data sample in FL training, and 2) data reconstruction attacks, which attempt to recover a victim client's original training data (Fig. \ref{fig:trust_attacks}).

\textbf{Membership Inference Attacks}: They aim to infer whether a specific data sample is in a victim client's local training dataset, which can be used to infer about the victim's private information such as identity for instance. Specifically, with a victim client's model (white box) or a trained shadow model that imitates it (black box), the attacker could train an attack model that infers whether a data sample belongs to the victim's local dataset. 
Schemes like \cite{shokri2017membership, DBLP:journals/tdsc/LiuWLPW23, DBLP:journals/tdsc/YanLWZSHL23} under black box settings usually train several shadow models with datasets partitioned in a variety of ways into training and validation sets. Then, correspondingly labeled as \textit{in} (indicating used for training) and \textit{out} (indicating not used for training), training and validation data, together with their predictions/logits, form the training dataset for the attack models. 
Recent studies have proven that such attacks can work on generative models (e.g., diffusion models \cite{DBLP:conf/icml/DuanK0SX23}), demonstrating the potential risks of membership inference attacks on \methodname{}.

However, the reliance on the original victim or shadow model makes this category of attacks difficult to implement under \methodname{} settings. The significant scale of FMs and the highly restricted external access to FMs (black box nature) make existing membership inference attacks designed for FL intractable for \methodname{}. Nevertheless, recent studies \cite{DBLP:journals/corr/abs-2311-06062} have proposed membership inference attacks which do not rely on a shadow model against prompt-based LLMs. Therefore, the threat still exists.

\textbf{Data Reconstruction Attacks}: These attacks aim to reconstruct the actual training data used from a victim client's model. Through various optimization methods (e.g., model inversion, gradient matching, adversarial training), an attacker can generate data samples that are close to the original ones in the victim's training dataset. Schemes based on diverse techniques (e.g., gradient matching \cite{DBLP:conf/nips/ZhuLH19, DBLP:journals/corr/abs-2001-02610}, GAN \cite{DBLP:conf/aaai/YuanCZ0YZ23}) have been developed and demonstrated to be effective under the traditional FL settings.

Due to the large sizes of FMs, data reconstruction attacks on \methodname{} can incur significantly costs. However, recent studies show that in various FM training scenarios, the model's superior ability of information representation and frequent model-client interactions can lead to privacy leakage, making data reconstruction attacks feasible. With specifically crafted prompts, FMs could generate sensitive feedback \cite{DBLP:journals/corr/abs-2311-09127, DBLP:journals/corr/abs-2310-02446, DBLP:journals/corr/abs-2307-08715}, from which the attackers can gain access to private information about the training datasets.

\subsubsection{Privacy Defenses}
\
\par
\indent
A wide range of defenses against privacy attacks in traditional FL settings have been proposed to preserve clients' data privacy. The mainstream schemes are generally designed to make trade-offs between knowledge integrity and privacy guarantees. Specifically, through model perturbation or compression techniques (e.g., differential privacy, confidence masking, model compression/sparsification), clients can reduce the risk of exposure of local private information through model parameters, while avoiding significantly negative impact on model performance. 

By adding statistical noises to the shared local model updates, differential privacy techniques can provide guarantees on privacy preservation to different extents \cite{DBLP:journals/tifs/XueXZLZSL24,DBLP:journals/jsac/OkegbileCZCY23,DBLP:journals/tdsc/LinWLSHD23}. The trade-off between privacy preservation and model performance has been widely studied, with Game Theory often being leveraged to constrain the variance of the noises added to the model while compensating the privacy risks of clients through incentive mechanisms \cite{sun2021pain,sun2022profit}.
The difficulty of recovering private local data could be further enhanced by incorporating mechanisms of gradient/model compression and sparsification \cite{hu2020federated, hu2023federated, wang2021datalens, chen2024privacy}. Due to the tension between model performance and degree of compression, the trade-off between performance and privacy is also an important topic of study \cite{chen2024privacy, zhang2023trading}, aiming to find potentially optimal solutions of privacy guarantee with limited impact on performance.

Due to their simplicity and flexibility, the approaches mentioned above are compatible with FL schemes with various modalities and scales. Hence, they could be promising solutions to addressing privacy issues under \methodname{} settings.

\begin{figure}[t]
    \centering
    \includegraphics[width=1.0\linewidth]{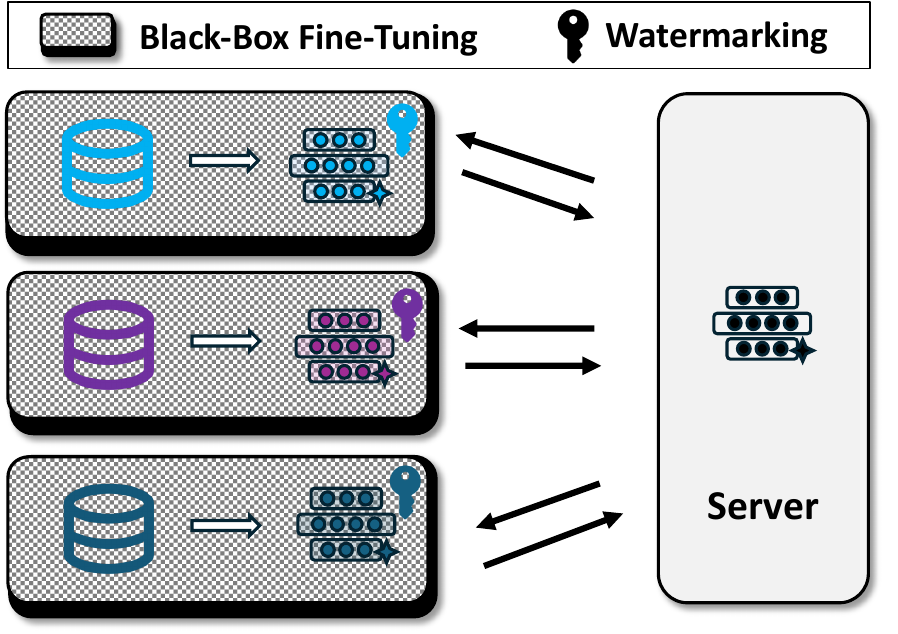}
\caption{Illustration of intellectual property protection strategies for \methodname{}.}
    \label{fig:IP}
    \vspace{-6pt}
\end{figure}

\subsection{\methodname{} Intellectual Property Protection}
{
Intellectual property protection is crucial for safeguarding \methodname{} against unauthorized use (e.g., model theft \cite{tekgul2021waffle}). In the context of \methodname{}, we explore two main intellectual property protection strategies: 1) black-box fine-tuning, and 2) watermarking (Fig. \ref{fig:IP}). By integrating these intellectual property protection strategies, \methodname{} can ensure the secure and efficient use of FMs across distributed networks, making \methodname{} more robust and secure for collaborative AI model development

\textbf{Black-Box Fine-Tuning}: Black-box fine-tuning refers to methods that fine-tune FMs without accessing the model's internal parameters \cite{sun2022black}. This approach is particularly relevant in FL, where maintaining the integrity and ownership of the original models is crucial. Black-box fine-tuning methods often add new parameters while keeping the original model unchanged, thus preserving the original model’s intellectual property. For example, Fed-BBPT \cite{lin2023efficient} is a prompt tuning framework that enables the joint training of a lightweight prompt generator across multiple clients. This framework allows clients to fine-tune their models locally without accessing or altering the core model parameters. Similarly, the method in \cite{sun2024fedbpt}, called FedBPT, uses an evolutionary algorithm to train optimal prompts that enhance the performance of frozen FMs. These black-box fine-tuning methods ensure that local fine-tuning of FMs can be achieved with robust intellectual property protection. However, current research mainly focuses on few-shot learning with small datasets for FM fine-tuning, indicating a need for further exploration with larger datasets and multiple data types to fully leverage the potential of black-box fine-tuning for \methodname{}.

\textbf{Watermarking}: Watermarking is a widely-adopted technique for intellectual property protection by embedding identifiers into models to prove ownership. In FL environments, \cite{tekgul2021waffle} introduced WAFFLE, a solution for embedding watermarks into global models. This technique ensures that the ownership of the model can be verified even when the model is distributed across multiple clients. More recently, \cite{yu2023leaked} developed DUW, which embeds a unique key into each client's local model. This key can identify the source of a leaked model and verify ownership, thereby providing a robust mechanism for intellectual property protection in federated settings. The use of watermarking in \methodname{} not only can secure the global model, but also can protect individual client models, thereby ensuring comprehensive intellectual property protection across the \methodname{} network.
}

\section{Towards Incentive Mechanisms for \methodname{}}
\label{sec:incentive}

The remarkable success of FL depends on clients (a.k.a., data owners) actively engaging during the training process. In reality, data owners might hesitate to join FL without proper compensation, especially if it involves significant commitment of local resources for training FMs. Thus, developing robust incentive mechanisms is imperative to motivating clients to participate in the \methodname{} training process, while deterring misbehaviours through punitive measures. Table \ref{tab:incentive_summary} provides an overview of existing incentive mechanisms in FL.

\subsection{FL Incentive Mechanisms}
Contract theory, game theory and auction mechanism are three widely adopted techniques in FL incentive mechanism design. Thus, we review and discuss existing works based on these three categories. 

\begin{table*}[!t]
  \centering
  \caption{Summary of the Main Categories of Incentive Mechanisms in FL}
  \resizebox*{2\columnwidth}{!}{
\begin{tabular}{p{1.8cm}|p{4.4cm}|p{4.6cm}|p{3.2cm}|p{3.6cm}|p{1.6cm}}
\hline
 & {\makecell[c]{\textsc{General} \\ \textsc{Objectives}} } 
  & {\makecell[c]{\textsc{Application}  \\ \textsc{Scenarios}}} 
  & {\makecell[c]{\textsc{Main}  \\ \textsc{Advantages}}}
  & {\makecell[c]{\textsc{Main}  \\ \textsc{Disadvantages}}}
  & {\makecell[c]{\textsc{Related}  \\ \textsc{Works}}} \\\hline
Contract theory & Maximizing the clients' key performance indicators & Scenarios with complete information asymmetry & Efficient resource-based reward & Reliance on traditional optimization methods  & \multicolumn{1}{c}{\multirow{2}{*}{\begin{tabular}[c]{@{}l@{}}\cite{kang2019incentive,kang2019toward,hou2017incentive,liu2017design,ding2020incentive,sarikaya2019motivating, feng2019joint,lim2021towards}\end{tabular}}}\\\hline
Game theory & Finding equilibrium solutions among participants to achieve utility maximization & Scenarios where interactions among participants are complex & Efficient handling of multifaceted goals & Prolonging training time &  
\multicolumn{1}{c}{\multirow{2}{*}{\begin{tabular}[c]{@{}l@{}}\cite{feng2019joint,sarikaya2019motivating,pandey2020crowdsourcing,dinh2020federated,khan2020federated,hu2020trading,lee2020market,sarikaya2020regulating,qu2020privacy,song2019profit,yu2020sustainable}\end{tabular}}}
\\\hline

Auction mechanism & Maximizing the social welfare & Scenarios with high competition & Fairness \& efficiency & May lead to dishonest behaviors & 
\multicolumn{1}{c}{\multirow{2}{*}{\begin{tabular}[c]{@{}l@{}}\cite{li2019credit, krishnaraj2022future, zavodovski2019decloud, hong2020optimizing,bahreini2018envy, gao2019auction, jiao2018social, jiao2019auction, yang2020task,jiao2020toward,zeng2020fmore,ying2020double,le2020auction,le2021incentive,zhang2021incentive,roy2021distributed,deng2021fair,zhang2022auction,zhang2022online,tang2023utility,tang2023competitive,tang2023multi}\end{tabular}}}

\\\hline

\end{tabular}
}
\label{tab:incentive_summary}
\vspace{-3pt}
\end{table*}

\subsubsection{Contract Theory-based Methods}
\
\par
\indent
To address the information asymmetry issue, \cite{kang2019incentive} devised an incentive mechanism categorizing FL participants based on data quality and compute resources, offering rewards based on contributions. FL participants select contracts to maximize profits, facing penalties for failures to meet the terms. This approach attracts high-quality data owners, enhancing FL performance and optimizing incentive payouts. Compared to Stackelberg game-based methods \cite{kang2019toward,hou2017incentive,liu2017design}, it is more adaptable to asymmetric information.

However, \cite{ding2020incentive} noted limitations in existing techniques \cite{kang2019incentive, sarikaya2019motivating, feng2019joint}, which only allowed the FL server to make decisions based on a single dimension of consideration. They proposed a two-dimensional incentive scheme considering training costs and communication delays, dealing with incomplete information regarding heterogeneous device networks. FL clients offering specific training data sizes and timely updates are selected, with penalties for non-compliance. Their method can effectively deal with a weak level of information asymmetry. However, strong information asymmetry challenges server decision-making, potentially leading to incentive mismatches.


\subsubsection{Game Theory-based Methods}
\
\par
\indent
\textbf{Stackelberg Game-based Methods}:
In FL, direct communication between the server and clients for exchanging model parameters has been identified as a source of inefficiency. To tackle this issue, \cite{feng2019joint} proposed a relay network to construct a communication platform, introducing a Stackelberg game to analyze the interaction between clients and the server. This game involves decisions about transmission power and relay node selection due to wireless network interference.
In \cite{sarikaya2019motivating}, the emphasis is placed on fair treatment of clients, recognizing potential selfish behaviours. They addressed challenges in synchronous batch tasks, introducing an incentive scheme to reduce time delay through a Stackelberg game.
Another contribution by \cite{pandey2020crowdsourcing} focused on enhancing communication efficiency by a crowdsourcing framework. They employed a two-stage Stackelberg game to model the interaction between mobile edge computing (MEC) servers and clients. Rewards are determined based on local model accuracy.
Building on these works, \cite{dinh2020federated,khan2020federated} extended Stackelberg game-based FL incentive mechanisms to model interactions in edge networks. The server acts as a leader offering a reward, motivating clients to perform more FL training rounds to improve model accuracy.
\cite{hu2020trading} criticized existing incentive schemes for privacy burdens on the FL server. They proposed a two-stage Stackelberg game to ensure privacy with a specialized budget. Users can strive towards optimizing utility under a given privacy budget.
In \cite{lee2020market}, a distributed market model involving IoT devices, MEC operators and a cloud operator was introduced. It leverages Stackelberg game to optimize revenue and energy consumption.
These studies highlight the importance of incentive design, privacy preservation, and efficient communication in FL, demonstrating the potential of Stackelberg games in optimizing server-client interactions.

\textbf{Yardstick Competition-based Methods}:
Stackelberg game-based approaches tend to be time-consuming \cite{sarikaya2020regulating}. To address this, \cite{sarikaya2020regulating} introduced a novel yardstick competition scheme, aiming to reduce training time in synchronous stochastic gradient descent. The yardstick serves as a benchmark, calculating acceptable delay with rewards being determined based on client deviations from it. However, this approach estimates delay solely based on CPU power without considering communications related factors.

\textbf{Shapley Value-based Methods}:
In \cite{qu2020privacy}, a two-phase framework incentivizing edge servers in a collaborative cloud-edge-device setting was proposed. It leverages Shapley Value (SV) to distribute rewards based on edge server contributions. However, it does not incentivize data-contributing devices. SV-based methods \cite{liu2020fedcoin} can assess contributions by FL clients, but often incur high computation costs. To address this issue, \cite{song2019profit} introduced the Contribution Index (CI) based on SVs. It is effective but only applicable to horizontal FL settings.

\textbf{Temporal Incentivization}: 
In \cite{yu2020sustainable}, the Federated Learning Incentivizer (FLI) was designed to address delays in compensating FL contributors. FLI is a real-time payment algorithm. It emphasizes fair treatment and efficient budget allocation. Compared with contemporary schemes, FLI is able to deal with practical FL scenarios in which the revenue generated by the FL model is gradually received and disseminated among contributors in a post-hoc fashion.

\begin{figure*}[t]
    \centering
    \includegraphics[width=\linewidth]{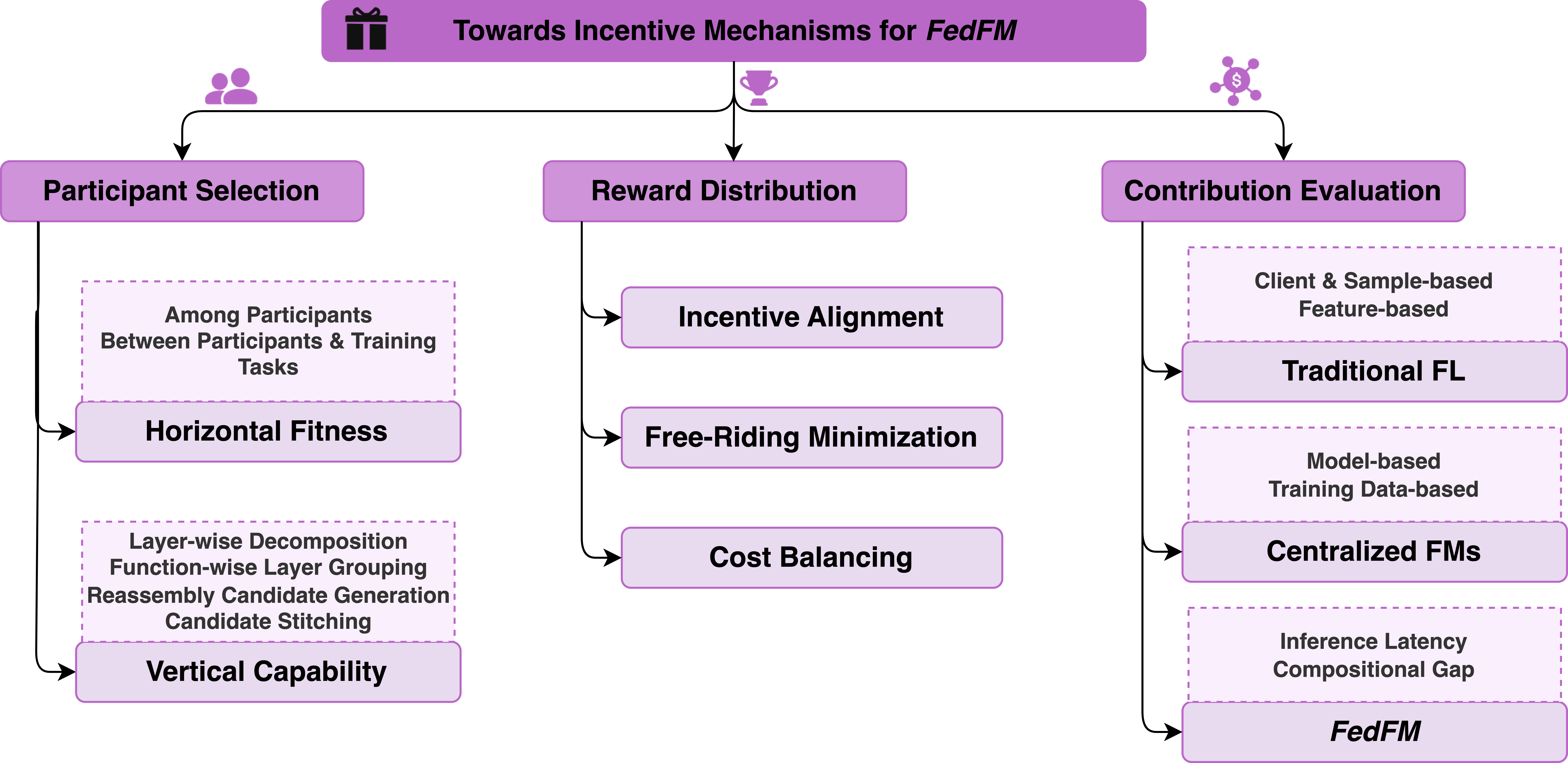}
    \caption{\methodname{} incentive mechanisms can be divided into three domains: 1) Participant Selection, 2) Reward Distribution, and 3) Contribution Evaluation, each of which is further divided into specific strategies that address the challenges of \methodname{}.}
    \label{fig:incentive-FedFMs}
    \vspace{-3pt}
\end{figure*}

\subsubsection{Auction-based Incentivization}
\
\par
\indent
\textbf{Reverse Auction-based Methods}:
Approaches based on reverse auction aim to help the server select clients in a monopoly market to maximize its utility \cite{jiao2020toward,zeng2020fmore,ying2020double,le2020auction,le2021incentive,zhang2021incentive,roy2021distributed,deng2021fair,zhang2022auction,zhang2022online}. They often leverage techniques such as reputation, blockchain, deep reinforcement learning, and graph neural networks. Moreover, they are typically tailored for monopoly markets, where there is only one server (i.e., data consumer) and multiple clients. For example, RRAFL \cite{zhang2021incentive} incorporates reputation and blockchain into a reverse auction. The data consumer publicizes its FL task, and clients bid for it unde RRAFL. The data consumer then determines the winning clients based on their reputation values, which are derived from their data quality and reliability track records in a blockchain.

\textbf{Forward Auction-based Methods}:
Methods in this category study how multiple data consumers shall bid for the same pool of clients to maximize their utility \cite{tang2023utility,tang2023competitive,tang2023multi}. In \cite{tang2023utility}, an optimal bidding function was proposed for data consumers, considering not only their limited budgets and the suitability of clients, but also prior auction-related knowledge such as the distribution of clients and the probability of winning the ongoing auction. It demonstrates that the estimation of client utility and the appropriate winning function significantly impact the optimal bidding strategy.

\textbf{Double Auction-based Methods}: FL incentive mechanisms based on double auction generally aim for social welfare maximization, social cost minimization or maximizing all servers or clients utility \cite{li2019credit, krishnaraj2022future, zavodovski2019decloud, hong2020optimizing}. They facilitate optimal client-server matching and pricing, and are applicable in cases where there are multiple FL servers and multiple FL clients. For example, in \cite{krishnaraj2022future}, an iterative double auction-based method for computing resource trade was proposed to achieve social welfare maximization. The method alternates between optimizing three objectives, while adhering to pricing rules to determine the winners and pricing.

\textbf{Combinatorial Auction-based Methods}:
Similar to those based on double auction mechanisms, methods based on combinatorial auction mechanisms also aim to maximize social welfare, minimize social cost and maximize the utility of all involved FL servers or clients \cite{bahreini2018envy, gao2019auction, jiao2018social, jiao2019auction, yang2020task}. They are suitable for scenarios in which FL clients sell resources in the form of packages, and FL servers compete for these packages. For instance, in \cite{yang2020task}, a multi-round sequential combinatorial auction model was adopted to allocate clients with limited resources to servers with heterogeneous resource requirements. In this approach, servers sequentially publicize their resource requirements and bidding values for different clients. The client-server matching and payments are then optimized.

\subsection{Discussions on \methodname{} Incentive Mechanisms}
It is worth noting that the current incentive mechanisms primarily emphasize participant selection based on a data-centric approach in traditional FL settings. Specifically, the FL server determines participant recruitment by assessing the characteristics of the data held by the candidates. Nevertheless, given the expansive scale and nuanced execution specifics of \methodname{}, it might be beneficial for \methodname{} to adopt a model-centric approach for participant selection, selecting participants based on the characteristics of their models.  
In the next section, we envision promising approaches for designing effective incentive mechanisms for \methodname{} from three interconnected and critical directions (as shown in Fig. \ref{fig:incentive-FedFMs}): \textit{1) participants selection}, \textit{2) contribution evaluation}, and \textit{3) reward distribution}.

\subsection{\methodname{} Participant Selection}
\label{sec:participant selection}
\begin{figure*}[t]
    \centering
    \includegraphics[width=0.95\linewidth]{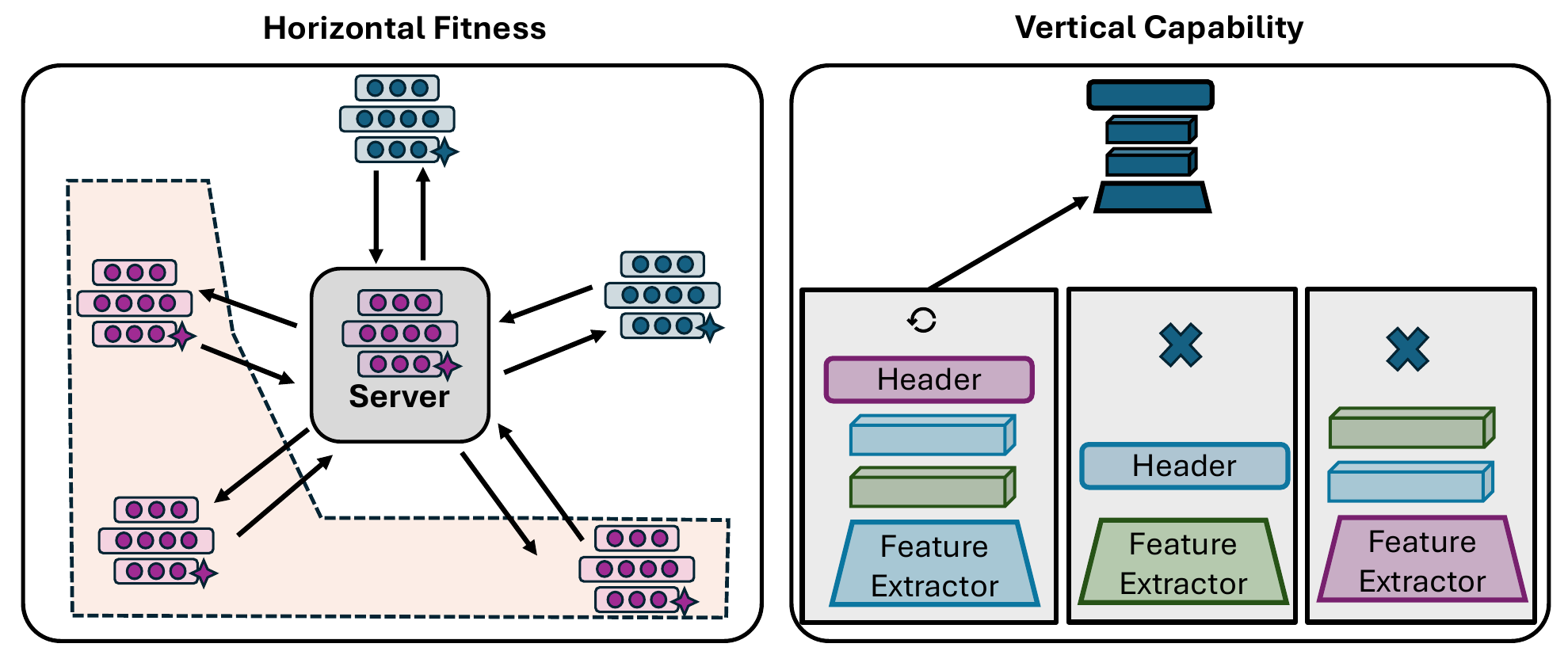}
\caption{Illustration of participant selection in \methodname{}.}
    \label{fig:participant_selection-FedFMs}
    \vspace{-6pt}
\end{figure*}

Fig. \ref{fig:participant_selection-FedFMs} summarizes \methodname{} participants selection. 
The model selection process in \methodname{} encompasses two critical dimensions: \textit{1) horizontal fitness for task-specific training} and \textit{2) vertical capability for training}.

\subsubsection{Horizontal Fitness for Task-Specific Training}
\
\par
\indent
The horizontal dimension, addressing the fitness of task-specific training, entails the selection of a model that meets the training requirements for various tasks. Given the huge scale of typical FMs, direct training from scratch can be challenging. Consequently, a strategy of pre-training followed by fine-tuning is commonly adopted. Typically, techniques such as LoRA \cite{hu2021lora} are applied for fine-tuning the pre-trained model to align with downstream tasks, which might turn out to be diverse (e.g., NLP, computer vision). In this context, the pre-trained FM should be robust and flexible enough to adapt to various downstream tasks. To achieve this goal, it is necessary to select a mixture of FMs hosted by FL participants based on their task training abilities.
To this end, it is crucial to gain insight into the relationships among participants as well as the relationships between participants and the training tasks. 

\textbf{Relationships among Participants}:
There are two types of relationships among participants: 1) the substitute relationship and 2) the complementary relationship. In the context of \methodname{}, a substitute relationship between two participants implies that the model or contribution of one participant can be substituted by those of another, without significantly affecting the overall performance or objective of \methodname{}. It suggests a degree of similarity or equivalence in the roles of the two participants in \methodname{}. The complementary relationship among participants in \methodname{} refers to a situation where the contributions, capabilities or characteristics of different participants complement each other, thereby enhancing the overall performance or effectiveness of \methodname{}.

To effectively manage various participants based on their relationships, clustering can adopted, drawing inspiration from their successful application in conventional FL \cite{ghosh2020efficient}. In particular, according to the relationship among participants, they can be grouped into clusters, with substitute participants assigned to the same cluster while complementary participants assigned to different clusters. Then, in the participation selection process, multi-agent mechanisms \cite{tang2023competitive} could be leveraged with each agent assigned to help manage a cluster, in order to achieve the target performance goals for \methodname{}.

\textbf{Relationships between Participants and Training Tasks}:
With FMs pre-trained on publicly available datasets with general knowledge, the significance of uniqueness of local data and domain-specific knowledge of each \methodname{} participant becomes important for enhancing domain-specific FM performance. 
To effectively manage the relationships among participants and training tasks, a possible approach is to adopt a combination of relationship networks, reputation systems, blockchains, and graph neural networks:
\begin{itemize}
    \item \textbf{Graph Neural Networks (GNNs)} \cite{liu2022federated}: GNNs offer powerful means to model the complex interconnections among participants, their local data, domain expertise, and the target tasks or domains. Through encapsulating these intricate relationships, the network can discern the relevance and synergies among participants' contributions to particular domains or tasks. Leveraging this insight, strategies for knowledge transfer, model personalization, and resource allocation can be orchestrated, thereby maximizing the utilization of participants' distinct data and expertise to boost domain-specific \methodname{} performance.
    \item \textbf{Reputation Systems} \cite{tang2023competitive,zhang2021incentive}: A reputation system can incentivize participants to contribute high-quality, domain-specific data and knowledge to the \methodname{} training process. Participants can earn higher reputations by consistently providing valuable contributions, which, in turn, can grant them increased training resources, greater influence over global model updates, or preferential access to personalized domain-specific models. The tracking of participants' reputation can be facilitated by blockchains \cite{zhang2021incentive}, which support transparency, immutability and decentralization. This encourages participants to curate and share their most relevant and unique local data, ultimately benefiting the performance of domain-specific \methodname{}.
\end{itemize}

\subsubsection{Vertical Capability for Training}
\
\par
\indent
The vertical dimension, focused on the ability to train, entails selecting a mixture of model components that can be assembled into a coherent FM. A large-scale FM can be divided into various components based on the role of each of them. These components can be combined together through techniques akin to conventional FL. For instance, in heterogeneous FL, the local model of each participant can be segmented into two primary parts: the feature extractor and the classifier \cite{FedGH:2023}. The feature extractor maps the input data into the latent space representation, while the classifier translates these representations into output logits. 

In \cite{wang2023fedlego,wang2023towards}, FedLEGO was proposed to address model heterogeneity issues in conventional FL. It treats participants' local models as a LEGO toy, disassembling them into bricks by layers. Subsequently, FedLEGO reassembles these bricks into new model structures. Drawing inspiration from heterogeneous FL, we envision that FM construction can be performed similar to LEGO, where it can consist of various functional building blocks hosted by diverse \methodname{} participants to be assembled together. 
We envision a multi-step approach for \methodname{} training through layer-wise decomposition, function-wise layer grouping, reassembly candidate generation, and candidate stitching.


\begin{itemize}
    \item \textbf{Layer-wise Decomposition}: Initially, the \methodname{} is disassembled into layers, focusing on the operation type of each model component. This step aims to identify the layers and their corresponding operation types within the FMs.
    \item \textbf{Function-wise Layer Grouping}: Subsequently, in the function-driven layer grouping step, these layers are grouped based on their functional similarities. Utilizing K-means style algorithms, these layers can be clustered together based on their functional attributes.
    \item \textbf{Reassembly Candidate Generation}: The reassembly candidate generation step involves assembling various functional candidates by combining the learned layer groups based on their functions. This step results in multiple functional candidates with diverse configurations.
    \item \textbf{Candidate Stitching}: Finally, the functional candidates are stitched together to form various FMs tailored to specific downstream tasks. This approach enhances the adaptability and flexibility of \methodname{}, allowing for the construction of task-specific FMs by reassembling learned functional components.
\end{itemize}

This envisioned modular approach can enhance the flexibility and adaptability of \methodname{}. It allows for the construction of diverse model configurations tailored to specific tasks or requirements. This Lego-inspired approach can be promising in dealing with model heterogeneity, promoting collaborative learning in \methodname{} settings.

\subsection{\methodname{} Contribution Evaluation}
\label{sec:contribution evaluation}
\label{sec:evaluation}

Participant contribution evaluation is crucial to the success of \methodname{} for the following reasons. 
Firstly, the intricate non-linearity and complexity of \methodname{} pose a challenge for stakeholders to understand the internal working mechanisms and decision making processes.
This lack of transparency can diminish trust and impede adoption, especially in safety-critical sectors like finance and healthcare \cite{li2023towards}. 
Secondly, the evaluation of \methodname{} contributes to an enhanced understanding of their strengths and weaknesses. For instance, the PromptBench benchmark \cite{zhu2023promptbench}  demonstrated that existing FMs are sensitive to adversarial prompts, thereby emphasizing the importance of prompt engineering for better model performance.

Contribution evaluation in \methodname{} is more challenging compared to the current approaches for centralized FMs or traditional FL. Firstly, different from centralized FM evaluations, \methodname{} cannot directly assess the quality of the training data used due to privacy protection requirements.
Secondly, \methodname{} evaluation is limited by the resource constraints in terms of communication power and local computation of the FL clients involved. 
Thirdly, compared to traditional FL, \methodname{} evaluation can involve a large number of sequences, resulting in insufficient test coverage \cite{kuchnik2023validating}. 

There are a number of existing evaluation methods for traditional FL and centralized FMs, but few for \methodname{}. In this section, we discuss existing evaluation methods and envision promising research directions for designing new \methodname{} evaluation protocols. 
We present approaches which cover relevant works to evaluate the model performance, analysis the decision results, and provide insights into the contributions of individual participants and datasets. 

\begin{table*}[ht!]
\centering
 \caption{Summary of Evaluations of Traditional FL Models} 
	\label{tab:evaluation}
\resizebox{1\linewidth}{!}{
\begin{tabular}{c|c|c|c|c|c|c|c|c|c|c|c|c|c} \toprule  
{\textbf {\begin{tabular}[c]{@{}l@{}}Evaluations of \\ Traditional FL Models \end{tabular}}} &  
{\textbf {\begin{tabular}[c]{@{}l@{}}Detailed Methods\\ \usym{1F5F8}: suitable  \end{tabular}}}  & \multicolumn{2}{c|}{\textsc{Stakeholders}} & \multicolumn{2}{c|}{\makecell[c]{\textsc{Threat} \\ \textsc{Models}} } & \multicolumn{3}{c|}{\makecell[c]{\textsc{Privacy} \\ \textsc{Protection} \\ \textsc{Targets}}} & \multicolumn{3}{c|}{\makecell[c]{\textsc{Privacy} \\ \textsc{Protection} \\ \textsc{Techniques}} } & \multicolumn{2}{c}{\makecell[c]{\textsc{Effectiveness} \\ \textsc{Metrics}} } \\ \cline {3-14}
& & S1 & S2 & A1 & A2 & P1 & P2 & P3 & T1 & T2 & T3 & E1 & E2   \\ \hline
\multirow{2}{*}{\makecell[c]{Client \& Sample \\ Contribution Evaluation}} & Influence-based     & \checkmark &&\checkmark &\checkmark  & \checkmark& & & \checkmark & & &\checkmark & \\ \cline{2-14} 
& Shapley value-based & \checkmark & \checkmark &\checkmark & &\checkmark & & &\checkmark  &\checkmark & &\checkmark &\checkmark \\ \hline 
\multirow{2}{*}{\makecell[c]{Feature Contribution\\ Evaluation}} & Model-Agnostic &  & \checkmark & & & & & && \checkmark  &\checkmark &\checkmark & \\ \cline{2-14} 
& Model-Specific&&\checkmark&\checkmark& && &\checkmark & &\checkmark & &\checkmark &   \\ \hline 
\end{tabular}
}

\raggedleft
\renewcommand\arraystretch{0.8}
\begin{tabular}{@{}lllll@{}}
\scriptsize S1: To FL Server & \scriptsize A1: Semi-Honest Participants &  \scriptsize P1: Raw Data     & \scriptsize T1: Differential Privacy   & \scriptsize E1: Post-Interpretation Performance   \\
 \scriptsize S2: To FL Client & \scriptsize A2: Malicious Participants            & \scriptsize P2: Data Distribution & \scriptsize T2: Homomorphic Encryption   & \scriptsize    E2: Faithfulness     \\
   &    &     \scriptsize P3: Label        & \scriptsize T3: Secure Multi-party Computation
\end{tabular}%
\vspace{-6pt}

\end{table*}

\subsubsection{Contribution Evaluation for Traditional FL} 
\
\par
\textbf{Client and Sample Contribution Evaluations}:
\
\par
\indent The performance of the global FL model highly relies on the quality of the local dataset. Client and sample contribution evaluation can help the FL server analysis this through tracing back to local data, and is important to model aggregation. FL client and sample contribution evaluation methods can be divided into two main categories: 1) Shapley value (SV)-based evaluation and 2) influence-based evaluation (Table \ref{tab:evaluation}).

\underline{Shapley Value-based Evaluation}: SV-based evaluation methods provide insights into FL clients' data via evaluating their contributions to the performance of the final FL model. Existing SV-based evaluation methods mostly focus on improving computational efficiency, while maintaining accuracy performance of the estimated SV. The original SV method is prohibitively expensive since the required utility function evaluation grows exponentially with the number of FL participants. 

Existing FL evaluations based on efficient SV calculation can be divided into two categories: 1) accelerating within-round evaluations, and 2) decreasing the number of rounds of sub-model evaluation. 
For the accelerating within-round evaluation approach, instead of re-training them from scratch, gradient-based SV \cite{nagalapatti2021game} and local embedding-based SV  \cite{fan2022fair} methods have been proposed to reconstruct sub-models. For the decreasing the number of rounds of sub-model evaluation approach, three popular methods have been proposed. The first evaluates every possible sub-model within the original SV setting based on gradient-based estimation \cite{song2019profit, wei2020efficient}. The second method leverages randomly sampled permutation evaluation \cite{wang2020principled,wang2022efficient} which produces the estimated SV of an FL participant as its expected contribution. However, since the number of selected permutations is fixed, potentially important permutations may be overlooked, leading to inaccurate estimation. Thus, the third approach \cite{liu2022gtg,liu2022contribution} combines the within-round and between-round truncation approach and the guided Monte Carlo sampling  and in order to prioritize sub-model permutations based on their importance. 

Another branch studies the problem of secure SV calculation under malicious settings. 
In \cite{zheng2022secure}, a two-server secure SV calculation protocol was designed, which leverages a hybrid scheme to avoid ciphertext-ciphertext multiplications. Another work \cite{ma2021transparent} designed a group-based SV computation scheme leveraging a blockchain-based secure aggregation framework in order to protect participants' data privacy. 

\underline{Influence-based Evaluation}: 
Although SV-based FL client and sample contribution evaluation consider the complex dependencies among clients, they are generally highly expensive to compute. 
Influence-based FL contribution evaluation methods have been proposed to efficiently evaluate the contribution of FL clients and their local data samples on FL model performance. 
Existing influence-based methods can be divided into two main branches. The first \cite{wang2019measure,zhang2022intrinsic} perturbs clients or their local samples to retrain FL models, and uses the difference in performance (\emph{e.g.}, test loss, test accuracy) between the new and the original model to approximate client or sample contribution.  
However, since this method requires retraining of FL models on all clients' datasets or individual samples, the evaluation procedure is prohibitively expensive \cite{li2021efficient, koh2017understanding}. 

The second category uses influence function methods \cite{koh2017understanding} that leverage the second-order optimization technique to avoid the expensive retraining. 
An early approach leveraging influence functions in client contribution evaluation is Fed-influence \cite{xue2021toward}. 
To measure the influence of a client, Fed-influence sums up all sample influence values since there is an additive property of 
the influence function when measuring changes in test predictions \cite{koh2019accuracy}. However, this approach requires clients to  calculate the inverse of the Hessian matrix and transmit it, which incurs large computation and communication overhead. 
To reduce the overhead of influence approximation, emerging methods \cite{li2021efficient,li2021privacy, kwon2023datainf, li2021sample} leverage the Hessian vector product to approximate the influence values. They are capable of achieving the linear operation costs, making them promising for practical adoption.

\textbf{Feature Contribution Evaluation}:
\
\par
\indent
\underline{Model-Agnostic Evaluation}:
These methods consider an FL model as a black-box and attempts to measure the relevance of each feature to the learning task to filter out irrelevant ones. 
They leverage various statistical measures (\emph{e.g.}, mutual information, F-statistics, Gini-impurity\cite{cassara2022federated, li2021privacy1, pansecure}) to compute per-feature relevance scores.   
Another branch of methods use efficient gradient-based SV estimation approaches for feature contribution evaluation \cite{wang2019measure,wang2019interpret}. However, they methods require the FL server to familiarize all the IDs of clients' features, which might violate clients privacy and make them unsuitable for practical vertical FL applications.


\underline{Model Specific Evaluation}: 
The utilization of attention-based evaluation method enables the server to interpret which specific parts of inputs are leveraged by the global FL model. 
In \cite{chen2020federated}, a hierarchical attention mechanism is proposed which develops task-specific attentions to access personal feature correlations. 
Besides, a temporal attention layer is designed to evaluate cross-client temporal correlations at the FL server level. The final visualization of the attention weights is used to determine which features the global model focuses for individual predictions. Another work, Flames2Graph \cite{younisflames2graph}, offers a personalized evaluation approach for the multivariate time series FL classification issue, which can extract and visualize the essential sequences that highly activate network neurons to capture the temporal dependencies among them. 
Furthermore, \cite{li2023fedsdg,li2023efficient} proposed a federated feature selection method, which introduces a Gaussian stochastic dual-gate based on the $l_0$ constraints to efficiently and privately approximate the probability of selected feature.


\subsubsection{Contribution Evaluation for Centralized FMs}
\
\par
Existing evaluation approaches designed for FMs can be divided into: 1) model-based evaluation, and 2) training data-based evaluation. 

\textbf{Model-based FM Evaluation}:
\
\par
\indent
The initial objective of language models, particularly FMs, is to enhance language processing task performance (\emph{e.g.}, inference accuracy, robustness, trustworthiness). There are two common evaluation methods for FMs: 1) automated evaluation, and 2) human evaluation. 

\underline{Automated Evaluation}: 
Automated evaluation of FMs commonly adopts standard indicators or  metrics and evaluation tools to assess model performance. These include accuracy, fairness, ROUGE \cite{lin2004rouge} and BERTScore \cite{zhang2019bertscore}. For instance, the BLEU score \cite{papineni2002bleu} has been leveraged to measure the similarity and quality between the text generated by the FM and the reference text by the machine translation task. 
Due to its simplicity and automatic computing, this evaluation metric is widely adopted by existing FM evaluation effort. In addition, with the deployment of FMs, more advanced automated evaluation approaches are emerging. For example, Lin et al. \cite{lin2023llm} proposed a unified multidimensional automated evaluation approach, LLM-EVAL, for domain conversations. PandaLM \cite{wang2023pandalm} trained an additional LLM to evaluate different models, which is suitable for reproducible and automated language model evaluation.

\underline{Human Evaluation}: 
Human evaluation aims to assess the quality and accuracy of FM-generated results by human participation. Different from automated evaluation, human evaluation focuses more on specific application scenarios and can provide more comprehensive results. Usually, evaluators consist of experts, researchers or the target users. Recent human evaluation methods involve tasks including generation, summarization and analogical reasoning tasks.  
Bubeck et al. \cite{bubeck2023sparks} conducted various human-crafted tests on GPT-4. It can be observed that GPT-4 performs close to or even exceeds human performance on various tasks. In practice, automated evaluation and human evaluation are considered for adoption on a case-by-case basis. 
\vspace{6pt}

\textbf{Training Data-based FM Evaluation}:
\
\par
\indent
During fine-tuning, the training data might not be equally important for a pre-trained FM. 
On the one hand, pre-trained FMs are prone to significant performance degradation with noisy data. This effect can be further amplified when noisy samples are highly influential to the model. 
On the other hand, specific knowledge embedded within some samples might have been extracted after several training rounds. Therefore, they can be ignored afterwards without affecting final fine-tuned model performance. 
Therefore, it is important to identify noisy samples and important samples in the fine-tuning data to improve training efficiency and model performance. 
Jain et al. \cite{jain2023bring} evaluate FMs from the data-based perspective by eliminating the need for laborious labeling of new data. 
In \cite{schoch2023data}, an efficient SV-based data evaluation method has been proposed. It achieves this design goal through an efficient sampling-based method that aggregates SV values calculated from subsets for valuation of the entire training set, and a value transfer method that leverages value information extracted from a simple classifier trained by representations. DataInf \cite{kwon2023datainf} is a computationally efficient influence approximation method that is based on an easy-to-compute closed-form expression. It can be easily applied to FMs.

\subsubsection{Contribution Evaluation for \methodname{}}
\
\par
As for evaluating \methodname{}, so far, there have been few relevant works. 
The work \cite{xing2023fedlogic} proposed a logic rule learning approach to select the optimal chain-of-thoughts prompts for improving the interpretability of federated prompt selection for multi-domain FMs. They cast this problem as a bi-level program, and solve it through variational expectation maximization. 
The work \cite{zhao2023privacy} proposed to evaluate \methodname{} by comparing the performance of the federated parameter-efficient fine-tuned model with traditional centralized fine-tuning methods. 
FedIT \cite{zhang2023towards} leverages FL framework for FM instruction tuning and conducted studies on the widely used GPT-4 to exploit the heterogeneous and diverse sets of instructions.  
FedNLP \cite{lin2021fednlp} is a benchmark framework for evaluating FL methods on four common formulations of NLP tasks: text classification, sequence tagging, question answering and seq2seq generation. They proposed a universal interface between Transformer-based language models (e.g., BERT, BART) and FL methods under various non-IID partitioning strategies. SV-based methods have been extensively employed in conventional FL scenarios to assess the contribution of each participant. However, applying SV-based methods to measure contributions in \methodname{} settings presents two notable challenges: inference latency and compositional gap. 

\textbf{Inference Latency}:
\
\par
\indent
Due to the substantial number of model parameters, directly utilizing SV-based methods for contribution evaluation in \methodname{} is impractical \cite{liu2022gtg}. The vast scale of participants, coupled with a large number of participation records, significantly amplifies the complexity and time-latency involved in both training and inference. The large model size further hinders the feasibility of incorporating timely contribution feedback to update and refine the utility of each participant, as is customary in traditional FL settings. This timely feedback plays a pivotal role in the participant selection process.

\textbf{Compositional Gap}:
\
\par
\indent
FMs often grapple with the issue of a compositional gap, where they struggle to generate correct answers to compositional problems, even though they can correctly answer all their sub-problems \cite{press2022measuring}. Directly requiring contributions from FMs for each participant is currently beyond their capabilities. FMs cannot fully exploit the open-world knowledge encoded in them \cite{kang2023llms, dai2023uncovering}, making it challenging to assess the contribution of each participant or even each component under \methodname{} settings.
\vspace{-3pt}

\subsection{\methodname{} Reward Distribution}
\label{sec:reward distribution}
\begin{figure*}[t]
    \centering
    \includegraphics[width=1\linewidth]{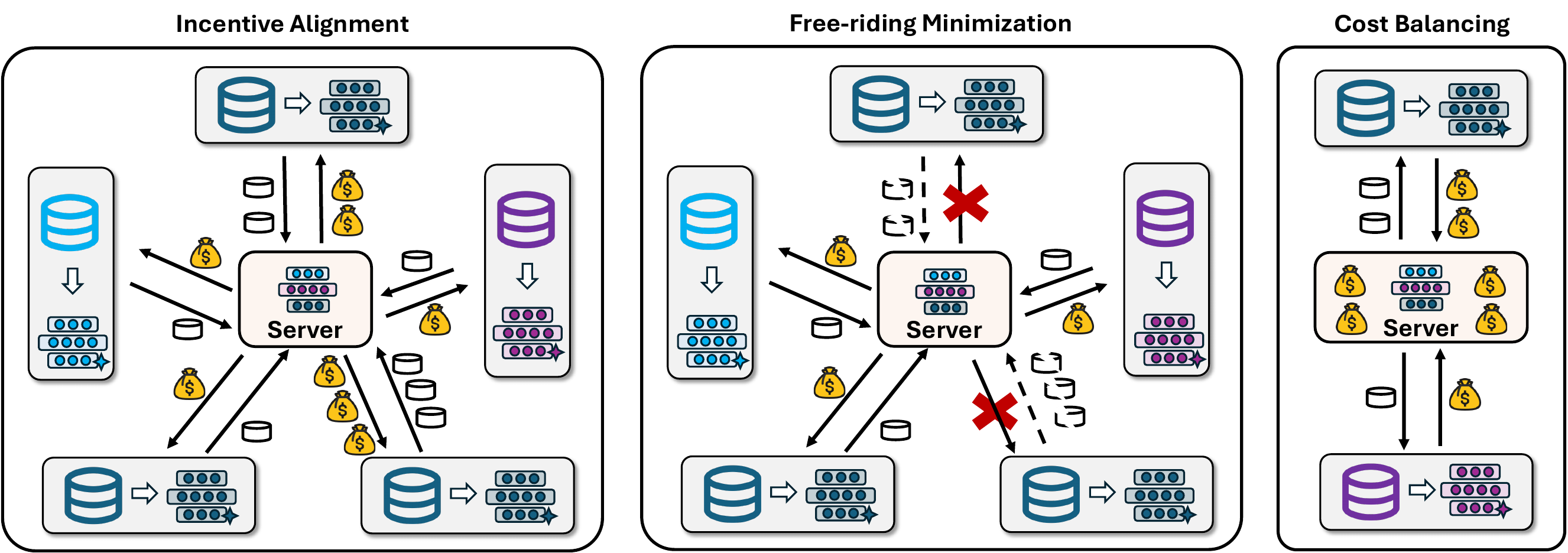}
    \caption{Illustration of reward distribution in \methodname{}.}
    \label{fig:reward_distribution-FedFMs}
    \vspace{-6pt}
\end{figure*}

Similar to traditional FL, \methodname{} reward distribution, shown in Fig. \ref{fig:reward_distribution-FedFMs}, needs to address the following issues.

\subsubsection{Incentive Alignment} 
\
\par
\indent
Reward distribution mechanisms must align with the overarching goals of \methodname{}. This includes decisions about the structure of rewards (e.g., monetary vs. non-monetary), whether rewards are distributed centrally or via smart contracts on a blockchain, and the criteria for reward eligibility. Ensuring that incentives align with the \methodname{} objectives is essential for motivating participation.

\subsubsection{Free-Riding Minimization}
\
\par
\indent
To maintain the integrity of the incentive structure, it is crucial to prevent free-riding, where some participants benefit without actively contributing to \methodname{}. Effective mechanisms, such as requiring a minimum level of participation or using reputation systems, can deter free-riding behaviours.

\subsubsection{Cost Balancing} 
\
\par
\indent
Managing the costs associated with rewards while ensuring the long-term sustainability of \methodname{} is a delicate balancing act. Striking the right balance between incentivizing participants adequately and maintaining the financial viability is essential for long-term sustainable operation of \methodname{}.


\section{\methodname{} Implementation and Applications}
\label{sec:implementation}

{
Implementing \methodname{} requires careful consideration of both hardware and platform capabilities for handling the computational and storage demands of large-scale models. Here, we provide a concise tutorial on the key components and tools commonly used in \methodname{} applications, with reference to platforms in Table \ref{tab:frameworks} and representative applications in Table \ref{tab:application}.
}
\subsection{\methodname{} Implementation}
\label{sec:framework}
{
Setting up \methodname{} requires a multi-faceted approach involving both software and hardware aspects to enable distributed training and fine-tuning across diverse devices, while maintaining data privacy. It can be facilitated by well-designed platforms and libraries supporting FL and the requirements of FMs. Here, we discuss leading FL frameworks which support large-scale FMs, including advanced hardware configurations and specialized platforms.
}

\begin{table*}[htbp]
\caption{Comparative Analysis of FL Platforms and Libraries that Support FMs.}
\label{tab:frameworks}
\resizebox{\textwidth}{!}{
\small
\begin{tabular}{lllllll}
\toprule
\textbf{Platforms}                              & \textsc{\begin{tabular}[c]{@{}l@{}}No. of \\ Aggregation \\ Methods \end{tabular}}                                          & \textsc{\begin{tabular}[c]{@{}l@{}}Methods to Improve\\ Computational \\ Efficiency\end{tabular}} & \textsc{\begin{tabular}[c]{@{}l@{}}Methods to Improve\\ Communication  \\ Efficiency\end{tabular}}     &  \textsc{\begin{tabular}[c]{@{}l@{}}No. of \\ Datasets \end{tabular}} & \textsc{\begin{tabular}[c]{@{}l@{}}Largest \\ Model\end{tabular}} \\
\midrule
FedML\cite{he2020fedml}  & 11    & PEFT (powered by HuggingFace)            & None       & Unknown      & LLaMa2 (7B)    \\ \hline
FederatedScope-LLM\cite{kuang2023federatedscope} & 1    & \begin{tabular}[c]{@{}l@{}}LoRA, P-Tuning, Prompt-Tuning, \\ Instruction-Tuning\end{tabular} & None    & 6                        & LLaMa (7B)          \\ \hline
Fate-LLM\cite{fan2023fate}      & 1               & \begin{tabular}[c]{@{}l@{}}LoRA, Prompt Tuning, \\ Full Fine-Tuning\end{tabular}            & \begin{tabular}[c]{@{}l@{}}Quantization,\\ Knowledge Distillation\end{tabular} & 3      & LLaMa (7B)     \\  \hline
OpenFedLLM \cite{ye2024openfedllm}         & 7     & \begin{tabular}[c]{@{}l@{}}LoRA, Instruction Tuning\end{tabular}                             & Quantization     & 8    & LLaMa2 (7B)    \\
\bottomrule
\end{tabular}
}
\vspace{3pt}
\end{table*}

\subsubsection{Platform and Library Considerations}
\
\par
\indent
{
An overview of prospective platforms and libraries for \methodname{} are shown in Table \ref{tab:frameworks}. We focus on frameworks including FedML \cite{he2020fedml}, FederatedScope-LLM \cite{kuang2023federatedscope}, FATE-LLM \cite{fan2023fate}, and OpenFedLLM \cite{ye2024openfedllm}, comparing the methods supported, dataset compatibility and maximum model sizes as showcased in their examples. 
}

\begin{itemize}
    \item FedML \cite{he2020fedml} is a flexible and scalable platform designed for implementing and experimenting with various FL models. It supports both simulation and real-world deployment, making it an ideal choice for research and practical applications of \methodname{}. FedML's integration with popular machine learning frameworks like TensorFlow and PyTorch allows for seamless adaptation of foundation models in federated settings.
    \item FederatedScope-LLM \cite{kuang2023federatedscope} is an extension of FederatedScope focused on LLMs. It addresses the unique challenges of training LLMs in a federated environment, such as efficient communication and privacy preservation. This platform is relevant for \methodname{} as it provides specialized tools to manage the complexities of \methodname{}.
    \item FATE-LLM \cite{fan2023fate}, a specialized version of the Federated AI Technology Enabler (FATE) platform, is tailored for LLMs. It leverages secure computation protocols and privacy-preserving techniques, making it suitable for \methodname{} applications that require strong data security and compliance with privacy regulations.
    \item OpenFedLLM \cite{ye2024openfedllm} is an open-source framework aimed at facilitating the \methodname{}. Its extensibility and ease of use make it a valuable tool for developing \methodname{} workflows. OpenFedLLM supports various FL paradigms and integrates with existing ML libraries, promoting innovation in \methodname{}.
\end{itemize}

{
The findings reveal a trend of integrating state-of-the-art techniques to facilitate \methodname{}. However, there appears to be little emphasis on computational efficiency. Notably, all frameworks have been assessed with contemporary large-scale FMs, underscoring their practical relevance. Utilizing these frameworks can facilitate the reassessment of techniques previously surveyed with actual FMs, potentially inspiring novel research directions. Implementing \methodname{} is an emerging and evolving research area, with continuously improving tools and practices. Selecting the right combination of platforms and libraries depends on the specific requirements of \methodname{} tasks, including the type of FMs involved, privacy, security and scalability requirements, as well as the computational resources available.
}

\subsubsection{Hardware Considerations}
\
\par
\indent
{
High-performance central servers are crucial for coordinating the \methodname{} process. These servers, often hosted on cloud platforms such as AWS\footnote{\url{https://aws.amazon.com/}}, Google Cloud\footnote{\url{https://cloud.google.com/?hl=en}}, Microsoft Azure\footnote{\url{https://azure.microsoft.com/en-gb}} or NVIDIA FLARE\footnote{\url{https://developer.nvidia.com/flare}}, need to have high-end configurations including multi-GPU setups, large memory, and extensive storage solutions to manage the aggregation and training of large-scale models. Distributed edge servers, equipped with multiple GPUs, high-core-count CPUs, and large memory, act as intermediaries between the central server and edge devices to reduce latency and enhance scalability.

Regarding edge devices, high-performance smartphones and tablets equipped with the latest processors, substantial RAM, and storage capabilities are suitable for initial phases of model training and inference tasks. Powerful laptops and desktops with multi-core CPUs, ample RAM, and dedicated GPUs are essential for edge clients participating in \methodname{}. In addition, advanced IoT devices with enhanced processing units, such as NVIDIA Jetson series\footnote{\url{https://www.nvidia.com/en-us/autonomous-machines/embedded-systems/}} or Google Coral AI accelerators, can support model inference and lightweight training tasks in distributed settings.
}

\subsubsection{Practical Implementation}
\
\par
\indent
{
For simulation and testing, high-performance computing clusters equipped with multiple GPUs and high-speed interconnects can mimic large-scale federated environments and provide insights into performance bottlenecks. Cloud-based simulation platforms such as AWS EC2 with p3/p4 instances, Google Cloud with TPU support, and Microsoft Azure with NDv2/NDv4 instances offer scalable simulation and testing environments for \methodname{} setups.

Deployment of \methodname{} systems benefits from containerization and orchestration tools like Docker and Kubernetes\footnote{\url{https://kubernetes.io/}}. Kubernetes operators can manage the lifecycle of FL jobs, ensuring efficient resource allocation and fault tolerance. For edge deployment, tools like NVIDIA Fleet Command\footnote{\url{https://resources.nvidia.com/en-us-retail-and-edge-ai/fleet-command-web-page?xs=205038&ncid=no-ncid}} and AWS IoT Greengrass\footnote{\url{https://aws.amazon.com/greengrass/}} facilitate the deployment and management of FL models on edge devices, supporting secure and efficient model inference and updates in distributed environments. 
}

\subsection{\methodname{} Applications}
\label{sec:application}
{
For NLP applications, \methodname{} enables collaborative training of FMs across diverse datasets, enhancing language understanding and generation capabilities while preserving data privacy. For speech recognition, \methodname{} facilitates the training of robust models that can generalize across different accents and languages, leveraging distributed data without centralizing sensitive voice data. In recommender systems, \methodname{} supports the development of personalized recommendations by aggregating user preferences from multiple sources, ensuring privacy and enhancing recommendation accuracy. In healthcare, \methodname{} allows for the aggregation of medical data from various institutions, enabling the development of more accurate diagnostic models and personalized treatment plans without compromising patient privacy. 
The \methodname{} applications are shown in Table \ref{tab:application} and the detailed descriptions of each application are given as follows.  
}

\begin{table*}[!ht]
\centering
\caption{Related Works on the Applications of \methodname{}}
\label{tab:application}
\resizebox{\textwidth}{!}{
\small
\begin{tabular}{c|cclcc}
\hline
\textbf{\makecell{Application \\ (Modality)}} & \textsc{Related Works} & \textsc{Year} & \textsc{Task and Domain} & \textsc{On-Device} & \textsc{Personalization} \\ \hline
\multirow{2}{*}{\makecell[c]{Healthcare (Vision and Text)}} & FedPR \cite{feng2023learning} &  2023 & MRI Reconstruction & No & No \\  
 & FedTherapist \cite{shin2023fedtherapist} & 2023 &Mental Health Prediction & Yes & No \\ \hline
\multirow{3}{*}{\makecell[c]{Recommender System (Text)}} & GPT-FedRec \cite{zeng2024federated} & 2024 &General & No & No \\ 
 & TransFR \cite{zhang2024transfr} & 2024 & General & Yes & Yes \\  
 & PPLR \cite{zhao2024llm} & 2024 & General & No & Yes \\ \hline
\multirow{3}{*}{\makecell[c]{Speech (Audio)}} & FedE2EASR \cite{azam2023federated} & 2023 &Speech Recognition & No & Yes \\ 
 & FedASR \cite{jia2023joint} & 2023 &Speech Recognition & Yes & Yes \\  
 & pFedS2T \cite{du2024communication} & 2024 &Speech-to-Text & No & Yes \\ \hline
\multirow{5}{*}{\makecell[c]{Multilingual NLP (Text)}} & MFPT \cite{zhao2024breaking} & 2024 &Multi-Tasks & Yes & No \\ 
 & FL-MetaSend \cite{chu2024only} & 2024 & Machine Translation & No & No \\ 
 & Fed-MNMT \cite{liu2023communication} &  2023 & Machine Translation & No & No \\ 
 & PMMFL \cite{weller2022pretrained} &  2022 &Multi-Tasks & No & No \\ 
 & FedKC \cite{wang2022fedkc} &  2022 &Language Understanding & No & No \\ \hline
\end{tabular}
}
\end{table*}

\begin{itemize}
    \item In healthcare, FedPR \cite{feng2023learning} focused on magnetic resonance imaging (MRI), using an FM pre-trained on public datasets and training visual prompts from decentralized clinical data via a personalized FL mechanism, achieving competitive performance with limited local data. Shin et al.\cite{shin2023fedtherapist} developed a mobile mental health monitoring system that uses user speech and keyboard input to fine-tune \methodname{}, demonstrating high accuracy in predicting mental health conditions such as depression, stress, and mood.
    \item For recommender systems, GPT-FedRec \cite{zeng2024federated} utilized ChatGPT's strong zero-shot generalization ability in federated recommendation, facilitating hybrid retrieval and re-ranking of recommendation systems. Zhang et al. \cite{zhang2024transfr} introduced a transparent federated recommendation framework that enhances recommendation accuracy while maintaining user privacy. Zhao et al. \cite{zhao2024llm} presented a personalized privacy-preserving language representation learning method, using FL to protect user data privacy.
    \item In the speech domain, Azam et al. \cite{azam2023federated} proposed federated end-to-end automatic speech recognition systems, leveraging \methodname{}'s advantages in privacy-sensitive audio scenarios. FedASR \cite{jia2023joint} showed that FL can improve automatic speech recognition by training collaboratively on decentralized audio data. Du et al. \cite{du2024communication} introduced personalized federated speech-to-text models, addressing the need for accurate speaker-independent performance while offering personalized adaptation.
    \item For multi-lingual NLP, Zhao et al. \cite{zhao2024breaking} aimed to enhance training efficiency in multilingual NLP by adapting pre-trained \methodname{} through prompt tuning. FL-MetaSend \cite{chu2024only} used knowledge distillation to selectively transfer global knowledge based on entropy measures, improving generalization across various domains. Liu et al. \cite{liu2023communication} explored different clustering methods to group adapter parameters, reducing the negative impact of data diversity. They found that language family-based clustering significantly outperforms other strategies. Weller et al. \cite{weller2022pretrained} demonstrated that fine-tuning pre-trained \methodname{} can achieve performance comparable to centralized fine-tuning in NLP settings. FedKC \cite{wang2022fedkc} employs k-means clustering on client data to generate representative knowledge, specifically in the form of clustered data centroids. These centroids are subsequently shared among clients for local training, enhancing data richness and addressing heterogeneity issues.

\end{itemize}

\section{Lessons Learned on \methodname{}}
\label{sec:lesson}

{
This section summarizes the key insights and takeaways from Sections \ref{sec: training and aggregation}- \ref{sec:framework}, providing readers with a clear understanding of the practical implications and broader context of the findings from this survey. In addition, we have connected these lessons to the topics discussed in Section \ref{sec:future} to form a cohesive narrative.

Section \ref{sec: training and aggregation} focuses on efficient training and aggregation methods in \methodname{}. In \methodname{}, the heterogeneity of client devices in terms of computational resources and data distributions necessitates sophisticated customization of model architectures and training protocols. Techniques such as PEFT and Prompt Tuning are pivotal for optimizing performance and efficiency. PEFT methods, including adapter tuning, BitFit and LoRA, allow selective fine-tuning of a small subset of parameters, significantly reducing computational and communication overhead while maintaining high model performance. These methods are crucial for adapting large models to resource-constrained environments by selectively updating the most relevant parameters. Prompt tuning involves incorporating task-specific prompts into the input data, providing an efficient means of adapting pre-trained models to new tasks with minimal additional training. Robust and efficient aggregation methods, such as hierarchical aggregation, are also essential to reduce communication costs by aggregating updates at multiple levels. The key lesson here is the necessity of developing adaptive and efficient training and aggregation strategies that can cater to the heterogeneous environments typical of \methodname{} deployments, ensuring robust and scalable model training.

Section \ref{sec:trustworthy} focuses on the critical areas of robustness, privacy, and intellectual property for \methodname{}. Ensuring robustness and privacy in \methodname{} is critical due to the sensitivity of the data involved and the potential for malicious attacks. Byzantine robustness addresses the challenge of dealing with malicious or faulty clients that can disrupt the learning process. Techniques such as robust aggregation methods, including geometric median and trimmed mean, help in mitigating the impact of Byzantine clients by aggregating updates in a way that reduces the influence of outliers. Privacy preservation is equally crucial, with techniques like DP and secure aggregation playing a vital role. Protecting intellectual property is vital to securing the ownership of \methodname{} from unauthorized exploitation, including model theft. The lesson learned here is the critical importance of developing methods that balance robustness and privacy with performance, meanwhile protecting intellectual property of \methodname{}. Continuous advancements in cryptographic techniques and robust aggregation algorithms are essential for enhancing the security and resilience of \methodname{} systems against adversarial threats.

Section \ref{sec:incentive} focuses on the incentive mechanism and contribution evaluation for \methodname{}. The successful deployment of \methodname{} relies heavily on the active participation of diverse stakeholders, necessitating effective incentive mechanisms. These mechanisms must ensure fairness, transparency, and security, motivating data owners to contribute their resources while deterring malicious behavior. Contract theory-based approaches can formalize participation terms, ensuring stakeholders are fairly compensated. Game theory-based methods model interactions between participants as strategic games, designing mechanisms that promote cooperative behavior and optimize collective outcomes. Auction-based methods provide a dynamic approach to incentivization, allowing participants to bid for the opportunity to contribute or receive rewards based on their resources and data quality. A critical takeaway from this section is that incentive mechanisms must be tailored to the specific needs and behaviors of \methodname{} participants. Combining multiple incentive strategies can address diverse motivations, ensuring sustained and meaningful participation, which is crucial for the success of \methodname{}.

Section \ref{sec:lesson} focuses on the implementation and applications of \methodname{}. The implementation of \methodname{} requires a comprehensive framework that integrates various components essential for effective deployment. This section discusses a structured approach to implementing \methodname{}, encompassing data management strategies, model training protocols, and efficient communication mechanisms. The framework emphasizes modularity, allowing for seamless integration of new methods and technologies as they evolve. Best practices in data preprocessing, feature extraction, and model optimization are incorporated to ensure robust and scalable implementations. In addition, this section highlights the transformative potential of \methodname{} across various applications, including NLP, speech recognition, recommendation systems, and healthcare. The lesson learned is that leveraging the comprehensive structure provided by the \methodname{} framework, along with its scalability and adaptability, can significantly enhance the effectiveness and reach of FL implementations, ensuring they meet the specific requirements of various real-world applications.
}

\begin{figure*}[t]
    \centering
    \includegraphics[width=\linewidth]{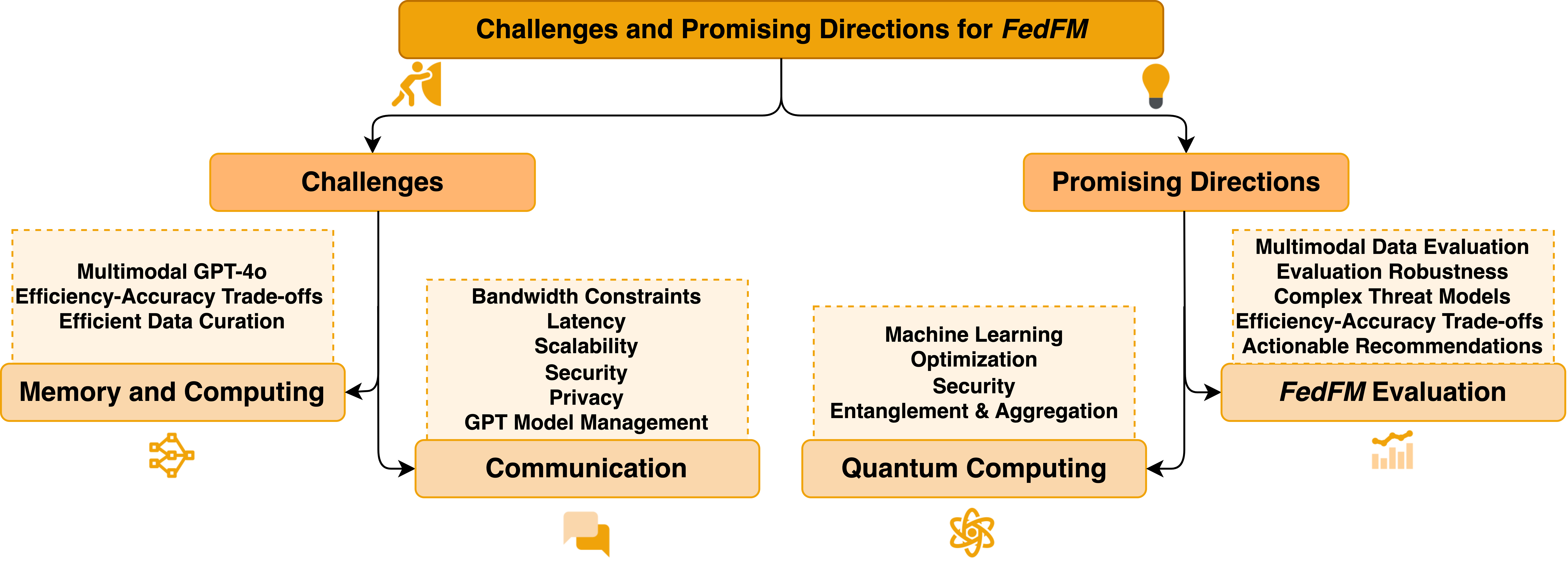}
    \caption{Challenges and promising directions for \methodname{}.}
    \label{fig:challenges-FedFMs}
    \vspace{-6pt}
\end{figure*}

In summary, the lessons learned from Section \ref{sec:lesson} provide a foundation for the envisioned future directions and open challenges discussed in Section \ref{sec:future}. By understanding the practical implications and challenges identified in our survey, researchers and practitioners can better develop key areas such as advanced customization, enhanced privacy mechanisms, comprehensive incentive strategies, and scalability solutions. Future research should focus on developing more adaptive strategies that can dynamically adjust to diverse client environments, robust privacy-preserving techniques that minimize performance trade-offs, and comprehensive incentive mechanisms that integrate multiple approaches to effectively motivate diverse participants. Additionally, exploring scalable and adaptable solutions will be crucial for meeting the evolving demands of real-world applications and advancing the field of \methodname{}. 

\section{Challenges and Promising Directions}
\label{sec:future}

Bringing \methodname{} into practice involves overcoming some technical and logistical challenges, including managing heterogeneous data and compute resources, ensuring robust and secure communication, and developing efficient algorithms for federated optimization and aggregation. 
Due to regulatory considerations, when implementing \methodname{}, it is crucial to monitor and mitigate biases that might arise from uneven data distributions across data owners. Auditing and fairness-aware algorithms can help address these concerns. 
In this section, we discuss the memory, computation and communication challenges of \methodname{}, as well as promising future research directions, exploring the synergy between quantum computing and \methodname{} as well as novel ways of evaluation to help the field advance further.

\subsection{Memory and Computing Challenges facing \methodname{}}
The integration of FMs with FL poses several key challenges, primarily focused on managing the substantial requirements for memory and computation resources.

{\bf Multi-modal GPT-4o}: The large scale of FMs, often involving billions of parameters, results in significant memory demands. This is particularly challenging in FL environments, where client devices may have limited memory capacities. The key challenge is developing strategies to accommodate these large models efficiently within the memory constraints of diverse client devices. For multi-modal GPT-4o Model, it integrates text, audio, and vision capabilities in real-time, promising to transform human-machine interaction\cite{gao2023examining,gao2023multi}. Further \methodname{} research can focus on seamless integration and processing of multi-modal data from diverse sources while preserving privacy. Enhancing cross-modal learning in federated settings enables models to leverage insights from different data types effectively. In addition, optimizing the performance of multi-modal models like GPT-4o in FL frameworks is vital, particularly for communication efficiency and computational demands.

In \methodname{}, the demand for high computational resources on FL client devices poses a persistent challenge. Despite strategies to mitigate this, it remains an open issue. 
Approaches such as model splitting \cite{thapa2022splitfed} and model compression \cite{shah2021model} have been proposed.
In addition, efforts are made to develop lighter models and more efficient training algorithms \cite{he2020group} that are less taxing on device resources. 
Yet, these solutions only offer a partial remedy. 
The research on federated optimization \cite{konevcny2016federated, mcmahan2017communication} indicate that finding the right balance between maintaining model performance and managing resource constraints is an ongoing struggle. 
Although these advances mark progress, ensuring that \methodname{} is adaptable for devices with diverse capabilities continues to be a significant open research question. 

{\bf Balancing Efficiency with Model Performance}: A pivotal aspect of \methodname{} research is finding a balance between reducing model sizes for compatibility with FL constraints while maintaining model performance~\cite{ren2023secfedsa}. New strategies need to be developed for achieving this balance, ensuring that models are both efficient in terms of resource utilization and effective in terms of performance.

{\bf Efficient Data Curation}: Managing the distribution of large datasets necessary for training FMs in a federated manner presents unique challenges. Key to this is the development of methods for efficient data handling, which minimizes data redundancy and optimizes data usage across multiple nodes during the training process.


\subsection{Communication Challenges facing \methodname{}}
Through revealing the inherent complexity of FMs, combined with the distributed nature of FL, we identify the following key communication challenges facing \methodname{}.
This section discusses the multifaceted challenges associated with communication overheads, bandwidth constraints, latency, resource demand, security, scalability, and privacy, which affect the operational efficiency of \methodname{} (Table \ref{tab:communication}).

{\bf Increased Communication Overhead vs. Bandwidth Constraints}: Bandwidth constraints present a critical challenge to \methodname{} \cite{fariborz2022llm}. 
The sheer size of FMs necessitates large transmission overhead, which is compounded by the need for frequent communication across distributed training devices (i.e., clients and servers).
While strategies like model compression (using techniques like pruning, quantization, and knowledge distillation) aim to reduce the size of the models, and thus the network load \cite{jaiswal2023compressing}, they only partially alleviate the bandwidth demand. 
Sparse communication techniques \cite{mestoukirdi2023sparser}, which focus on sharing only essential model updates, help to some extent but do not entirely solve the problem. Asynchronous communication strategies \cite{xia2021survey} and edge computing \cite{wang2019edge} can reduce network congestion and data transmission volume. Yet, they cannot fully compensate for the inherent high transmission requirements of \methodname{}. 
Efficient coding schemes optimize data transmission \cite{yue2022communication}, but the fundamental challenge of transmitting large volumes of data in a bandwidth-limited environment persists. 
Collectively, these approaches would make strides in addressing bandwidth issues, but they do not completely solve the challenge, highlighting the pressing need for innovative solutions in the field of \methodname{} \cite{zhuang2023foundation}.


{\bf Latency Issues}: The long time it takes to transmit large volumes of parameter across clients and the FL server is a significant obstacle in \methodname{}. 
This delay is more pronounced when dealing with complex models and numerous, often geographically dispersed, FL clients. 
To mitigate this, 
techniques such as efficient data serialization \cite{beutel2020flower} and optimizing network protocols have been investigated. 
Moreover, implementing edge computing \cite{wang2019edge}, where data processing occurs closer to the sources, helps in reducing the round-trip time for parameter transmission. 
Despite these efforts, latency remains an open challenge for \methodname{}, often impacting the overall speed and responsiveness of the model training and fine-tuning processes \cite{xu2023federated}.

\begin{table*}[t]
\caption{Summary of Communication Challenges and Potential Mitigation Strategies in \methodname{}}
\label{tab:communication}
\resizebox{\textwidth}{!}{
\small
\begin{tabular}{l|l|l}
\hline
\textbf{Challenge}                              & \textsc{\begin{tabular}[c]{@{}l@{}}Mitigation \\ Strategies\end{tabular}} & \textsc{Relevant Works} \\ 
\hline
{Bandwidth Constraints / Latency Issues} & \begin{tabular}[c]{@{}l@{}}Model Compression, Sparse Communication,\\
 Asynchronous Communication, Efficient Coding Schemes\end{tabular}  & \cite{zhang2023fedpetuning,shen2024large,wu2023fedcomp,mao2023safari,xu2023asynchronous,liu2021federated}\\
\hline
{Resource Demands on Clients} &\begin{tabular}[c]{@{}l@{}} Model Splitting, Model Compression, \\ Lighter Models, Efficient Training Algorithms \end{tabular} & \cite{thapa2022splitfed,chawla2024beyond,zhang2024fedsl,liao2023accelerating,konevcny2016federated}\\
\hline
Communication Overheads / Scalability Issues  &  \begin{tabular}[c]{@{}l@{}} Efficient Network Protocols, Load Balancing, \\ Active Client Selection\end{tabular} & \cite{aledhari2020federated,tran2019federated,charles2024towards,zhang2023fedpetuning}\\
\hline
Privacy Concerns and Security Issues &  Homomorphic Encryption & \cite{jin2023fedml,rieyan2024advanced,aziz2023exploring}\\
\hline
\end{tabular}
}
\vspace{-3pt}
\end{table*}

{\bf Scalability Issues}: This problem arises because as more devices join the network, managing and processing their inputs becomes more complex and resource-intensive. 
To tackle this, strategies such as efficient network protocols \cite{aledhari2020federated,ren2021interpretable} and load balancing techniques \cite{tran2019federated} have been studied to manage the increased traffic and computational demands. 
Moreover, client selection \cite{shi2023towards}, where only a subset of clients are active at any given time, can help manage the load. 
However, despite these efforts, scalability remains a challenge for \methodname{}.
As the number of FL participants and the complexity of the FMs increase, achieving scalability remains a major obstacle, affecting both the effectiveness and adoption of \methodname{}.

{\bf Security Issues}: In \methodname{} supported by advanced network technologies like 5G and beyond (B5G), security issues pose a significant challenge. 
The diversity in computing and communication capabilities across different participants in the network, stemming from variations in hardware (e.g., CPU, GPU), network connections (e.g., 4G, 5G, B5G, WiFi) and energy resources (e.g., battery, charged), leads to system heterogeneity ~\cite{ren2022robustness,ren2021vulnerability,ren2022universal,zhang2024vulnerability}. 
This diversity can introduce inconsistencies and vulnerabilities in \methodname{} \cite{liu2020secure}. 
Moreover, the presence of unreliable devices within the network might lead to Byzantine failures as highlighted in \cite{fang2020local,tian2021adversarial}.
These failures refer to scenarios where certain nodes in the network act in a faulty or malicious manner, further complicating the security landscape of \methodname{}.
The varying levels of security across different devices exacerbate the difficulty in defending against attacks and ensuring system reliability. 
Despite ongoing efforts to enhance FL security \cite{li2020federated}, these inherent vulnerabilities in \methodname{} remain open.

{\bf Privacy Issues}: Privacy concerns in \methodname{} remain a significant challenge, despite the ongoing privacy protection research in traditional FL settings. 
While FL enhances privacy by sharing model updates rather than raw data, there are still vulnerabilities during interactions between FL participants \cite{lin2017deep}. 
For instance, adversaries can exploit these vulnerabilities to perform attacks such as membership inference or gradient leakage \cite{lin2017deep,tian2024lesson,tian2021joint}, aiming to extract local training data from devices.
Existing countermeasures, such as homomorphic encryption (HE) and secure multi-party computation (SMC), while effective in enhancing FL privacy to some extent, fall short in fully addressing these malicious attacks.
As detailed in \cite{huang2019survey}, HE can protect against data leakage but is less effective against more sophisticated threats like membership inference and gradient leakage attacks. However, it cannot be efficiently applied on large-scale FMs.
Despite these efforts in improving privacy protection, the evolving nature of threats and the complexity of FL systems, particularly in the context of FMs, means that privacy concerns remain a pressing and unresolved issue in \methodname{}. The development of more advanced and resilient privacy-preserving mechanisms continues to be a critical need in this field.


{
\textbf{GPT Model Management}: 
Currently, there are millions of new GPTs in the GPT Store. The GPT store concept envisions a vast repository of pre-trained GPT models, each specialized for different tasks, domains, or languages. \methodname{} can explore dynamic selection and personalization of the most suitable GPT models for specific participant needs. In addition, efficient \methodname{} techniques to aggregate updates from diverse GPT models while maintaining model performance and privacy are crucial. Optimal management of computational and data resources is also essential to support millions of GPT models in a federated environment, ensuring robust performance and low latency.
}

\subsection{Promising Direction: Quantum Computing for \methodname{}}

The integration of quantum computing with \methodname{} presents a transformative opportunity to enhance \methodname{}~\cite{group2023quafu}. This integration could revolutionize the efficiency and capability of these models~\cite{liang2023unleashing}. In this section, we envision how four promising quantum computing techniques (Table \ref{tab:quantum}): 1) quantum machine learning, 2) quantum optimization, 3) quantum security and 4) a combination of quantum entanglement and quantum aggregation hold promise to improve the efficiency, trustworthiness and incentives aspects of \methodname{}.

\textbf{Quantum ML for \methodname{}}: A key area in which quantum computing can make a significant impact is matrix operations, which are widely utilized in ML training. Quantum ML algorithms can perform these operations much faster than classical algorithms ~\cite{ren2023towards}. In the context of \methodname{}, this speedup can be particularly beneficial given the large matrices involved in such models~\cite{khan2017quantum}. Another crucial element in ML is gradient-based optimization, vital for training neural networks (NNs) in LLMs. Quantum algorithms promise more efficient gradient computation, potentially expediting the optimization process in \methodname{} settings. This enhancement hold the promise to lead to faster convergence during the training process.
    
The ability of quantum computing to efficiently perform certain types of sampling and approximation (e.g., Monte Carlo sampling) can be advantageous for many ML tasks. For example,~\cite{khan2021physical} have demonstrated a framework for reservoir computing with nonlinear quantum reservoirs, showcasing the computational capabilities across classical and quantum regimes. Quantum systems enable more efficient data structure representation, which can expedite computation in NNs during inference. When fine-tuning \methodname{} for specific tasks, feature selection is a crucial process. Quantum computing can offer efficient solutions for this process, enhancing the performance of the resulting FMs~\cite{bausch2020recurrent}. Lastly, quantum parallelism can support simultaneous evaluation of different hyper-parameter settings. This capability is significant for the hyper-parameter tuning process, a time-consuming aspect of FM fine-tuning. Moreover, the capability of quantum computing for handling high-dimensional spaces can aid in optimizing the high-dimensional parameter space of FMs more efficiently. 

Overall, quantum computing can accelerate training and inference, enhancing fine-tuning, and offering innovative approaches to complex computational tasks in \methodname{} settings. This paradigm shift can notably improve the efficiency and effectiveness of \methodname{}.

\begin{table*}[!htb]
\caption{Potential for Quantum Computing Techniques to Enhance \methodname{}}
\label{tab:quantum}
\resizebox{\textwidth}{!}{
\small
\begin{tabular}{ll|ccc}
\hline

\multicolumn{1}{l}{\multirow{2}{*}{\textbf{\begin{tabular}[l]{@{}l@{}}Quantum Computing\\  Techniques (\usym{1F5F8} means suitable) \end{tabular}}}} &
  \multirow{2}{*}{\textsc{General Objectives}} &
  \multicolumn{3}{c}{\textsc{Aspects of \methodname{}}} \\  \cline{3-5} 
\multicolumn{1}{c}{} &
   &
\begin{tabular}[c]{@{}l@{}}Efficiency\end{tabular} &
\begin{tabular}[c]{@{}l@{}}Incentivization\end{tabular} &
 \begin{tabular}[c]{@{}l@{}}Trustworthiness\end{tabular} \\ \hline

\multicolumn{1}{l}{Quantum Machine Learning}                                                          & \begin{tabular}[c]{@{}l@{}} Quantum Speedup for \methodname{} \\(including Training, Inference, and Fine-tuning)\end{tabular} 
 & \usym{1F5F8}              & \usym{1F5F8}               &                  \\ \hline
Quantum Optimization                                                                    &    \begin{tabular}[c]{@{}l@{}} Leveraging Quantum and Classical Capabilities \\to Optimize \methodname{} Objective and Cost Functions\end{tabular}   &          \usym{1F5F8}      &      \usym{1F5F8}          &                    \\ \hline
Quantum Security                                                                          &  \begin{tabular}[c]{@{}l@{}}   Quantum Enhanced Data Encryption \\and Quantum Communications for \methodname{}\end{tabular} 
&                  &                  &      \usym{1F5F8}            \\ \hline
\begin{tabular}[c]{@{}l@{}}Quantum Entanglement and Aggregation\end{tabular} &  
\begin{tabular}[c]{@{}l@{}}Quantum Entanglement Synchronized Updating \\and Quantum Aggregation for \methodname{}\end{tabular}  &        \usym{1F5F8}             &                  &        \usym{1F5F8}                \\ \hline
\end{tabular}
}
\end{table*}

\textbf{Quantum Optimization for \methodname{}}: Quantum optimization in ML, particularly in the context of \methodname{}, stands as a significant advancement in training methodologies. Leveraging quantum computing capabilities, these techniques aim to optimize ML models more effectively, focusing on finding the most suitable parameters to minimize loss functions and enhance overall performance. Methods like the Quantum Approximate Optimization Algorithm (QAOA) \cite{farhi2014quantum} and Variational Quantum Eigensolver (VQE) \cite{peruzzo2014variational} represent a leap forward in solving optimization problems. These quantum-based solutions can outperform classical algorithms, especially in navigating complex, high-dimensional solution spaces typical in FMs. Such algorithms are not only efficient in locating global minima, which can be challenging for classical methods, but also provide more effective ways to fine-tune model parameters, thereby accelerating convergence to optimal solutions and enhancing the efficiency of fine-tuning~\cite{kirby2021variational}. Quantum Annealing leverages quantum tunneling to efficiently explore solution spaces and find global minima of objective functions. Its ability to escape local minima makes it a valuable tool in the optimization process. It offers a novel approach to optimizing parameters, especially in \methodname{}. 

The integration of quantum and classical optimization steps in hybrid algorithms offers a balanced approach. By utilizing quantum processors for the more computationally demanding parts and classical processors for other parts, these hybrid models leverage the strengths of both quantum and classical computing. They can be particularly effective in overcoming current limitations of quantum computing, while still harnessing its advantages for optimization tasks~\cite{kuete2023universal}. Global optimization is a key factor in \methodname{} settings, where the objective is to achieve consensus among distributed models. Quantum algorithms can efficiently aggregate local updates and conduct global optimization. This can produce more effective and cohesive FMs in a collaborative learning environment. The incorporation of quantum optimization into \methodname{} represents a promising direction. By enhancing the efficiency of finding optimal solutions, these quantum-based techniques have the potential to significantly improve the performance and effectiveness of \methodname{}.

\textbf{Quantum Security for \methodname{}}: Quantum computing-based encryption methods are promising for enhancing the privacy and security of \methodname{}~\cite{ren2023towards}. 
Quantum Key Distribution (QKD) employs quantum properties to create a secure communication channel~\cite{kaewpuang2023adaptive, ren2024icc}. It enables the generation of a shared random secret key for encryption and decryption. Its sensitivity to eavesdropping ensures high security. In addition, device-independent QKD (DI-QKD) offers an even higher level of security by not requiring trust in the quantum devices used in the protocol ~\cite{ren2023qfdsa}. In \methodname{} settings, QKD can securely distribute keys among participants, ensuring secure communication and model updates. 

Quantum encryption methods use quantum states for secure information representation and transmission. They can help \methodname{} protect the model parameters transmitted across the federated network. As quantum adversaries threaten classical encryption methods, quantum-secure cryptography develops new techniques that are secure against them. Its incorporation into \methodname{} is essential for long-term security. Quantum authentication protocols ensure the authenticity of data and participants, which is crucial in \methodname{} settings to defend against malicious activities and maintain integrity. Quantum-secure hash functions can provide data integrity and authentication. They remain secure against quantum attacks, thus enhancing data integrity in \methodname{}~\cite{gurung2023performance}. Quantum homomorphic encryption allows computations on encrypted data without decryption. It can offer privacy protection on transmitted model parameters during the collaborative learning effort in \methodname{}~\cite{buyukates2022lightverifl}. Blind quantum computing enables private quantum computations \cite{barz2012demonstration}. It can be leveraged to allow \methodname{} participants to contribute to the learning processes without revealing their data. 
Quantum secure multi-party computation (SMPC) can extend classical SMPC to the quantum realm for FL without revealing individual data \cite{qu2021secure}, thereby ensuring secure communication in \methodname{}. 

Overall, these secure quantum computing methods can be leveraged to build a comprehensive framework for ensuring unparalleled security and privacy in data encryption and communication in \methodname{} settings.

\textbf{Quantum Entanglement and Quantum Aggregation for \methodname{}}: Quantum entanglement offers a novel approach to synchronizing updates in \methodname{}. This quantum phenomenon, where two or more particles become interconnected so that the state of one instantly influences the others, can be useful for \methodname{} in multiple ways. Theoretically, quantum entanglement might enable immediate synchronization of model updates across different nodes in a federated network \cite{horodecki2009quantum}. Using entangled particles to represent model parameters allows changes in one node to be instantly reflected in all other entangled nodes, ensuring real-time synchronization of updates. Reaching consensus on model updates among distributed nodes is a significant challenge in \methodname{}. 
Quantum entanglement can achieve highly resource-efficient synchronization by reducing communication overhead and allowing for more effective error correction~\cite{troupe2022quantum}. This can save bandwidth and computational resources, with quantum entanglement-assisted aggregation potentially correlating model updates from different nodes more efficiently. While still in the theoretical stage, the potential impact of quantum entanglement on \methodname{} could revolutionize the way model updates are synchronized, with highly enhanced efficiency, security and resource management.

Quantum aggregation in \methodname{} involves leveraging quantum computing principles for aggregating model updates from various nodes. This process aims to compute a global model update that effectively integrates local model changes from each node, potentially enhancing both efficiency and security. Key approaches in quantum aggregation include the follows. Quantum superposition \cite{friedman2000quantum} and quantum interference \cite{romero2011large} involve using quantum superposition to represent multiple model updates on a single quantum state. This state can then be processed to compute a global model updating. Quantum interference, where quantum states interfere constructively or destructively, can facilitate efficient \methodname{} aggregation. Quantum summation algorithms can perform aggregation tasks more efficiently than classical algorithms~\cite{park2019quantum}. Leveraging quantum cryptography techniques, quantum secure aggregation of model updates can preserve privacy while enabling effective aggregation. While detailed studies specific to \methodname{} and quantum aggregation are not yet available, the principles of quantum computing offer promising pathways for enhancing the efficiency and security of \methodname{} model aggregation.

\subsection{Promising Direction: Model Evaluation of \methodname{}}

\textbf{Multimodal Data Evaluation}: 
Existing FL evaluation approaches are generally designed for single-modal data and cannot be directly adopted by \methodname{} where multimodal data which are combinatorial objects (\emph{e.g.}, integers, dates, URL strings, phone numbers) are often involved. The multiple modalities and heterogeneity among data owners make it hard to match samples and conduct model aggregations. Furthermore, when there are noisy samples in certain modalities, how to select high-quality data and train high-performance models via \methodname{} is still an open challenge.

\textbf{Robust Evaluation}: 
It is crucial to maintain robustness against varying inputs in FMs trained via \methodname{}. For example, employing identical prompts with distinct grammars and expressions might result in different outputs from ChatGPT and other LLMs, which indicates the lack of robustness of current LLMs to input variations. Although there exist prior work on robustness evaluation, there is still considerable room for improving, including using more diverse evaluation sets and developing more efficient evaluations to achieve robustness task performance. 

\textbf{Evaluation under Complex Threat Models}:
Existing FL evaluation methods are mostly based on the simple threat model of semi-honest participants. This renders them susceptible to scenarios in which the server or clients are colluding or malicious. 
It is imperative to relax this simplifying assumption to enable future \methodname{} evaluation methods to handle more realistic threats in practice. Moreover, understanding how the adversaries can exploit the interpretations derived from \methodname{} evaluation methods to compromise the system is also crucial for the adoption of \methodname{} by mission-critical applications.

\textbf{Efficiency and Accuracy Trade-offs}: Existing FL evaluation techniques (\emph{e.g.}, shapely value, influence functions) incur high computation and communication costs if directly used to evaluate \methodname{}. This is especially challenging if resource-constrained devices are involved. Therefore, research focusing on trade-off between efficiency and accuracy is very important for \methodname{} evaluation for practical adoption. 

\textbf{Actionable Recommendation based on \methodname{} Evaluation}:
\methodname{} evaluation is not the ultimate goal, but rather a means to its enhancement. 
A powerful evaluation system should not only provide benchmark results, but also deliver insightful analyses, recommendations and guidance for future \methodname{} enhancement. 
With evaluation results, we obtain conclusions regarding model performance, robustness, stability and other factors. It is important to explore adaptive explainable federated learning techniques (e.g., federated neuro-symbolic learning) to provide \methodname{} designers with actionable recommendations on how to enhance \methodname{}.


\section{Conclusions}
\label{sec:conclusion}

In this paper, we provide a comprehensive and perspective survey on the development of \methodname{}, representing a significant step towards building FM-based AI systems in an efficient and collaborative manner. 
The survey reveals a complex yet promising set of topics intersecting FL and FMs. As \methodname{} continue to evolve, the focus on training and aggregation methodologies, trustworthiness measures, incentive mechanisms, and participant management strategies will be crucial for its success. 
This field is essential for unlocking the full potential of FMs by enabling efficient and trustworthy access to privately owned valuable data via FL.
The proposed taxonomy for \methodname{} and novel strategies for participant selection and model training pave the way for future research, particularly in enhancing privacy-preserving techniques and leveraging quantum computing for improved security and efficiency. 
Overall, the convergence of \methodname{} demonstrates significant potential in advancing AI capabilities while effectively addressing critical challenges in data privacy and model trustworthiness.

\bibliographystyle{IEEEtran}

\bibliography{cite}

\begin{thebibliography}{100}
\providecommand{\url}[1]{#1}
\csname url@samestyle\endcsname
\providecommand{\newblock}{\relax}
\providecommand{\bibinfo}[2]{#2}
\providecommand{\BIBentrySTDinterwordspacing}{\spaceskip=0pt\relax}
\providecommand{\BIBentryALTinterwordstretchfactor}{4}
\providecommand{\BIBentryALTinterwordspacing}{\spaceskip=\fontdimen2\font plus
\BIBentryALTinterwordstretchfactor\fontdimen3\font minus \fontdimen4\font\relax}
\providecommand{\BIBforeignlanguage}[2]{{%
\expandafter\ifx\csname l@#1\endcsname\relax
\typeout{** WARNING: IEEEtran.bst: No hyphenation pattern has been}%
\typeout{** loaded for the language `#1'. Using the pattern for}%
\typeout{** the default language instead.}%
\else
\language=\csname l@#1\endcsname
\fi
#2}}
\providecommand{\BIBdecl}{\relax}
\BIBdecl

\bibitem{yuan2023power}
Y.~Yuan, ``On the power of foundation models,'' in \emph{International Conference on Machine Learning}.\hskip 1em plus 0.5em minus 0.4em\relax PMLR, 2023, pp. 40\,519--40\,530.

\bibitem{bommasani2021opportunities}
R.~Bommasani, D.~A. Hudson, E.~Adeli, R.~Altman, S.~Arora, S.~von Arx, M.~S. Bernstein, J.~Bohg, A.~Bosselut, E.~Brunskill \emph{et~al.}, ``On the opportunities and risks of foundation models,'' \emph{arXiv preprint arXiv:2108.07258}, 2021.

\bibitem{zhou2023comprehensive}
C.~Zhou, Q.~Li, C.~Li, J.~Yu, Y.~Liu, G.~Wang, K.~Zhang, C.~Ji, Q.~Yan, L.~He \emph{et~al.}, ``A comprehensive survey on pretrained foundation models: A history from bert to chatgpt,'' \emph{arXiv preprint arXiv:2302.09419}, 2023.

\bibitem{radford2019language}
A.~Radford, J.~Wu, R.~Child, D.~Luan, D.~Amodei, I.~Sutskever \emph{et~al.}, ``Language models are unsupervised multitask learners,'' \emph{OpenAI blog}, vol.~1, no.~8, p.~9, 2019.

\bibitem{brown2020language}
T.~Brown, B.~Mann, N.~Ryder, M.~Subbiah, J.~D. Kaplan, P.~Dhariwal, A.~Neelakantan, P.~Shyam, G.~Sastry, A.~Askell \emph{et~al.}, ``Language models are few-shot learners,'' \emph{Advances in neural information processing systems}, vol.~33, pp. 1877--1901, 2020.

\bibitem{OpenAI_GPT4_2023}
\BIBentryALTinterwordspacing
OpenAI, ``Gpt-4 technical report,'' \emph{ArXiv}, vol. abs/2303.08774, 2023. [Online]. Available: \url{https://arxiv.org/abs/2303.08774}
\BIBentrySTDinterwordspacing

\bibitem{llama}
H.~Touvron, T.~Lavril, G.~Izacard, X.~Martinet, M.-A. Lachaux, T.~Lacroix, B.~Rozi{\`e}re, N.~Goyal, E.~Hambro, F.~Azhar \emph{et~al.}, ``Llama: Open and efficient foundation language models,'' \emph{arXiv preprint arXiv:2302.13971}, 2023.

\bibitem{chowdhery2022palm}
A.~Chowdhery, S.~Narang, J.~Devlin, M.~Bosma, G.~Mishra, A.~Roberts, P.~Barham, H.~W. Chung, C.~Sutton, S.~Gehrmann \emph{et~al.}, ``Palm: Scaling language modeling with pathways,'' \emph{arXiv preprint arXiv:2204.02311}, 2022.

\bibitem{segmentanyting}
A.~Kirillov, E.~Mintun, N.~Ravi, H.~Mao, C.~Rolland, L.~Gustafson, T.~Xiao, S.~Whitehead, A.~C. Berg, W.-Y. Lo \emph{et~al.}, ``Segment anything,'' \emph{arXiv preprint arXiv:2304.02643}, 2023.

\bibitem{stablediffusion}
R.~Rombach, A.~Blattmann, D.~Lorenz, P.~Esser, and B.~Ommer, ``High-resolution image synthesis with latent diffusion models,'' in \emph{Proceedings of the IEEE/CVF conference on computer vision and pattern recognition}, 2022, pp. 10\,684--10\,695.

\bibitem{dalle}
A.~Ramesh, M.~Pavlov, G.~Goh, S.~Gray, C.~Voss, A.~Radford, M.~Chen, and I.~Sutskever, ``Zero-shot text-to-image generation,'' in \emph{International Conference on Machine Learning}.\hskip 1em plus 0.5em minus 0.4em\relax PMLR, 2021, pp. 8821--8831.

\bibitem{clip}
A.~Radford, J.~W. Kim, C.~Hallacy, A.~Ramesh, G.~Goh, S.~Agarwal, G.~Sastry, A.~Askell, P.~Mishkin, J.~Clark \emph{et~al.}, ``Learning transferable visual models from natural language supervision,'' in \emph{International conference on machine learning}.\hskip 1em plus 0.5em minus 0.4em\relax PMLR, 2021, pp. 8748--8763.

\bibitem{kairouz2021advances}
P.~Kairouz, H.~B. McMahan, B.~Avent, A.~Bellet, M.~Bennis, A.~N. Bhagoji, K.~Bonawitz, Z.~Charles, G.~Cormode, R.~Cummings \emph{et~al.}, ``Advances and open problems in federated learning,'' \emph{Foundations and trends{\textregistered} in machine learning}, vol.~14, no. 1--2, pp. 1--210, 2021.

\bibitem{chen2023federated}
C.~Chen, X.~Feng, J.~Zhou, J.~Yin, and X.~Zheng, ``Federated large language model: A position paper,'' \emph{arXiv preprint arXiv:2307.08925}, 2023.

\bibitem{zhuang2023foundation}
W.~Zhuang, C.~Chen, and L.~Lyu, ``When foundation model meets federated learning: Motivations, challenges, and future directions,'' \emph{arXiv preprint arXiv:2306.15546}, 2023.

\bibitem{yu2023federated}
S.~Yu, J.~P. Mu{\~n}oz, and A.~Jannesari, ``Federated foundation models: Privacy-preserving and collaborative learning for large models,'' \emph{arXiv preprint arXiv:2305.11414}, 2023.

\bibitem{kang2023grounding}
Y.~Kang, T.~Fan, H.~Gu, L.~Fan, and Q.~Yang, ``Grounding foundation models through federated transfer learning: A general framework,'' \emph{arXiv preprint arXiv:2311.17431}, 2023.

\bibitem{woisetschlager2024survey}
H.~Woisetschl{\"a}ger, A.~Isenko, S.~Wang, R.~Mayer, and H.-A. Jacobsen, ``A survey on efficient federated learning methods for foundation model training,'' \emph{arXiv preprint arXiv:2401.04472}, 2024.

\bibitem{li2024position}
X.~Li and J.~Wang, ``Position paper: Assessing robustness, privacy, and fairness in federated learning integrated with foundation models,'' \emph{arXiv preprint arXiv:2402.01857}, 2024.

\bibitem{li2024synergizing}
S.~Li, F.~Ye, M.~Fang, J.~Zhao, Y.-H. Chan, E.~C.-H. Ngai, and T.~Voigt, ``Synergizing foundation models and federated learning: A survey,'' \emph{arXiv preprint arXiv:2406.12844}, 2024.

\bibitem{sani2024future}
L.~Sani, A.~Iacob, Z.~Cao, B.~Marino, Y.~Gao, T.~Paulik, W.~Zhao, W.~F. Shen, P.~Aleksandrov, X.~Qiu \emph{et~al.}, ``The future of large language model pre-training is federated,'' \emph{arXiv preprint arXiv:2405.10853}, 2024.

\bibitem{myers2024foundation}
D.~Myers, R.~Mohawesh, V.~I. Chellaboina, A.~L. Sathvik, P.~Venkatesh, Y.-H. Ho, H.~Henshaw, M.~Alhawawreh, D.~Berdik, and Y.~Jararweh, ``Foundation and large language models: fundamentals, challenges, opportunities, and social impacts,'' \emph{Cluster Computing}, vol.~27, no.~1, pp. 1--26, 2024.

\bibitem{li2020federated}
T.~Li, A.~K. Sahu, M.~Zaheer, M.~Sanjabi, A.~Talwalkar, and V.~Smith, ``Federated optimization in heterogeneous networks,'' \emph{Proceedings of Machine learning and systems}, vol.~2, pp. 429--450, 2020.

\bibitem{yang2019federated}
Q.~Yang, Y.~Liu, Y.~Cheng, Y.~Kang, T.~Chen, and H.~Yu, \emph{Federated Learning}.\hskip 1em plus 0.5em minus 0.4em\relax Springer, Cham, 2020, vol. Synthesis Lectures on Artificial Intelligence and Machine Learning.

\bibitem{TransferLearning2010}
S.~J. Pan and Q.~Yang, ``A survey on transfer learning,'' \emph{IEEE Transactions on Knowledge and Data Engineering}, vol.~22, no.~10, pp. 1345--1359, 2010.

\bibitem{villalobos2022will}
P.~Villalobos, J.~Sevilla, L.~Heim, T.~Besiroglu, M.~Hobbhahn, and A.~Ho, ``Will we run out of data? an analysis of the limits of scaling datasets in machine learning,'' \emph{arXiv preprint arXiv:2211.04325}, 2022.

\bibitem{ibrahim2024simple}
A.~Ibrahim, B.~Th{\'e}rien, K.~Gupta, M.~L. Richter, Q.~Anthony, T.~Lesort, E.~Belilovsky, and I.~Rish, ``Simple and scalable strategies to continually pre-train large language models,'' \emph{arXiv preprint arXiv:2403.08763}, 2024.

\bibitem{subramanian2024towards}
S.~Subramanian, P.~Harrington, K.~Keutzer, W.~Bhimji, D.~Morozov, M.~W. Mahoney, and A.~Gholami, ``Towards foundation models for scientific machine learning: Characterizing scaling and transfer behavior,'' \emph{Advances in Neural Information Processing Systems}, vol.~36, 2024.

\bibitem{dosovitskiy2020image}
A.~Dosovitskiy, L.~Beyer, A.~Kolesnikov, D.~Weissenborn, X.~Zhai, T.~Unterthiner, M.~Dehghani, M.~Minderer, G.~Heigold, S.~Gelly \emph{et~al.}, ``An image is worth 16x16 words: Transformers for image recognition at scale,'' \emph{arXiv preprint arXiv:2010.11929}, 2020.

\bibitem{pan2024unifying}
S.~Pan, L.~Luo, Y.~Wang, C.~Chen, J.~Wang, and X.~Wu, ``Unifying large language models and knowledge graphs: A roadmap,'' \emph{IEEE Transactions on Knowledge and Data Engineering}, 2024.

\bibitem{nguyen2022federated}
D.~C. Nguyen, Q.-V. Pham, P.~N. Pathirana, M.~Ding, A.~Seneviratne, Z.~Lin, O.~Dobre, and W.-J. Hwang, ``Federated learning for smart healthcare: A survey,'' \emph{ACM Computing Surveys (Csur)}, vol.~55, no.~3, pp. 1--37, 2022.

\bibitem{long2020federated}
G.~Long, Y.~Tan, J.~Jiang, and C.~Zhang, ``Federated learning for open banking,'' in \emph{Federated Learning: Privacy and Incentive}.\hskip 1em plus 0.5em minus 0.4em\relax Springer, 2020, pp. 240--254.

\bibitem{mcmahan2017communication}
B.~McMahan, E.~Moore, D.~Ramage, S.~Hampson, and B.~A. y~Arcas, ``Communication-efficient learning of deep networks from decentralized data,'' in \emph{Artificial intelligence and statistics}.\hskip 1em plus 0.5em minus 0.4em\relax PMLR, 2017, pp. 1273--1282.

\bibitem{tan2022towards}
A.~Z. Tan, H.~Yu, L.~Cui, and Q.~Yang, ``Towards personalized federated learning,'' \emph{IEEE Transactions on Neural Networks and Learning Systems}, 2022.

\bibitem{scaffold}
S.~P. Karimireddy, S.~Kale, M.~Mohri, S.~Reddi, S.~Stich, and A.~T. Suresh, ``Scaffold: Stochastic controlled averaging for federated learning,'' in \emph{International conference on machine learning}.\hskip 1em plus 0.5em minus 0.4em\relax PMLR, 2020, pp. 5132--5143.

\bibitem{reisizadeh2020fedpaq}
A.~Reisizadeh, A.~Mokhtari, H.~Hassani, A.~Jadbabaie, and R.~Pedarsani, ``Fedpaq: A communication-efficient federated learning method with periodic averaging and quantization,'' in \emph{International Conference on Artificial Intelligence and Statistics}.\hskip 1em plus 0.5em minus 0.4em\relax PMLR, 2020, pp. 2021--2031.

\bibitem{diao2020heterofl}
E.~Diao, J.~Ding, and V.~Tarokh, ``Heterofl: Computation and communication efficient federated learning for heterogeneous clients,'' \emph{arXiv preprint arXiv:2010.01264}, 2020.

\bibitem{li2020lotteryfl}
A.~Li, J.~Sun, B.~Wang, L.~Duan, S.~Li, Y.~Chen, and H.~Li, ``Lotteryfl: Personalized and communication-efficient federated learning with lottery ticket hypothesis on non-iid datasets,'' \emph{arXiv preprint arXiv:2008.03371}, 2020.

\bibitem{horvath2021fjord}
S.~Horvath, S.~Laskaridis, M.~Almeida, I.~Leontiadis, S.~Venieris, and N.~Lane, ``Fjord: Fair and accurate federated learning under heterogeneous targets with ordered dropout,'' \emph{Advances in Neural Information Processing Systems}, vol.~34, pp. 12\,876--12\,889, 2021.

\bibitem{yang2021h-fl}
H.~Yang, ``H-fl: A hierarchical communication-efficient and privacy-protected architecture for federated learning,'' \emph{arXiv preprint arXiv:2106.00275}, 2021.

\bibitem{jiang2022pruneFL}
Y.~Jiang, S.~Wang, V.~Valls, B.~J. Ko, W.-H. Lee, K.~K. Leung, and L.~Tassiulas, ``Model pruning enables efficient federated learning on edge devices,'' \emph{IEEE Transactions on Neural Networks and Learning Systems}, 2022.

\bibitem{isik2022fedpm}
B.~Isik, F.~Pase, D.~Gunduz, T.~Weissman, and M.~Zorzi, ``Sparse random networks for communication-efficient federated learning,'' \emph{arXiv preprint arXiv:2209.15328}, 2022.

\bibitem{huang2023fedtiny}
H.~Huang, L.~Zhang, C.~Sun, R.~Fang, X.~Yuan, and D.~Wu, ``Distributed pruning towards tiny neural networks in federated learning,'' in \emph{2023 IEEE 43rd International Conference on Distributed Computing Systems (ICDCS)}.\hskip 1em plus 0.5em minus 0.4em\relax IEEE, 2023, pp. 190--201.

\bibitem{li2022soteriafl}
Z.~Li, H.~Zhao, B.~Li, and Y.~Chi, ``Soteriafl: A unified framework for private federated learning with communication compression,'' \emph{Advances in Neural Information Processing Systems}, vol.~35, pp. 4285--4300, 2022.

\bibitem{zhao2023fedprompt}
H.~Zhao, W.~Du, F.~Li, P.~Li, and G.~Liu, ``Fedprompt: Communication-efficient and privacy-preserving prompt tuning in federated learning,'' in \emph{ICASSP 2023-2023 IEEE International Conference on Acoustics, Speech and Signal Processing (ICASSP)}.\hskip 1em plus 0.5em minus 0.4em\relax IEEE, 2023, pp. 1--5.

\bibitem{tian2022fedbert}
Y.~Tian, Y.~Wan, L.~Lyu, D.~Yao, H.~Jin, and L.~Sun, ``Fedbert: When federated learning meets pre-training,'' \emph{ACM Transactions on Intelligent Systems and Technology (TIST)}, vol.~13, no.~4, pp. 1--26, 2022.

\bibitem{lu2023fedclip}
W.~Lu, X.~Hu, J.~Wang, and X.~Xie, ``Fedclip: Fast generalization and personalization for clip in federated learning,'' \emph{arXiv preprint arXiv:2302.13485}, 2023.

\bibitem{fedpeft2022}
G.~Sun, M.~Mendieta, T.~Yang, and C.~Chen, ``Exploring parameter-efficient fine-tuning for improving communication efficiency in federated learning,'' \emph{arXiv preprint arXiv:2210.01708}, 2022.

\bibitem{zhang2023fedpetuning}
Z.~Zhang, Y.~Yang, Y.~Dai, Q.~Wang, Y.~Yu, L.~Qu, and Z.~Xu, ``Fedpetuning: When federated learning meets the parameter-efficient tuning methods of pre-trained language models,'' in \emph{Annual Meeting of the Association of Computational Linguistics 2023}.\hskip 1em plus 0.5em minus 0.4em\relax Association for Computational Linguistics (ACL), 2023, pp. 9963--9977.

\bibitem{babakniya2023slora}
S.~Babakniya, A.~R. Elkordy, Y.~H. Ezzeldin, Q.~Liu, K.-B. Song, M.~El-Khamy, and S.~Avestimehr, ``Slora: Federated parameter efficient fine-tuning of language models,'' \emph{arXiv preprint arXiv:2308.06522}, 2023.

\bibitem{chen2022fedobd}
Y.~Chen, Z.~Chen, P.~Wu, and H.~Yu, ``Fedobd: Opportunistic block dropout for efficiently training large-scale neural networks through federated learning,'' \emph{arXiv preprint arXiv:2208.05174}, 2022.

\bibitem{zhang2023fedit}
J.~Zhang, S.~Vahidian, M.~Kuo, C.~Li, R.~Zhang, G.~Wang, and Y.~Chen, ``Towards building the federated gpt: Federated instruction tuning,'' \emph{arXiv preprint arXiv:2305.05644}, 2023.

\bibitem{xu2023fwdllm}
M.~Xu, Y.~Wu, D.~Cai, X.~Li, and S.~Wang, ``Fwdllm: Efficient fedllm using forward gradient,'' \emph{arXiv preprint arXiv:2308.13894}, 2023.

\bibitem{devlin2018bert}
J.~Devlin, M.-W. Chang, K.~Lee, and K.~Toutanova, ``Bert: Pre-training of deep bidirectional transformers for language understanding,'' \emph{arXiv preprint arXiv:1810.04805}, 2018.

\bibitem{radford2018improving}
A.~Radford, K.~Narasimhan, T.~Salimans, and I.~Sutskever, ``Improving language understanding with unsupervised learning,'' \emph{Open AI}, 2018.

\bibitem{dosovitskiy2020ViT}
A.~Dosovitskiy, L.~Beyer, A.~Kolesnikov, D.~Weissenborn, X.~Zhai, T.~Unterthiner, M.~Dehghani, M.~Minderer, G.~Heigold, S.~Gelly \emph{et~al.}, ``An image is worth 16x16 words: Transformers for image recognition at scale,'' \emph{arXiv preprint arXiv:2010.11929}, 2020.

\bibitem{raffel2020exploring}
C.~Raffel, N.~Shazeer, A.~Roberts, K.~Lee, S.~Narang, M.~Matena, Y.~Zhou, W.~Li, and P.~J. Liu, ``Exploring the limits of transfer learning with a unified text-to-text transformer,'' \emph{Journal of machine learning research}, vol.~21, no. 140, pp. 1--67, 2020.

\bibitem{ramesh2021zero}
A.~Ramesh, M.~Pavlov, G.~Goh, S.~Gray, C.~Voss, A.~Radford, M.~Chen, and I.~Sutskever, ``Zero-shot text-to-image generation,'' in \emph{International conference on machine learning}.\hskip 1em plus 0.5em minus 0.4em\relax Pmlr, 2021, pp. 8821--8831.

\bibitem{radford2021learning}
A.~Radford, J.~W. Kim, C.~Hallacy, A.~Ramesh, G.~Goh, S.~Agarwal, G.~Sastry, A.~Askell, P.~Mishkin, J.~Clark \emph{et~al.}, ``Learning transferable visual models from natural language supervision,'' in \emph{International conference on machine learning}.\hskip 1em plus 0.5em minus 0.4em\relax PMLR, 2021, pp. 8748--8763.

\bibitem{reed2022generalist}
S.~Reed, K.~Zolna, E.~Parisotto, S.~G. Colmenarejo, A.~Novikov, G.~Barth-Maron, M.~Gimenez, Y.~Sulsky, J.~Kay, J.~T. Springenberg \emph{et~al.}, ``A generalist agent,'' \emph{arXiv preprint arXiv:2205.06175}, 2022.

\bibitem{chowdhery2023palm}
A.~Chowdhery, S.~Narang, J.~Devlin, M.~Bosma, G.~Mishra, A.~Roberts, P.~Barham, H.~W. Chung, C.~Sutton, S.~Gehrmann \emph{et~al.}, ``Palm: Scaling language modeling with pathways,'' \emph{Journal of Machine Learning Research}, vol.~24, no. 240, pp. 1--113, 2023.

\bibitem{touvron2023llama}
H.~Touvron, T.~Lavril, G.~Izacard, X.~Martinet, M.-A. Lachaux, T.~Lacroix, B.~Rozi{\`e}re, N.~Goyal, E.~Hambro, F.~Azhar \emph{et~al.}, ``Llama: Open and efficient foundation language models,'' \emph{arXiv preprint arXiv:2302.13971}, 2023.

\bibitem{GPT-3.5}
OpenAI, ``Gpt-3.5,'' 2023.

\bibitem{hoffmann2022training}
J.~Hoffmann, S.~Borgeaud, A.~Mensch, E.~Buchatskaya, T.~Cai, E.~Rutherford, D.~de~Las~Casas, L.~A. Hendricks, J.~Welbl, A.~Clark \emph{et~al.}, ``Training compute-optimal large language models,'' in \emph{Proceedings of the 36th International Conference on Neural Information Processing Systems}, 2022, pp. 30\,016--30\,030.

\bibitem{alayrac2022flamingo}
J.-B. Alayrac, J.~Donahue, P.~Luc, A.~Miech, I.~Barr, Y.~Hasson, K.~Lenc, A.~Mensch, K.~Millican, M.~Reynolds \emph{et~al.}, ``Flamingo: a visual language model for few-shot learning,'' \emph{Advances in neural information processing systems}, vol.~35, pp. 23\,716--23\,736, 2022.

\bibitem{liu2024visual}
H.~Liu, C.~Li, Q.~Wu, and Y.~J. Lee, ``Visual instruction tuning,'' \emph{Advances in neural information processing systems}, vol.~36, 2024.

\bibitem{LLaMA-3}
\BIBentryALTinterwordspacing
M.~AI, ``Llama 3,'' 2024. [Online]. Available: \url{https://llama.meta.com/docs/model-cards-and-prompt-formats/meta-llama-3}
\BIBentrySTDinterwordspacing

\bibitem{GPT-4o}
\BIBentryALTinterwordspacing
OpenAI, ``Gpt-4o,'' 2024. [Online]. Available: \url{https://platform.openai.com/docs/models/gpt-4o}
\BIBentrySTDinterwordspacing

\bibitem{LLaMA-3.1}
\BIBentryALTinterwordspacing
M.~AI, ``Llama 3.1,'' 2024. [Online]. Available: \url{https://llama.meta.com/docs/model-cards-and-prompt-formats/llama3_1}
\BIBentrySTDinterwordspacing

\bibitem{liu2023recent}
B.~Liu, N.~Lv, Y.~Guo, and Y.~Li, ``Recent advances on federated learning: A systematic survey,'' \emph{arXiv preprint arXiv:2301.01299}, 2023.

\bibitem{wang2020fedma}
H.~Wang, M.~Yurochkin, Y.~Sun, D.~Papailiopoulos, and Y.~Khazaeni, ``Federated learning with matched averaging,'' \emph{arXiv preprint arXiv:2002.06440}, 2020.

\bibitem{ye2024openfedllm}
R.~Ye, W.~Wang, J.~Chai, D.~Li, Z.~Li, Y.~Xu, Y.~Du, Y.~Wang, and S.~Chen, ``Openfedllm: Training large language models on decentralized private data via federated learning,'' \emph{arXiv preprint arXiv:2402.06954}, 2024.

\bibitem{wortsman2022model}
M.~Wortsman, G.~Ilharco, S.~Y. Gadre, R.~Roelofs, R.~Gontijo-Lopes, A.~S. Morcos, H.~Namkoong, A.~Farhadi, Y.~Carmon, S.~Kornblith \emph{et~al.}, ``Model soups: averaging weights of multiple fine-tuned models improves accuracy without increasing inference time,'' in \emph{International Conference on Machine Learning}.\hskip 1em plus 0.5em minus 0.4em\relax PMLR, 2022, pp. 23\,965--23\,998.

\bibitem{pmlr-v202-rame23a}
\BIBentryALTinterwordspacing
A.~Rame, K.~Ahuja, J.~Zhang, M.~Cord, L.~Bottou, and D.~Lopez-Paz, ``Model ratatouille: Recycling diverse models for out-of-distribution generalization,'' in \emph{Proceedings of the 40th International Conference on Machine Learning}, ser. Proceedings of Machine Learning Research, A.~Krause, E.~Brunskill, K.~Cho, B.~Engelhardt, S.~Sabato, and J.~Scarlett, Eds., vol. 202.\hskip 1em plus 0.5em minus 0.4em\relax PMLR, 23--29 Jul 2023, pp. 28\,656--28\,679. [Online]. Available: \url{https://proceedings.mlr.press/v202/rame23a.html}
\BIBentrySTDinterwordspacing

\bibitem{MoE}
W.~Fedus, B.~Zoph, and N.~Shazeer, ``Switch transformers: Scaling to trillion parameter models with simple and efficient sparsity,'' \emph{J. Mach. Learn. Res.}, vol.~23, pp. 120:1--120:39, 2022.

\bibitem{du2022glam}
N.~Du, Y.~Huang, A.~M. Dai, S.~Tong, D.~Lepikhin, Y.~Xu, M.~Krikun, Y.~Zhou, A.~W. Yu, O.~Firat \emph{et~al.}, ``Glam: Efficient scaling of language models with mixture-of-experts,'' in \emph{International Conference on Machine Learning}.\hskip 1em plus 0.5em minus 0.4em\relax PMLR, 2022, pp. 5547--5569.

\bibitem{zoph2022st}
B.~Zoph, I.~Bello, S.~Kumar, N.~Du, Y.~Huang, J.~Dean, N.~Shazeer, and W.~Fedus, ``St-moe: Designing stable and transferable sparse expert models,'' \emph{arXiv preprint arXiv:2202.08906}, 2022.

\bibitem{xue2024openmoe}
F.~Xue, Z.~Zheng, Y.~Fu, J.~Ni, Z.~Zheng, W.~Zhou, and Y.~You, ``Openmoe: An early effort on open mixture-of-experts language models,'' \emph{arXiv preprint arXiv:2402.01739}, 2024.

\bibitem{MoE-FL}
S.~Parsaeefard, S.~E. Etesami, and A.~Leon{-}Garcia, ``Robust federated learning by mixture of experts,'' \emph{CoRR}, vol. abs/2104.11700, 2021.

\bibitem{FedMix}
E.~L. Zec, O.~Mogren, J.~Martinsson, L.~R. S{\"{u}}tfeld, and D.~Gillblad, ``Federated learning using a mixture of experts,'' \emph{CoRR}, vol. abs/2010.02056, 2020.

\bibitem{Fed-MoEs}
T.~Bai, Y.~Zhang, Y.~Wang, Y.~Qin, and F.~Zhang, ``Multi-site {MRI} classification using weighted federated learning based on mixture of experts domain adaptation,'' in \emph{Proc. {IEEE} {BIBM}}.\hskip 1em plus 0.5em minus 0.4em\relax Las Vegas, NV, USA: {IEEE}, 2022, pp. 916--921.

\bibitem{PFL-MoE}
B.~Guo, Y.~Mei, D.~Xiao, and W.~Wu, ``Pfl-moe: Personalized federated learning based on mixture of experts,'' in \emph{Proc. {APWeb-WAIM}}, vol. 12858.\hskip 1em plus 0.5em minus 0.4em\relax Guangzhou, China: Springer, 2021, pp. 480--486.

\bibitem{pFedMoE}
L.~Yi, H.~Yu, C.~Ren, H.~Zhang, G.~Wang, X.~Liu, and X.~Li, ``pfedmoe: Data-level personalization with mixture of experts for model-heterogeneous personalized federated learning,'' \emph{CoRR}, vol. abs/2402.01350, 2024.

\bibitem{ding2023parameter}
N.~Ding, Y.~Qin, G.~Yang, F.~Wei, Z.~Yang, Y.~Su, S.~Hu, Y.~Chen, C.-M. Chan, W.~Chen \emph{et~al.}, ``Parameter-efficient fine-tuning of large-scale pre-trained language models,'' \emph{Nature Machine Intelligence}, vol.~5, no.~3, pp. 220--235, 2023.

\bibitem{houlsby2019adaptor}
N.~Houlsby, A.~Giurgiu, S.~Jastrzebski, B.~Morrone, Q.~De~Laroussilhe, A.~Gesmundo, M.~Attariyan, and S.~Gelly, ``Parameter-efficient transfer learning for nlp,'' in \emph{International Conference on Machine Learning}.\hskip 1em plus 0.5em minus 0.4em\relax PMLR, 2019, pp. 2790--2799.

\bibitem{zaken2021bitfit}
E.~B. Zaken, S.~Ravfogel, and Y.~Goldberg, ``Bitfit: Simple parameter-efficient fine-tuning for transformer-based masked language-models,'' \emph{arXiv preprint arXiv:2106.10199}, 2021.

\bibitem{hu2021lora}
E.~J. Hu, Y.~Shen, P.~Wallis, Z.~Allen{-}Zhu, Y.~Li, S.~Wang, L.~Wang, and W.~Chen, ``Lora: Low-rank adaptation of large language models,'' in \emph{Proc. {ICLR}}.\hskip 1em plus 0.5em minus 0.4em\relax Virtual Event: OpenReview.net, 2022.

\bibitem{pFedLoRA}
L.~Yi, H.~Yu, G.~Wang, and X.~Liu, ``Fedlora: Model-heterogeneous personalized federated learning with lora tuning,'' \emph{CoRR}, vol. abs/2310.13283, 2023.

\bibitem{SA-FedLoRA}
Y.~Yang, X.~Liu, T.~Gao, X.~Xu, and G.~Wang, ``Sa-fedlora: Adaptive parameter allocation for efficient federated learning with lora tuning,'' \emph{CoRR}, vol. abs/2405.09394, 2024.

\bibitem{FFA-LoRA}
Y.~Sun, Z.~Li, Y.~Li, and B.~Ding, ``Improving lora in privacy-preserving federated learning,'' \emph{CoRR}, vol. abs/2403.12313, 2024.

\bibitem{chen2024feddat}
H.~Chen, Y.~Zhang, D.~Krompass, J.~Gu, and V.~Tresp, ``Feddat: An approach for foundation model finetuning in multi-modal heterogeneous federated learning,'' in \emph{Proceedings of the AAAI Conference on Artificial Intelligence}, vol.~38, no.~10, 2024, pp. 11\,285--11\,293.

\bibitem{su2023fedra}
S.~Su, B.~Li, and X.~Xue, ``Fedra: A random allocation strategy for federated tuning to unleash the power of heterogeneous clients,'' \emph{arXiv preprint arXiv:2311.11227}, 2023.

\bibitem{bai2024federated}
J.~Bai, D.~Chen, B.~Qian, L.~Yao, and Y.~Li, ``Federated fine-tuning of large language models under heterogeneous language tasks and client resources,'' \emph{arXiv preprint arXiv:2402.11505}, 2024.

\bibitem{cho2023heterogeneous}
Y.~J. Cho, L.~Liu, Z.~Xu, A.~Fahrezi, M.~Barnes, and G.~Joshi, ``Heterogeneous lora for federated fine-tuning of on-device foundation models,'' in \emph{International Workshop on Federated Learning in the Age of Foundation Models in Conjunction with NeurIPS 2023}, 2023.

\bibitem{min2023recentPt}
B.~Min, H.~Ross, E.~Sulem, A.~P.~B. Veyseh, T.~H. Nguyen, O.~Sainz, E.~Agirre, I.~Heintz, and D.~Roth, ``Recent advances in natural language processing via large pre-trained language models: A survey,'' \emph{ACM Computing Surveys}, vol.~56, no.~2, pp. 1--40, 2023.

\bibitem{yuan2021bartscore}
W.~Yuan, G.~Neubig, and P.~Liu, ``Bartscore: Evaluating generated text as text generation,'' \emph{Advances in Neural Information Processing Systems}, vol.~34, pp. 27\,263--27\,277, 2021.

\bibitem{scao2021many}
T.~L. Scao and A.~M. Rush, ``How many data points is a prompt worth?'' \emph{arXiv preprint arXiv:2103.08493}, 2021.

\bibitem{zhang2023instruction}
S.~Zhang, L.~Dong, X.~Li, S.~Zhang, X.~Sun, S.~Wang, J.~Li, R.~Hu, T.~Zhang, F.~Wu \emph{et~al.}, ``Instruction tuning for large language models: A survey,'' \emph{arXiv preprint arXiv:2308.10792}, 2023.

\bibitem{cheng2017surveycompression}
Y.~Cheng, D.~Wang, P.~Zhou, and T.~Zhang, ``A survey of model compression and acceleration for deep neural networks,'' \emph{arXiv preprint arXiv:1710.09282}, 2017.

\bibitem{klema1980singular}
V.~Klema and A.~Laub, ``The singular value decomposition: Its computation and some applications,'' \emph{IEEE Transactions on automatic control}, vol.~25, no.~2, pp. 164--176, 1980.

\bibitem{Lyu-et-al:2022TNNLS}
L.~Lyu, H.~Yu, X.~Ma, C.~Chen, L.~Sun, J.~Zhao, Q.~Yang, and P.~S. Yu, ``Privacy and robustness in federated learning: Attacks and defenses,'' \emph{IEEE Transactions on Neural Networks and Learning Systems}, 2022.

\bibitem{BadNets:IdentifyingVulnerabilitiesintheMachineLearningModel}
T.~Gu, B.~Dolan{-}Gavitt, and S.~Garg, ``Badnets: Identifying vulnerabilities in the machine learning model supply chain,'' \emph{CoRR}, vol. abs/1708.06733, 2017.

\bibitem{DBLP:journals/compsec/HeSXHTZ24}
Y.~He, Z.~Shen, C.~Xia, J.~Hua, W.~Tong, and S.~Zhong, ``{SGBA:} {A} stealthy scapegoat backdoor attack against deep neural networks,'' \emph{Comput. Secur.}, vol. 136, p. 103523, 2024.

\bibitem{DBLP:journals/www/HouHYKT23}
R.~Hou, T.~Huang, H.~Yan, L.~Ke, and W.~Tang, ``A stealthy and robust backdoor attack via frequency domain transform,'' \emph{World Wide Web {(WWW)}}, vol.~26, no.~5, pp. 2767--2783, 2023.

\bibitem{DBLP:conf/aistats/BagdasaryanVHES20}
E.~Bagdasaryan, A.~Veit, Y.~Hua, D.~Estrin, and V.~Shmatikov, ``How to backdoor federated learning,'' in \emph{The 23rd International Conference on Artificial Intelligence and Statistics, {AISTATS} 2020, 26-28 August 2020, Online [Palermo, Sicily, Italy]}, ser. Proceedings of Machine Learning Research, vol. 108.\hskip 1em plus 0.5em minus 0.4em\relax {PMLR}, 2020, pp. 2938--2948.

\bibitem{DBLP:conf/aaai/LyuHWLWL023}
X.~Lyu, Y.~Han, W.~Wang, J.~Liu, B.~Wang, J.~Liu, and X.~Zhang, ``Poisoning with cerberus: Stealthy and colluded backdoor attack against federated learning,'' in \emph{Thirty-Seventh {AAAI} Conference on Artificial Intelligence, {AAAI} 2023}.\hskip 1em plus 0.5em minus 0.4em\relax {AAAI} Press, 2023, pp. 9020--9028.

\bibitem{PoisoningAttacksagainstSupportVectorMachines}
B.~Biggio, B.~Nelson, and P.~Laskov, ``Poisoning attacks against support vector machines,'' in \emph{Proceedings of the 29th International Conference on Machine Learning, {ICML} 2012, Edinburgh, Scotland, UK, June 26 - July 1, 2012}.\hskip 1em plus 0.5em minus 0.4em\relax icml.cc / Omnipress, 2012.

\bibitem{TargetedBackdoorAttacksonDeepLearning}
\BIBentryALTinterwordspacing
X.~Chen, C.~Liu, B.~Li, K.~Lu, and D.~Song, ``Targeted backdoor attacks on deep learning systems using data poisoning,'' \emph{CoRR}, vol. abs/1712.05526, 2017. [Online]. Available: \url{http://arxiv.org/abs/1712.05526}
\BIBentrySTDinterwordspacing

\bibitem{PoisoningAttackstoGraph-BasedRecommenderSystems}
M.~Fang, G.~Yang, N.~Z. Gong, and J.~Liu, ``Poisoning attacks to graph-based recommender systems,'' in \emph{Proceedings of the 34th Annual Computer Security Applications Conference, {ACSAC} 2018, San Juan, PR, USA, December 03-07, 2018}.\hskip 1em plus 0.5em minus 0.4em\relax {ACM}, 2018, pp. 381--392.

\bibitem{ManipulatingMachineLearning:Poisoning}
M.~Jagielski, A.~Oprea, B.~Biggio, C.~Liu, C.~Nita{-}Rotaru, and B.~Li, ``Manipulating machine learning: Poisoning attacks and countermeasures for regression learning,'' in \emph{2018 {IEEE} Symposium on Security and Privacy, {SP} 2018, Proceedings, 21-23 May 2018, San Francisco, California, {USA}}.\hskip 1em plus 0.5em minus 0.4em\relax {IEEE} Computer Society, 2018, pp. 19--35.

\bibitem{PoisonFrogs!TargetedClean-LabelPoisoning}
A.~Shafahi, W.~R. Huang, M.~Najibi, O.~Suciu, C.~Studer, T.~Dumitras, and T.~Goldstein, ``Poison frogs! targeted clean-label poisoning attacks on neural networks,'' in \emph{Advances in Neural Information Processing Systems 31: Annual Conference on Neural Information Processing Systems 2018, NeurIPS 2018, December 3-8, 2018, Montr{\'{e}}al, Canada}, 2018, pp. 6106--6116.

\bibitem{DBLP:conf/acl/MeiLWZM23}
K.~Mei, Z.~Li, Z.~Wang, Y.~Zhang, and S.~Ma, ``{NOTABLE:} transferable backdoor attacks against prompt-based {NLP} models,'' in \emph{Proceedings of the 61st Annual Meeting of the Association for Computational Linguistics (Volume 1: Long Papers), {ACL} 2023, Toronto, Canada, July 9-14, 2023}.\hskip 1em plus 0.5em minus 0.4em\relax Association for Computational Linguistics, 2023, pp. 15\,551--15\,565.

\bibitem{DBLP:conf/emnlp/ZhaoWLZF23}
S.~Zhao, J.~Wen, A.~T. Luu, J.~Zhao, and J.~Fu, ``Prompt as triggers for backdoor attack: Examining the vulnerability in language models,'' in \emph{Proceedings of the 2023 Conference on Empirical Methods in Natural Language Processing, {EMNLP} 2023, Singapore, December 6-10, 2023}.\hskip 1em plus 0.5em minus 0.4em\relax Association for Computational Linguistics, 2023, pp. 12\,303--12\,317.

\bibitem{DBLP:conf/ijcai/DuZLLW22}
W.~Du, Y.~Zhao, B.~Li, G.~Liu, and S.~Wang, ``{PPT:} backdoor attacks on pre-trained models via poisoned prompt tuning,'' in \emph{Proceedings of the Thirty-First International Joint Conference on Artificial Intelligence, {IJCAI} 2022, Vienna, Austria, 23-29 July 2022}.\hskip 1em plus 0.5em minus 0.4em\relax ijcai.org, 2022, pp. 680--686.

\bibitem{DBLP:conf/sp/WangYSLVZZ19}
B.~Wang, Y.~Yao, S.~Shan, H.~Li, B.~Viswanath, H.~Zheng, and B.~Y. Zhao, ``Neural cleanse: Identifying and mitigating backdoor attacks in neural networks,'' in \emph{2019 {IEEE} Symposium on Security and Privacy, {SP} 2019, San Francisco, CA, USA, May 19-23, 2019}.\hskip 1em plus 0.5em minus 0.4em\relax {IEEE}, 2019, pp. 707--723.

\bibitem{DBLP:conf/cvpr/0002T0SXL0M023}
S.~Feng, G.~Tao, S.~Cheng, G.~Shen, X.~Xu, Y.~Liu, K.~Zhang, S.~Ma, and X.~Zhang, ``Detecting backdoors in pre-trained encoders,'' in \emph{{IEEE/CVF} Conference on Computer Vision and Pattern Recognition, {CVPR} 2023, Vancouver, BC, Canada, June 17-24, 2023}.\hskip 1em plus 0.5em minus 0.4em\relax {IEEE}, 2023, pp. 16\,352--16\,362.

\bibitem{Machine_Learning_with_Adversaries}
P.~Blanchard, E.~M.~E. Mhamdi, R.~Guerraoui, and J.~Stainer, ``Machine learning with adversaries: Byzantine tolerant gradient descent,'' in \emph{Advances in Neural Information Processing Systems 30: Annual Conference on Neural Information Processing Systems 2017, December 4-9, 2017, Long Beach, CA, {USA}}, 2017, pp. 119--129.

\bibitem{Byzantine-RobustDistributedLearning:Towards}
D.~Yin, Y.~Chen, K.~Ramchandran, and P.~L. Bartlett, ``Byzantine-robust distributed learning: Towards optimal statistical rates,'' in \emph{Proceedings of the 35th International Conference on Machine Learning, {ICML} 2018, Stockholmsm{\"{a}}ssan, Stockholm, Sweden, July 10-15, 2018}, ser. Proceedings of Machine Learning Research, vol.~80.\hskip 1em plus 0.5em minus 0.4em\relax {PMLR}, 2018, pp. 5636--5645.

\bibitem{DBLP:conf/icml/XieKG19}
C.~Xie, S.~Koyejo, and I.~Gupta, ``Zeno: Distributed stochastic gradient descent with suspicion-based fault-tolerance,'' in \emph{Proceedings of the 36th International Conference on Machine Learning, {ICML} 2019, 9-15 June 2019, Long Beach, California, {USA}}, ser. Proceedings of Machine Learning Research, vol.~97.\hskip 1em plus 0.5em minus 0.4em\relax {PMLR}, 2019, pp. 6893--6901.

\bibitem{DBLP:conf/icml/XieKG20}
------, ``Zeno++: Robust fully asynchronous {SGD},'' in \emph{Proceedings of the 37th International Conference on Machine Learning, {ICML} 2020, 13-18 July 2020, Virtual Event}, ser. Proceedings of Machine Learning Research, vol. 119.\hskip 1em plus 0.5em minus 0.4em\relax {PMLR}, 2020, pp. 10\,495--10\,503.

\bibitem{LocalModelPoisoningAttacksto}
M.~Fang, X.~Cao, J.~Jia, and N.~Z. Gong, ``Local model poisoning attacks to byzantine-robust federated learning,'' in \emph{29th {USENIX} Security Symposium, {USENIX} Security 2020, August 12-14, 2020}.\hskip 1em plus 0.5em minus 0.4em\relax {USENIX} Association, 2020, pp. 1605--1622.

\bibitem{DBLP:conf/acsac/HaoLXC021}
M.~Hao, H.~Li, G.~Xu, H.~Chen, and T.~Zhang, ``Efficient, private and robust federated learning,'' in \emph{{ACSAC} '21: Annual Computer Security Applications Conference, Virtual Event, USA, December 6 - 10, 2021}.\hskip 1em plus 0.5em minus 0.4em\relax {ACM}, 2021, pp. 45--60.

\bibitem{DBLP:conf/ndss/CaoF0G21}
X.~Cao, M.~Fang, J.~Liu, and N.~Z. Gong, ``Fltrust: Byzantine-robust federated learning via trust bootstrapping,'' in \emph{28th Annual Network and Distributed System Security Symposium, {NDSS} 2021, virtually, February 21-25, 2021}.\hskip 1em plus 0.5em minus 0.4em\relax The Internet Society, 2021.

\bibitem{prakash2020mitigating}
S.~Prakash and A.~S. Avestimehr, ``Mitigating byzantine attacks in federated learning,'' \emph{arXiv preprint arXiv:2010.07541}, 2020.

\bibitem{shokri2017membership}
R.~Shokri, M.~Stronati, C.~Song, and V.~Shmatikov, ``Membership inference attacks against machine learning models,'' in \emph{2017 IEEE symposium on security and privacy (SP)}.\hskip 1em plus 0.5em minus 0.4em\relax IEEE, 2017, pp. 3--18.

\bibitem{DBLP:journals/tdsc/LiuWLPW23}
L.~Liu, Y.~Wang, G.~Liu, K.~Peng, and C.~Wang, ``Membership inference attacks against machine learning models via prediction sensitivity,'' \emph{{IEEE} Trans. Dependable Secur. Comput.}, vol.~20, no.~3, pp. 2341--2347, 2023.

\bibitem{DBLP:journals/tdsc/YanLWZSHL23}
H.~Yan, S.~Li, Y.~Wang, Y.~Zhang, K.~Sharif, H.~Hu, and Y.~Li, ``Membership inference attacks against deep learning models via logits distribution,'' \emph{{IEEE} Trans. Dependable Secur. Comput.}, vol.~20, no.~5, pp. 3799--3808, 2023.

\bibitem{DBLP:conf/icml/DuanK0SX23}
J.~Duan, F.~Kong, S.~Wang, X.~Shi, and K.~Xu, ``Are diffusion models vulnerable to membership inference attacks?'' in \emph{International Conference on Machine Learning, {ICML} 2023, 23-29 July 2023, Honolulu, Hawaii, {USA}}, ser. Proceedings of Machine Learning Research, A.~Krause, E.~Brunskill, K.~Cho, B.~Engelhardt, S.~Sabato, and J.~Scarlett, Eds., vol. 202.\hskip 1em plus 0.5em minus 0.4em\relax {PMLR}, 2023, pp. 8717--8730.

\bibitem{DBLP:journals/corr/abs-2311-06062}
W.~Fu, H.~Wang, C.~Gao, G.~Liu, Y.~Li, and T.~Jiang, ``Practical membership inference attacks against fine-tuned large language models via self-prompt calibration,'' \emph{CoRR}, vol. abs/2311.06062, 2023.

\bibitem{DBLP:conf/nips/ZhuLH19}
L.~Zhu, Z.~Liu, and S.~Han, ``Deep leakage from gradients,'' in \emph{Advances in Neural Information Processing Systems 32: Annual Conference on Neural Information Processing Systems 2019, NeurIPS 2019, December 8-14, 2019, Vancouver, BC, Canada}, H.~M. Wallach, H.~Larochelle, A.~Beygelzimer, F.~d'Alch{\'{e}}{-}Buc, E.~B. Fox, and R.~Garnett, Eds., 2019, pp. 14\,747--14\,756.

\bibitem{DBLP:journals/corr/abs-2001-02610}
B.~Zhao, K.~R. Mopuri, and H.~Bilen, ``idlg: Improved deep leakage from gradients,'' \emph{CoRR}, vol. abs/2001.02610, 2020.

\bibitem{DBLP:conf/aaai/YuanCZ0YZ23}
X.~Yuan, K.~Chen, J.~Zhang, W.~Zhang, N.~Yu, and Y.~Zhang, ``Pseudo label-guided model inversion attack via conditional generative adversarial network,'' in \emph{Thirty-Seventh {AAAI} Conference on Artificial Intelligence, {AAAI} 2023}.\hskip 1em plus 0.5em minus 0.4em\relax {AAAI} Press, 2023, pp. 3349--3357.

\bibitem{DBLP:journals/corr/abs-2311-09127}
Y.~Wu, X.~Li, Y.~Liu, P.~Zhou, and L.~Sun, ``Jailbreaking {GPT-4V} via self-adversarial attacks with system prompts,'' \emph{CoRR}, vol. abs/2311.09127, 2023.

\bibitem{DBLP:journals/corr/abs-2310-02446}
Z.~X. Yong, C.~Menghini, and S.~H. Bach, ``Low-resource languages jailbreak {GPT-4},'' \emph{CoRR}, vol. abs/2310.02446, 2023.

\bibitem{DBLP:journals/corr/abs-2307-08715}
G.~Deng, Y.~Liu, Y.~Li, K.~Wang, Y.~Zhang, Z.~Li, H.~Wang, T.~Zhang, and Y.~Liu, ``Jailbreaker: Automated jailbreak across multiple large language model chatbots,'' \emph{CoRR}, vol. abs/2307.08715, 2023.

\bibitem{DBLP:journals/tifs/XueXZLZSL24}
R.~Xue, K.~Xue, B.~Zhu, X.~Luo, T.~Zhang, Q.~Sun, and J.~Lu, ``Differentially private federated learning with an adaptive noise mechanism,'' \emph{{IEEE} Trans. Inf. Forensics Secur.}, vol.~19, pp. 74--87, 2024.

\bibitem{DBLP:journals/jsac/OkegbileCZCY23}
S.~D. Okegbile, J.~Cai, H.~Zheng, J.~Chen, and C.~Yi, ``Differentially private federated multi-task learning framework for enhancing human-to-virtual connectivity in human digital twin,'' \emph{{IEEE} J. Sel. Areas Commun.}, vol.~41, no.~11, pp. 3533--3547, 2023.

\bibitem{DBLP:journals/tdsc/LinWLSHD23}
X.~Lin, J.~Wu, J.~Li, C.~Sang, S.~Hu, and M.~J. Deen, ``Heterogeneous differential-private federated learning: Trading privacy for utility truthfully,'' \emph{{IEEE} Trans. Dependable Secur. Comput.}, vol.~20, no.~6, pp. 5113--5129, 2023.

\bibitem{sun2021pain}
P.~Sun, H.~Che, Z.~Wang, Y.~Wang, T.~Wang, L.~Wu, and H.~Shao, ``Pain-fl: Personalized privacy-preserving incentive for federated learning,'' \emph{IEEE Journal on Selected Areas in Communications}, vol.~39, no.~12, pp. 3805--3820, 2021.

\bibitem{sun2022profit}
P.~Sun, X.~Chen, G.~Liao, and J.~Huang, ``A profit-maximizing model marketplace with differentially private federated learning,'' in \emph{IEEE INFOCOM 2022-IEEE Conference on Computer Communications}.\hskip 1em plus 0.5em minus 0.4em\relax IEEE, 2022, pp. 1439--1448.

\bibitem{hu2020federated}
R.~Hu, Y.~Gong, and Y.~Guo, ``Federated learning with sparsification-amplified privacy and adaptive optimization,'' \emph{arXiv preprint arXiv:2008.01558}, 2020.

\bibitem{hu2023federated}
R.~Hu, Y.~Guo, and Y.~Gong, ``Federated learning with sparsified model perturbation: Improving accuracy under client-level differential privacy,'' \emph{IEEE Transactions on Mobile Computing}, 2023.

\bibitem{wang2021datalens}
B.~Wang, F.~Wu, Y.~Long, L.~Rimanic, C.~Zhang, and B.~Li, ``Datalens: Scalable privacy preserving training via gradient compression and aggregation,'' in \emph{Proceedings of the 2021 ACM SIGSAC Conference on Computer and Communications Security}, 2021, pp. 2146--2168.

\bibitem{chen2024privacy}
W.-N. Chen, D.~Song, A.~Ozgur, and P.~Kairouz, ``Privacy amplification via compression: Achieving the optimal privacy-accuracy-communication trade-off in distributed mean estimation,'' \emph{Advances in Neural Information Processing Systems}, vol.~36, 2024.

\bibitem{zhang2023trading}
X.~Zhang, Y.~Kang, K.~Chen, L.~Fan, and Q.~Yang, ``Trading off privacy, utility, and efficiency in federated learning,'' \emph{ACM Transactions on Intelligent Systems and Technology}, vol.~14, no.~6, pp. 1--32, 2023.

\bibitem{tekgul2021waffle}
B.~G. Tekgul, Y.~Xia, S.~Marchal, and N.~Asokan, ``Waffle: Watermarking in federated learning,'' in \emph{2021 40th International Symposium on Reliable Distributed Systems (SRDS)}.\hskip 1em plus 0.5em minus 0.4em\relax IEEE, 2021, pp. 310--320.

\bibitem{sun2022black}
T.~Sun, Y.~Shao, H.~Qian, X.~Huang, and X.~Qiu, ``Black-box tuning for language-model-as-a-service,'' in \emph{International Conference on Machine Learning}.\hskip 1em plus 0.5em minus 0.4em\relax PMLR, 2022, pp. 20\,841--20\,855.

\bibitem{lin2023efficient}
Z.~Lin, Y.~Sun, Y.~Shi, X.~Wang, L.~Huang, L.~Shen, and D.~Tao, ``Efficient federated prompt tuning for black-box large pre-trained models,'' \emph{arXiv preprint arXiv:2310.03123}, 2023.

\bibitem{sun2024fedbpt}
J.~Sun, Z.~Xu, H.~Yin, D.~Yang, D.~Xu, Y.~Liu, Z.~Du, Y.~Chen, and H.~R. Roth, ``Fedbpt: Efficient federated black-box prompt tuning for large language models,'' in \emph{Forty-first International Conference on Machine Learning}, 2024.

\bibitem{yu2023leaked}
S.~Yu, J.~Hong, Y.~Zeng, F.~Wang, R.~Jia, and J.~Zhou, ``Who leaked the model? tracking ip infringers in accountable federated learning,'' in \emph{NeurIPS 2023 Workshop on Regulatable ML}, 2023.

\bibitem{kang2019incentive}
J.~Kang, Z.~Xiong, S.~Niyato, and J.~Zhang, ``Incentive mechanism for reliable federated learning: A joint optimization approach to combining reputation and contract theory,'' \emph{IEEE Internet of Things Journal}, vol.~6, no.~6, pp. 10\,700--10\,714, 2019.

\bibitem{kang2019toward}
J.~Kang, Z.~Xiong, D.~Niyato, D.~Ye, D.~I. Kim, and J.~Zhao, ``Toward secure blockchain-enabled internet of vehicles: Optimizing consensus management using reputation and contract theory,'' \emph{IEEE Transactions on Vehicular Technology}, vol.~68, no.~3, pp. 2906--2920, 2019.

\bibitem{hou2017incentive}
Z.~Hou, H.~Chen, Y.~Li, and B.~Vucetic, ``Incentive mechanism design for wireless energy harvesting-based internet of things,'' \emph{IEEE Internet of Things Journal}, vol.~5, no.~4, pp. 2620--2632, 2017.

\bibitem{liu2017design}
T.~Liu, J.~Li, F.~Shu, M.~Tao, W.~Chen, and Z.~Han, ``Design of contract-based trading mechanism for a small-cell caching system,'' \emph{IEEE Transactions on Wireless Communications}, vol.~16, no.~10, pp. 6602--6617, 2017.

\bibitem{ding2020incentive}
N.~Ding, Z.~Fang, and J.~Huang, ``Incentive mechanism design for federated learning with multi-dimensional private information,'' in \emph{2020 18th International Symposium on Modeling and Optimization in Mobile, Ad Hoc, and Wireless Networks (WiOPT)}.\hskip 1em plus 0.5em minus 0.4em\relax IEEE, 2020, pp. 1--8.

\bibitem{sarikaya2019motivating}
Y.~Sarikaya and O.~Ercetin, ``Motivating workers in federated learning: A stackelberg game perspective,'' \emph{IEEE Networking Letters}, vol.~2, no.~1, pp. 23--27, 2019.

\bibitem{feng2019joint}
S.~Feng, D.~Niyato, P.~Wang, D.~I. Kim, and Y.-C. Liang, ``Joint service pricing and cooperative relay communication for federated learning,'' in \emph{2019 International Conference on Internet of Things (iThings) and IEEE Green Computing and Communications (GreenCom) and IEEE Cyber, Physical and Social Computing (CPSCom) and IEEE Smart Data (SmartData)}.\hskip 1em plus 0.5em minus 0.4em\relax IEEE, 2019, pp. 815--820.

\bibitem{lim2021towards}
W.~Y.~B. Lim, J.~Huang, Z.~Xiong, J.~Kang, D.~Niyato, X.-S. Hua, C.~Leung, and C.~Miao, ``Towards federated learning in uav-enabled internet of vehicles: A multi-dimensional contract-matching approach,'' \emph{IEEE Transactions on Intelligent Transportation Systems}, vol.~22, no.~8, pp. 5140--5154, 2021.

\bibitem{pandey2020crowdsourcing}
S.~R. Pandey, N.~H. Tran, and C.~S. Hong, ``A crowdsourcing framework for on-device federated learning,'' \emph{IEEE Transactions on Wireless Communications}, vol.~19, no.~5, pp. 3241--3256, 2020.

\bibitem{dinh2020federated}
C.~T. Dinh, N.~H. Tran, M.~N. Nguyen, C.~S. Hong, W.~Bao, A.~Y. Zomaya, and V.~Gramoli, ``Federated learning over wireless networks: Convergence analysis and resource allocation,'' \emph{IEEE/ACM Transadinh2020federatedctions on Networking}, vol.~29, no.~1, pp. 398--409, 2020.

\bibitem{khan2020federated}
L.~U. Khan, S.~R. Pandey, N.~H. Tran, W.~Saad, Z.~Han, M.~N. Nguyen, and C.~S. Hong, ``Federated learning for edge networks: Resource optimization and incentive mechanism,'' \emph{IEEE Communications Magazine}, vol.~58, no.~10, pp. 88--93, 2020.

\bibitem{hu2020trading}
R.~Hu and Y.~Gong, ``Trading data for learning: Incentive mechanism for on-device federated learning,'' in \emph{GLOBECOM 2020-2020 IEEE Global Communications Conference}.\hskip 1em plus 0.5em minus 0.4em\relax IEEE, 2020, pp. 1--6.

\bibitem{lee2020market}
J.~Lee, D.~Kim, and D.~Niyato, ``Market analysis of distributed learning resource management for internet of things: A game-theoretic approach,'' \emph{IEEE Internet of Things Journal}, vol.~7, no.~9, pp. 8430--8439, 2020.

\bibitem{sarikaya2020regulating}
Y.~Sarikaya and O.~Ercetin, ``Regulating workers in federated learning by yardstick competition,'' in \emph{Proceedings of the 13th EAI International Conference on Performance Evaluation Methodologies and Tools}, 2020, pp. 150--155.

\bibitem{qu2020privacy}
X.~Qu, Q.~Hu, and S.~Wang, ``Privacy-preserving model training architecture for intelligent edge computing,'' \emph{Computer Communications}, vol. 162, pp. 94--101, 2020.

\bibitem{song2019profit}
T.~Song, Y.~Tong, and S.~Wei, ``Profit allocation for federated learning,'' in \emph{IEEE BigData}, 2019, pp. 2577--2586.

\bibitem{yu2020sustainable}
H.~Yu, Z.~Liu, Y.~Liu, T.~Chen, M.~Cong, X.~Weng, D.~Niyato, and Q.~Yang, ``A sustainable incentive scheme for federated learning,'' \emph{IEEE Intelligent Systems}, vol.~35, no.~4, pp. 58--69, 2020.

\bibitem{li2019credit}
Z.~Li, Z.~Yang, S.~Xie, W.~Chen, and K.~Liu, ``Credit-based payments for fast computing resource trading in edge-assisted internet of things,'' \emph{IEEE Internet of Things Journal}, vol.~6, no.~4, pp. 6606--6617, 2019.

\bibitem{krishnaraj2022future}
N.~Krishnaraj, K.~Bellam, B.~Sivakumar, and A.~Daniel, ``The future of cloud computing: Blockchain-based decentralized cloud/fog solutions--challenges, opportunities, and standards,'' \emph{Blockchain Security in Cloud Computing}, pp. 207--226, 2022.

\bibitem{zavodovski2019decloud}
A.~Zavodovski, S.~Bayhan, N.~Mohan, P.~Zhou, W.~Wong, and J.~Kangasharju, ``Decloud: Truthful decentralized double auction for edge clouds,'' in \emph{Proceedings of the 2019 IEEE 39th International Conference on Distributed Computing Systems (ICDCS'19)}, 2019, pp. 2157--2167.

\bibitem{hong2020optimizing}
H.-J. Hong, W.~Fan, C.~E. Chow, X.~Zhou, and S.-Y. Chang, ``Optimizing social welfare for task offloading in mobile edge computing,'' in \emph{Proceedings of the 2020 IFIP Networking Conference (Networking'20)}, 2020, pp. 524--528.

\bibitem{bahreini2018envy}
T.~Bahreini, H.~Badri, and D.~Grosu, ``An envy-free auction mechanism for resource allocation in edge computing systems,'' in \emph{Proceedings of the 2018 IEEE/ACM Symposium on Edge Computing (SEC'18)}, 2018, pp. 313--322.

\bibitem{gao2019auction}
G.~Gao, M.~Xiao, J.~Wu, H.~Huang, S.~Wang, and G.~Chen, ``Auction-based vm allocation for deadline-sensitive tasks in distributed edge cloud,'' \emph{IEEE Transactions on Services Computing}, vol.~14, no.~6, pp. 1702--1716, 2019.

\bibitem{jiao2018social}
Y.~Jiao, P.~Wang, D.~Niyato, and Z.~Xiong, ``Social welfare maximization auction in edge computing resource allocation for mobile blockchain,'' in \emph{Proceedings of the 2018 IEEE International Conference on Communications (ICC'18)}, 2018, pp. 1--6.

\bibitem{jiao2019auction}
Y.~Jiao, P.~Wang, D.~Niyato, and K.~Suankaewmanee, ``Auction mechanisms in cloud/fog computing resource allocation for public blockchain networks,'' \emph{IEEE Transactions on Parallel and Distributed Systems}, vol.~30, no.~9, pp. 1975--1989, 2019.

\bibitem{yang2020task}
S.~Yang, ``A task offloading solution for internet of vehicles using combination auction matching model based on mobile edge computing,'' \emph{IEEE Access}, vol.~8, pp. 53\,261--53\,273, 2020.

\bibitem{jiao2020toward}
Y.~Jiao, P.~Wang, D.~Niyato, B.~Lin, and D.~I. Kim, ``Toward an automated auction framework for wireless federated learning services market,'' \emph{IEEE Transactions on Mobile Computing}, vol.~20, no.~10, pp. 3034--3048, 2020.

\bibitem{zeng2020fmore}
R.~Zeng, S.~Zhang, J.~Wang, and X.~Chu, ``Fmore: An incentive scheme of multi-dimensional auction for federated learning in {MEC},'' in \emph{ICDCS}, 2020, pp. 278--288.

\bibitem{ying2020double}
C.~Ying, H.~Jin, X.~Wang, and Y.~Luo, ``Double insurance: Incentivized federated learning with differential privacy in mobile crowdsensing,'' in \emph{2020 International Symposium on Reliable Distributed Systems (SRDS)}.\hskip 1em plus 0.5em minus 0.4em\relax IEEE, 2020, pp. 81--90.

\bibitem{le2020auction}
T.~H.~T. Le, N.~H. Tran, Y.~K. Tun, Z.~Han, and C.~S. Hong, ``Auction based incentive design for efficient federated learning in cellular wireless networks,'' in \emph{WCNC}, 2020, pp. 1--6.

\bibitem{le2021incentive}
T.~H. Thi~Le, N.~H. Tran, Y.~K. Tun, M.~N.~H. Nguyen, S.~R. Pandey, Z.~Han, and C.~S. Hong, ``An incentive mechanism for federated learning in wireless cellular networks: An auction approach,'' \emph{IEEE Transactions on Wireless Communications}, vol.~20, no.~8, pp. 4874--4887, 2021.

\bibitem{zhang2021incentive}
J.~Zhang, Y.~Wu, and R.~Pan, ``Incentive mechanism for horizontal federated learning based on reputation and reverse auction,'' in \emph{WWW}, 2021, p. 947–956.

\bibitem{roy2021distributed}
P.~Roy, S.~Sarker, M.~A. Razzaque, M.~Mamun-or Rashid, M.~M. Hassan, and G.~Fortino, ``Distributed task allocation in mobile device cloud exploiting federated learning and subjective logic,'' \emph{Journal of Systems Architecture}, vol. 113, p. 101972, 2021.

\bibitem{deng2021fair}
Y.~Deng, F.~Lyu, J.~Ren, Y.-C. Chen, P.~Yang, Y.~Zhou, and Y.~Zhang, ``Fair: Quality-aware federated learning with precise user incentive and model aggregation,'' in \emph{INFOCOM}, 2021.

\bibitem{zhang2022auction}
J.~Zhang, Y.~Wu, and R.~Pan, ``Auction-based ex-post-payment incentive mechanism design for horizontal federated learning with reputation and contribution measurement,'' \emph{arXiv preprint arXiv:2201.02410}, 2022.

\bibitem{zhang2022online}
------, ``Online auction-based incentive mechanism design for horizontal federated learning with budget constraint,'' \emph{arXiv preprint}, p. 2201.09047, 2022.

\bibitem{tang2023utility}
X.~Tang and H.~Yu, ``Utility-maximizing bidding strategy for data consumers in auction-based federated learning,'' in \emph{Proceedings of the 2023 IEEE International Conference on Multimedia and Expo (ICME'23)}, 2023.

\bibitem{tang2023competitive}
X.~Tang and H.~Yu", ``Competitive-cooperative multi-agent reinforcement learning for auction-based federated learning,'' in \emph{Proceedings of the 32nd International Joint Conference on Artificial Intelligence (IJCAI'23)}, 2023.

\bibitem{tang2023multi}
X.~Tang and H.~Yu, ``Multi-session budget optimization for forward auction-based federated learning,'' \emph{arXiv preprint arXiv:2311.12548}, 2023.

\bibitem{liu2020fedcoin}
Y.~Liu, Z.~Ai, S.~Sun, S.~Zhang, Z.~Liu, and H.~Yu, ``Fedcoin: A peer-to-peer payment system for federated learning,'' in \emph{Federated learning: privacy and incentive}.\hskip 1em plus 0.5em minus 0.4em\relax Springer, 2020, pp. 125--138.

\bibitem{ghosh2020efficient}
A.~Ghosh, J.~Chung, D.~Yin, and K.~Ramchandran, ``An efficient framework for clustered federated learning,'' \emph{Advances in Neural Information Processing Systems}, vol.~33, pp. 19\,586--19\,597, 2020.

\bibitem{liu2022federated}
R.~Liu, P.~Xing, Z.~Deng, A.~Li, C.~Guan, and H.~Yu, ``Federated graph neural networks: Overview, techniques and challenges,'' \emph{arXiv preprint arXiv:2202.07256}, 2022.

\bibitem{FedGH:2023}
L.~Yi, G.~Wang, X.~Liu, Z.~Shi, and H.~Yu, ``{FedGH}: Heterogeneous federated learning with generalized global header,'' in \emph{Proceedings of the 31st ACM International Conference on Multimedia (ACM MM'23)}, 2023, pp. 8686--8696.

\bibitem{wang2023fedlego}
J.~Wang, S.~Cui, and F.~Ma, ``Fedlego: Enabling heterogenous model cooperation via brick reassembly in federated learning,'' in \emph{International Workshop on Federated Learning for Distributed Data Mining}, 2023.

\bibitem{wang2023towards}
J.~Wang, X.~Yang, S.~Cui, L.~Che, L.~Lyu, D.~Xu, and F.~Ma, ``Towards personalized federated learning via heterogeneous model reassembly,'' \emph{arXiv preprint arXiv:2308.08643}, 2023.

\bibitem{li2023towards}
A.~Li, R.~Liu, M.~Hu, L.~A. Tuan, and H.~Yu, ``Towards interpretable federated learning,'' \emph{arXiv preprint arXiv:2302.13473}, 2023.

\bibitem{zhu2023promptbench}
K.~Zhu, J.~Wang, J.~Zhou, Z.~Wang, H.~Chen, Y.~Wang, L.~Yang, W.~Ye, N.~Z. Gong, Y.~Zhang \emph{et~al.}, ``Promptbench: Towards evaluating the robustness of large language models on adversarial prompts,'' \emph{arXiv preprint arXiv:2306.04528}, 2023.

\bibitem{kuchnik2023validating}
M.~Kuchnik, V.~Smith, and G.~Amvrosiadis, ``Validating large language models with relm,'' \emph{Proceedings of Machine Learning and Systems}, vol.~5, 2023.

\bibitem{nagalapatti2021game}
L.~Nagalapatti and R.~Narayanam, ``Game of gradients: Mitigating irrelevant clients in federated learning,'' in \emph{Proceedings of the AAAI Conference on Artificial Intelligence}, vol.~35, no.~10, 2021, pp. 9046--9054.

\bibitem{fan2022fair}
Z.~Fan, H.~Fang, Z.~Zhou, J.~Pei, M.~P. Friedlander, and Y.~Zhang, ``Fair and efficient contribution valuation for vertical federated learning,'' \emph{arXiv preprint arXiv:2201.02658}, 2022.

\bibitem{wei2020efficient}
S.~Wei, Y.~Tong, Z.~Zhou, and T.~Song, ``Efficient and fair data valuation for horizontal federated learning,'' in \emph{Federated Learning}.\hskip 1em plus 0.5em minus 0.4em\relax Springer, 2020, pp. 139--152.

\bibitem{wang2020principled}
T.~Wang, J.~Rausch, C.~Zhang, R.~Jia, and D.~Song, ``A principled approach to data valuation for federated learning,'' in \emph{Federated Learning}.\hskip 1em plus 0.5em minus 0.4em\relax Springer, 2020, pp. 153--167.

\bibitem{wang2022efficient}
J.~Wang, L.~Zhang, A.~Li, X.~You, and H.~Cheng, ``Efficient participant contribution evaluation for horizontal and vertical federated learning,'' in \emph{2022 IEEE 38th International Conference on Data Engineering (ICDE)}.\hskip 1em plus 0.5em minus 0.4em\relax IEEE, 2022, pp. 911--923.

\bibitem{liu2022gtg}
Z.~Liu, Y.~Chen, H.~Yu, Y.~Liu, and L.~Cui, ``Gtg-shapley: Efficient and accurate participant contribution evaluation in federated learning,'' \emph{ACM Transactions on Intelligent Systems and Technology (TIST)}, vol.~13, no.~4, pp. 1--21, 2022.

\bibitem{liu2022contribution}
Z.~Liu, Y.~Chen, Y.~Zhao, H.~Yu, Y.~Liu, R.~Bao, J.~Jiang, Z.~Nie, Q.~Xu, and Q.~Yang, ``Contribution-aware federated learning for smart healthcare,'' in \emph{Proceedings of the AAAI Conference on Artificial Intelligence}, vol.~36, no.~11, 2022, pp. 12\,396--12\,404.

\bibitem{zheng2022secure}
S.~Zheng, Y.~Cao, and M.~Yoshikawa, ``Secure shapley value for cross-silo federated learning,'' \emph{arXiv preprint arXiv:2209.04856}, 2022.

\bibitem{ma2021transparent}
S.~Ma, Y.~Cao, and L.~Xiong, ``Transparent contribution evaluation for secure federated learning on blockchain,'' in \emph{2021 IEEE 37th International Conference on Data Engineering Workshops (ICDEW)}.\hskip 1em plus 0.5em minus 0.4em\relax IEEE, 2021, pp. 88--91.

\bibitem{wang2019measure}
G.~Wang, C.~X. Dang, and Z.~Zhou, ``Measure contribution of participants in federated learning,'' in \emph{IEEE Big Data}, 2019, pp. 2597--2604.

\bibitem{zhang2022intrinsic}
L.~Zhang, L.~Fan, Y.~Luo, and L.-Y. Duan, ``Intrinsic performance influence-based participant contribution estimation for horizontal federated learning,'' \emph{ACM Transactions on Intelligent Systems and Technology (TIST)}, vol.~13, no.~6, pp. 1--24, 2022.

\bibitem{li2021efficient}
A.~Li, L.~Zhang, J.~Wang, J.~Tan, F.~Han, Y.~Qin, N.~M. Freris, and X.-Y. Li, ``Efficient federated-learning model debugging,'' in \emph{2021 IEEE 37th International Conference on Data Engineering (ICDE)}.\hskip 1em plus 0.5em minus 0.4em\relax IEEE, 2021, pp. 372--383.

\bibitem{koh2017understanding}
P.~W. Koh and P.~Liang, ``Understanding black-box predictions via influence functions,'' in \emph{International conference on machine learning}.\hskip 1em plus 0.5em minus 0.4em\relax PMLR, 2017, pp. 1885--1894.

\bibitem{xue2021toward}
Y.~Xue, C.~Niu, Z.~Zheng, S.~Tang, C.~Lyu, F.~Wu, and G.~Chen, ``Toward understanding the influence of individual clients in federated learning,'' in \emph{Proceedings of the AAAI Conference on Artificial Intelligence}, vol.~35, no.~12, 2021, pp. 10\,560--10\,567.

\bibitem{koh2019accuracy}
P.~W.~W. Koh, K.-S. Ang, H.~Teo, and P.~S. Liang, ``On the accuracy of influence functions for measuring group effects,'' \emph{Advances in neural information processing systems}, vol.~32, 2019.

\bibitem{li2021privacy}
A.~Li, L.~Zhang, J.~Wang, F.~Han, and X.-Y. Li, ``Privacy-preserving efficient federated-learning model debugging,'' \emph{IEEE Transactions on Parallel and Distributed Systems}, vol.~33, no.~10, pp. 2291--2303, 2021.

\bibitem{kwon2023datainf}
Y.~Kwon, E.~Wu, K.~Wu, and J.~Zou, ``Datainf: Efficiently estimating data influence in lora-tuned llms and diffusion models,'' \emph{arXiv preprint arXiv:2310.00902}, 2023.

\bibitem{li2021sample}
A.~Li, L.~Zhang, J.~Tan, Y.~Qin, J.~Wang, and X.-Y. Li, ``Sample-level data selection for federated learning,'' in \emph{IEEE INFOCOM 2021-IEEE Conference on Computer Communications}.\hskip 1em plus 0.5em minus 0.4em\relax IEEE, 2021, pp. 1--10.

\bibitem{cassara2022federated}
P.~Cassara, A.~Gotta, and L.~Valerio, ``Federated feature selection for cyber-physical systems of systems,'' \emph{IEEE Transactions on Vehicular Technology}, vol.~71, no.~9, pp. 9937--9950, 2022.

\bibitem{li2021privacy1}
X.~Li, R.~Dowsley, and M.~De~Cock, ``Privacy-preserving feature selection with secure multiparty computation,'' in \emph{International Conference on Machine Learning}.\hskip 1em plus 0.5em minus 0.4em\relax PMLR, 2021, pp. 6326--6336.

\bibitem{pansecure}
F.~Pan, D.~Meng, Y.~Zhang, and X.~Li, ``Secure federated feature selection for cross-feature federated learning,'' \emph{arXiv preprint}, 2020.

\bibitem{wang2019interpret}
G.~Wang, ``Interpret federated learning with shapley values,'' \emph{arXiv preprint arXiv:1905.04519}, 2019.

\bibitem{chen2020federated}
Y.~Chen, Y.~Ning, Z.~Chai, and H.~Rangwala, ``Federated multi-task learning with hierarchical attention for sensor data analytics,'' in \emph{2020 International Joint Conference on Neural Networks (IJCNN)}.\hskip 1em plus 0.5em minus 0.4em\relax IEEE, 2020, pp. 1--8.

\bibitem{younisflames2graph}
R.~Younis, Z.~Ahmadi, A.~Hakmeh, and M.~Fisichella, ``Flames2graph: An interpretable federated multivariate time series classification framework,'' 2023.

\bibitem{li2023fedsdg}
A.~Li, H.~Peng, L.~Zhang, J.~Huang, Q.~Guo, H.~Yu, and Y.~Liu, ``Fedsdg-fs: Efficient and secure feature selection for vertical federated learning,'' \emph{arXiv preprint arXiv:2302.10417}, 2023.

\bibitem{li2023efficient}
A.~Li, J.~Huang, J.~Jia, H.~Peng, L.~Zhang, L.~A. Tuan, H.~Yu, and X.-Y. Li, ``Efficient and privacy-preserving feature importance-based vertical federated learning,'' \emph{IEEE Transactions on Mobile Computing}, no.~01, pp. 1--17, 2023.

\bibitem{lin2004rouge}
C.-Y. Lin, ``Rouge: A package for automatic evaluation of summaries,'' in \emph{Text summarization branches out}, 2004, pp. 74--81.

\bibitem{zhang2019bertscore}
T.~Zhang, V.~Kishore, F.~Wu, K.~Q. Weinberger, and Y.~Artzi, ``Bertscore: Evaluating text generation with bert,'' \emph{arXiv preprint arXiv:1904.09675}, 2019.

\bibitem{papineni2002bleu}
K.~Papineni, S.~Roukos, T.~Ward, and W.-J. Zhu, ``Bleu: a method for automatic evaluation of machine translation,'' in \emph{Proceedings of the 40th annual meeting of the Association for Computational Linguistics}, 2002, pp. 311--318.

\bibitem{lin2023llm}
Y.-T. Lin and Y.-N. Chen, ``Llm-eval: Unified multi-dimensional automatic evaluation for open-domain conversations with large language models,'' \emph{arXiv preprint arXiv:2305.13711}, 2023.

\bibitem{wang2023pandalm}
Y.~Wang, Z.~Yu, Z.~Zeng, L.~Yang, C.~Wang, H.~Chen, C.~Jiang, R.~Xie, J.~Wang, X.~Xie \emph{et~al.}, ``Pandalm: An automatic evaluation benchmark for llm instruction tuning optimization,'' \emph{arXiv preprint arXiv:2306.05087}, 2023.

\bibitem{bubeck2023sparks}
S.~Bubeck, V.~Chandrasekaran, R.~Eldan, J.~Gehrke, E.~Horvitz, E.~Kamar, P.~Lee, Y.~T. Lee, Y.~Li, S.~Lundberg \emph{et~al.}, ``Sparks of artificial general intelligence: Early experiments with gpt-4,'' \emph{arXiv preprint arXiv:2303.12712}, 2023.

\bibitem{jain2023bring}
N.~Jain, K.~Saifullah, Y.~Wen, J.~Kirchenbauer, M.~Shu, A.~Saha, M.~Goldblum, J.~Geiping, and T.~Goldstein, ``Bring your own data! self-supervised evaluation for large language models,'' \emph{arXiv preprint arXiv:2306.13651}, 2023.

\bibitem{schoch2023data}
S.~Schoch, R.~Mishra, and Y.~Ji, ``Data selection for fine-tuning large language models using transferred shapley values,'' \emph{arXiv preprint arXiv:2306.10165}, 2023.

\bibitem{xing2023fedlogic}
P.~Xing, S.~Lu, and H.~Yu, ``Fedlogic: Interpretable federated multi-domain chain-of-thought prompt selection for large language models,'' \emph{arXiv preprint arXiv:2308.15324}, 2023.

\bibitem{zhao2023privacy}
J.~Zhao, ``Privacy-preserving fine-tuning of artificial intelligence (ai) foundation models with federated learning, differential privacy, offsite tuning, and parameter-efficient fine-tuning (peft),'' \emph{Authorea Preprints}, 2023.

\bibitem{zhang2023towards}
J.~Zhang, S.~Vahidian, M.~Kuo, C.~Li, R.~Zhang, G.~Wang, and Y.~Chen, ``Towards building the federated gpt: Federated instruction tuning,'' \emph{arXiv preprint arXiv:2305.05644}, 2023.

\bibitem{lin2021fednlp}
B.~Y. Lin, C.~He, Z.~Zeng, H.~Wang, Y.~Huang, C.~Dupuy, R.~Gupta, M.~Soltanolkotabi, X.~Ren, and S.~Avestimehr, ``Fednlp: Benchmarking federated learning methods for natural language processing tasks,'' \emph{arXiv preprint arXiv:2104.08815}, 2021.

\bibitem{press2022measuring}
O.~Press, M.~Zhang, S.~Min, L.~Schmidt, N.~A. Smith, and M.~Lewis, ``Measuring and narrowing the compositionality gap in language models,'' \emph{arXiv preprint arXiv:2210.03350}, 2022.

\bibitem{kang2023llms}
W.-C. Kang, J.~Ni, N.~Mehta, M.~Sathiamoorthy, L.~Hong, E.~Chi, and D.~Z. Cheng, ``Do llms understand user preferences? evaluating llms on user rating prediction,'' \emph{arXiv preprint arXiv:2305.06474}, 2023.

\bibitem{dai2023uncovering}
S.~Dai, N.~Shao, H.~Zhao, W.~Yu, Z.~Si, C.~Xu, Z.~Sun, X.~Zhang, and J.~Xu, ``Uncovering chatgpt's capabilities in recommender systems,'' \emph{arXiv preprint arXiv:2305.02182}, 2023.

\bibitem{he2020fedml}
C.~He, S.~Li, J.~So, X.~Zeng, M.~Zhang, H.~Wang, X.~Wang, P.~Vepakomma, A.~Singh, H.~Qiu \emph{et~al.}, ``Fedml: A research library and benchmark for federated machine learning,'' \emph{arXiv preprint arXiv:2007.13518}, 2020.

\bibitem{kuang2023federatedscope}
W.~Kuang, B.~Qian, Z.~Li, D.~Chen, D.~Gao, X.~Pan, Y.~Xie, Y.~Li, B.~Ding, and J.~Zhou, ``Federatedscope-llm: A comprehensive package for fine-tuning large language models in federated learning,'' \emph{arXiv preprint arXiv:2309.00363}, 2023.

\bibitem{fan2023fate}
T.~Fan, Y.~Kang, G.~Ma, W.~Chen, W.~Wei, L.~Fan, and Q.~Yang, ``Fate-llm: A industrial grade federated learning framework for large language models,'' \emph{arXiv preprint arXiv:2310.10049}, 2023.

\bibitem{feng2023learning}
C.-M. Feng, B.~Li, X.~Xu, Y.~Liu, H.~Fu, and W.~Zuo, ``Learning federated visual prompt in null space for mri reconstruction,'' in \emph{Proceedings of the IEEE/CVF Conference on Computer Vision and Pattern Recognition}, 2023, pp. 8064--8073.

\bibitem{shin2023fedtherapist}
J.~Shin, H.~Yoon, S.~Lee, S.~Park, Y.~Liu, J.~D. Choi, and S.-J. Lee, ``Fedtherapist: Mental health monitoring with user-generated linguistic expressions on smartphones via federated learning,'' in \emph{Proceedings of the 2023 Conference on Empirical Methods in Natural Language Processing}, 2023, pp. 11\,971--11\,988.

\bibitem{zeng2024federated}
H.~Zeng, Z.~Yue, Q.~Jiang, and D.~Wang, ``Federated recommendation via hybrid retrieval augmented generation,'' \emph{arXiv preprint arXiv:2403.04256}, 2024.

\bibitem{zhang2024transfr}
H.~Zhang, H.~Liu, H.~Li, and Y.~Li, ``Transfr: Transferable federated recommendation with pre-trained language models,'' \emph{arXiv preprint arXiv:2402.01124}, 2024.

\bibitem{zhao2024llm}
J.~Zhao, W.~Wang, C.~Xu, Z.~Ren, S.-K. Ng, and T.-S. Chua, ``Llm-based federated recommendation,'' \emph{arXiv preprint arXiv:2402.09959}, 2024.

\bibitem{azam2023federated}
S.~S. Azam, M.~Pelikan, V.~Feldman, K.~Talwar, J.~Silovsky, and T.~Likhomanenko, ``Federated learning for speech recognition: Revisiting current trends towards large-scale asr,'' in \emph{International Workshop on Federated Learning in the Age of Foundation Models in Conjunction with NeurIPS 2023}, 2023.

\bibitem{jia2023joint}
J.~Jia, K.~Li, M.~Malek, K.~Malik, J.~Mahadeokar, O.~Kalinli, and F.~Seide, ``Joint federated learning and personalization for on-device asr,'' in \emph{2023 IEEE Automatic Speech Recognition and Understanding Workshop (ASRU)}.\hskip 1em plus 0.5em minus 0.4em\relax IEEE, 2023, pp. 1--8.

\bibitem{du2024communication}
Y.~Du, Z.~Zhang, L.~Yue, X.~Huang, Y.~Zhang, T.~Xu, L.~Xu, and E.~Chen, ``Communication-efficient personalized federated learning for speech-to-text tasks,'' in \emph{ICASSP 2024-2024 IEEE International Conference on Acoustics, Speech and Signal Processing (ICASSP)}.\hskip 1em plus 0.5em minus 0.4em\relax IEEE, 2024, pp. 10\,001--10\,005.

\bibitem{zhao2024breaking}
W.~Zhao, Y.~Chen, R.~Lee, X.~Qiu, Y.~Gao, H.~Fan, and N.~D. Lane, ``Breaking physical and linguistic borders: Multilingual federated prompt tuning for low-resource languages,'' in \emph{The Twelfth International Conference on Learning Representations}, 2024.

\bibitem{chu2024only}
Y.-W. Chu, D.-J. Han, and C.~G. Brinton, ``Only send what you need: Learning to communicate efficiently in federated multilingual machine translation,'' in \emph{Companion Proceedings of the ACM on Web Conference 2024}, 2024, pp. 1548--1557.

\bibitem{liu2023communication}
Y.~Liu, X.~Bi, L.~Li, S.~Chen, W.~Yang, and X.~Sun, ``Communication efficient federated learning for multilingual neural machine translation with adapter,'' in \emph{Findings of the Association for Computational Linguistics: ACL 2023}, 2023, pp. 5315--5328.

\bibitem{weller2022pretrained}
O.~Weller, M.~Marone, V.~Braverman, D.~Lawrie, and B.~Van~Durme, ``Pretrained models for multilingual federated learning,'' in \emph{Proceedings of the 2022 Conference of the North American Chapter of the Association for Computational Linguistics: Human Language Technologies}, 2022, pp. 1413--1421.

\bibitem{wang2022fedkc}
H.~Wang, H.~Zhao, Y.~Wang, T.~Yu, J.~Gu, and J.~Gao, ``Fedkc: Federated knowledge composition for multilingual natural language understanding,'' in \emph{Proceedings of the ACM Web Conference 2022}, 2022, pp. 1839--1850.

\bibitem{gao2023examining}
K.~Gao, S.~He, Z.~He, J.~Lin, Q.~Pei, J.~Shao, and W.~Zhang, ``Examining user-friendly and open-sourced large gpt models: A survey on language, multimodal, and scientific gpt models,'' \emph{arXiv preprint arXiv:2308.14149}, 2023.

\bibitem{gao2023multi}
Y.~Gao, Y.~Zhao, and H.~Yu, ``Multi-tier client selection for mobile federated learning networks,'' in \emph{ICME}, 2023.

\bibitem{thapa2022splitfed}
C.~Thapa, P.~C.~M. Arachchige, S.~Camtepe, and L.~Sun, ``Splitfed: When federated learning meets split learning,'' in \emph{Proceedings of the AAAI Conference on Artificial Intelligence}, vol.~36, no.~8, 2022, pp. 8485--8493.

\bibitem{shah2021model}
S.~M. Shah and V.~K. Lau, ``Model compression for communication efficient federated learning,'' \emph{IEEE Transactions on Neural Networks and Learning Systems}, 2021.

\bibitem{he2020group}
C.~He, M.~Annavaram, and S.~Avestimehr, ``Group knowledge transfer: Federated learning of large {CNN}s at the edge,'' \emph{Advances in Neural Information Processing Systems}, vol.~33, pp. 14\,068--14\,080, 2020.

\bibitem{konevcny2016federated}
J.~Kone{\v{c}}n{\`y}, H.~B. McMahan, D.~Ramage, and P.~Richt{\'a}rik, ``Federated optimization: Distributed machine learning for on-device intelligence,'' \emph{arXiv preprint arXiv:1610.02527}, 2016.

\bibitem{ren2023secfedsa}
C.~Ren, H.~Yu, R.~Yan, Q.~Li, Y.~Xu, D.~Niyato, and Z.~Y. Dong, ``Secfedsa: A secure differential privacy-based federated learning approach for smart cyber-physical grid stability assessment,'' \emph{IEEE Internet of Things Journal}, 2023.

\bibitem{fariborz2022llm}
M.~Fariborz, M.~Samani, P.~Fotouhi, R.~Proietti, I.-M. Yi, V.~Akella, J.~Lowe-Power, S.~Palermo, and S.~B. Yoo, ``{LLM}: Realizing low-latency memory by exploiting embedded silicon photonics for irregular workloads,'' in \emph{International Conference on High Performance Computing}.\hskip 1em plus 0.5em minus 0.4em\relax Springer, 2022, pp. 44--64.

\bibitem{jaiswal2023compressing}
A.~Jaiswal, Z.~Gan, X.~Du, B.~Zhang, Z.~Wang, and Y.~Yang, ``Compressing {LLMs}: The truth is rarely pure and never simple,'' \emph{arXiv preprint arXiv:2310.01382}, 2023.

\bibitem{mestoukirdi2023sparser}
M.~Mestoukirdi, O.~Esrafilian, D.~Gesbert, Q.~Li, and N.~Gresset, ``Sparser random networks exist: Enforcing communication-efficient federated learning via regularization,'' \emph{arXiv preprint arXiv:2309.10834}, 2023.

\bibitem{xia2021survey}
Q.~Xia, W.~Ye, Z.~Tao, J.~Wu, and Q.~Li, ``A survey of federated learning for edge computing: Research problems and solutions,'' \emph{High-Confidence Computing}, vol.~1, no.~1, p. 100008, 2021.

\bibitem{wang2019edge}
X.~Wang, Y.~Han, C.~Wang, Q.~Zhao, X.~Chen, and M.~Chen, ``In-edge ai: Intelligentizing mobile edge computing, caching and communication by federated learning,'' \emph{Ieee Network}, vol.~33, no.~5, pp. 156--165, 2019.

\bibitem{yue2022communication}
K.~Yue, R.~Jin, C.-W. Wong, and H.~Dai, ``Communication-efficient federated learning via predictive coding,'' \emph{IEEE Journal of Selected Topics in Signal Processing}, vol.~16, no.~3, pp. 369--380, 2022.

\bibitem{beutel2020flower}
D.~J. Beutel, T.~Topal, A.~Mathur, X.~Qiu, J.~Fernandez-Marques, Y.~Gao, L.~Sani, K.~H. Li, T.~Parcollet, P.~P.~B. de~Gusm{\~a}o \emph{et~al.}, ``Flower: A friendly federated learning research framework,'' \emph{arXiv preprint arXiv:2007.14390}, 2020.

\bibitem{xu2023federated}
M.~Xu, Y.~Wu, D.~Cai, X.~Li, and S.~Wang, ``Federated fine-tuning of billion-sized language models across mobile devices,'' \emph{arXiv preprint arXiv:2308.13894}, 2023.

\bibitem{shen2024large}
Y.~Shen, J.~Shao, X.~Zhang, Z.~Lin, H.~Pan, D.~Li, J.~Zhang, and K.~B. Letaief, ``Large language models empowered autonomous edge ai for connected intelligence,'' \emph{IEEE Communications Magazine}, 2024.

\bibitem{wu2023fedcomp}
D.~Wu, W.~Yang, H.~Jin, X.~Zou, W.~Xia, and B.~Fang, ``Fedcomp: A federated learning compression framework for resource-constrained edge computing devices,'' \emph{IEEE Transactions on Computer-Aided Design of Integrated Circuits and Systems}, 2023.

\bibitem{mao2023safari}
Y.~Mao, Z.~Zhao, M.~Yang, L.~Liang, Y.~Liu, W.~Ding, T.~Lan, and X.-P. Zhang, ``Safari: Sparsity-enabled federated learning with limited and unreliable communications,'' \emph{IEEE Transactions on Mobile Computing}, 2023.

\bibitem{xu2023asynchronous}
C.~Xu, Y.~Qu, Y.~Xiang, and L.~Gao, ``Asynchronous federated learning on heterogeneous devices: A survey,'' \emph{Computer Science Review}, vol.~50, p. 100595, 2023.

\bibitem{liu2021federated}
M.~Liu, S.~Ho, M.~Wang, L.~Gao, Y.~Jin, and H.~Zhang, ``Federated learning meets natural language processing: A survey,'' \emph{arXiv preprint arXiv:2107.12603}, 2021.

\bibitem{chawla2024beyond}
M.~Chawla, G.~R. Gupta, S.~Gaddam, and M.~Wadhwa, ``Beyond federated learning for iot: Efficient split learning with caching \& model customization,'' \emph{IEEE INTERNET OF THINGS JOURNAL}, p.~1, 2024.

\bibitem{zhang2024fedsl}
W.~Zhang, T.~Zhou, Q.~Lu, Y.~Yuan, A.~Tolba, and W.~Said, ``Fedsl: A communication efficient federated learning with split layer aggregation,'' \emph{IEEE Internet of Things Journal}, 2024.

\bibitem{liao2023accelerating}
Y.~Liao, Y.~Xu, H.~Xu, Z.~Yao, L.~Wang, and C.~Qiao, ``Accelerating federated learning with data and model parallelism in edge computing,'' \emph{IEEE/ACM Transactions on Networking}, 2023.

\bibitem{aledhari2020federated}
M.~Aledhari, R.~Razzak, R.~M. Parizi, and F.~Saeed, ``Federated learning: A survey on enabling technologies, protocols, and applications,'' \emph{IEEE Access}, vol.~8, pp. 140\,699--140\,725, 2020.

\bibitem{tran2019federated}
N.~H. Tran, W.~Bao, A.~Zomaya, M.~N. Nguyen, and C.~S. Hong, ``Federated learning over wireless networks: Optimization model design and analysis,'' in \emph{IEEE INFOCOM 2019-IEEE conference on computer communications}.\hskip 1em plus 0.5em minus 0.4em\relax IEEE, 2019, pp. 1387--1395.

\bibitem{charles2024towards}
Z.~Charles, N.~Mitchell, K.~Pillutla, M.~Reneer, and Z.~Garrett, ``Towards federated foundation models: Scalable dataset pipelines for group-structured learning,'' \emph{Advances in Neural Information Processing Systems}, vol.~36, 2024.

\bibitem{jin2023fedml}
W.~Jin, Y.~Yao, S.~Han, C.~Joe-Wong, S.~Ravi, S.~Avestimehr, and C.~He, ``Fedml-he: An efficient homomorphic-encryption-based privacy-preserving federated learning system,'' \emph{arXiv preprint arXiv:2303.10837}, 2023.

\bibitem{rieyan2024advanced}
S.~A. Rieyan, M.~R.~K. News, A.~M. Rahman, S.~A. Khan, S.~T.~J. Zaarif, M.~G.~R. Alam, M.~M. Hassan, M.~Ianni, and G.~Fortino, ``An advanced data fabric architecture leveraging homomorphic encryption and federated learning,'' \emph{Information Fusion}, vol. 102, p. 102004, 2024.

\bibitem{aziz2023exploring}
R.~Aziz, S.~Banerjee, S.~Bouzefrane, and T.~Le~Vinh, ``Exploring homomorphic encryption and differential privacy techniques towards secure federated learning paradigm,'' \emph{Future internet}, vol.~15, no.~9, p. 310, 2023.

\bibitem{ren2021interpretable}
C.~Ren, Y.~Xu, and R.~Zhang, ``An interpretable deep learning method for power system transient stability assessment via tree regularization,'' \emph{IEEE Transactions on Power Systems}, vol.~37, no.~5, pp. 3359--3369, 2021.

\bibitem{shi2023towards}
Y.~Shi, H.~Yu, and C.~Leung, ``Towards fairness-aware federated learning,'' \emph{IEEE Transactions on Neural Networks and Learning Systems}, 2023.

\bibitem{ren2022robustness}
C.~Ren and Y.~Xu, ``Robustness verification for machine-learning-based power system dynamic security assessment models under adversarial examples,'' \emph{IEEE Transactions on Control of Network Systems}, vol.~9, no.~4, pp. 1645--1654, 2022.

\bibitem{ren2021vulnerability}
C.~Ren, X.~Du, Y.~Xu, Q.~Song, Y.~Liu, and R.~Tan, ``Vulnerability analysis, robustness verification, and mitigation strategy for machine learning-based power system stability assessment model under adversarial examples,'' \emph{IEEE Transactions on Smart Grid}, vol.~13, no.~2, pp. 1622--1632, 2021.

\bibitem{ren2022universal}
C.~Ren and Y.~Xu, ``A universal defense strategy for data-driven power system stability assessment models under adversarial examples,'' \emph{IEEE Internet of Things Journal}, 2022.

\bibitem{zhang2024vulnerability}
Z.~Zhang, M.~Liu, M.~Sun, R.~Deng, P.~Cheng, D.~Niyato, M.-Y. Chow, and J.~Chen, ``Vulnerability of machine learning approaches applied in iot-based smart grid: A review,'' \emph{IEEE Internet of Things Journal}, 2024.

\bibitem{liu2020secure}
Y.~Liu, J.~Peng, J.~Kang, A.~M. Iliyasu, D.~Niyato, and A.~A. Abd El-Latif, ``A secure federated learning framework for 5g networks,'' \emph{IEEE Wireless Communications}, vol.~27, no.~4, pp. 24--31, 2020.

\bibitem{fang2020local}
M.~Fang, X.~Cao, J.~Jia, and N.~Gong, ``Local model poisoning attacks to $\{$Byzantine-Robust$\}$ federated learning,'' in \emph{29th USENIX security symposium (USENIX Security 20)}, 2020, pp. 1605--1622.

\bibitem{tian2021adversarial}
J.~Tian, B.~Wang, R.~Guo, Z.~Wang, K.~Cao, and X.~Wang, ``Adversarial attacks and defenses for deep-learning-based unmanned aerial vehicles,'' \emph{IEEE Internet of Things Journal}, vol.~9, no.~22, pp. 22\,399--22\,409, 2021.

\bibitem{lin2017deep}
Y.~Lin, S.~Han, H.~Mao, Y.~Wang, and W.~J. Dally, ``Deep gradient compression: Reducing the communication bandwidth for distributed training,'' \emph{arXiv preprint arXiv:1712.01887}, 2017.

\bibitem{tian2024lesson}
J.~Tian, C.~Shen, B.~Wang, X.~Xia, M.~Zhang, C.~Lin, and Q.~Li, ``Lesson: Multi-label adversarial false data injection attack for deep learning locational detection,'' \emph{IEEE Transactions on Dependable and Secure Computing}, 2024.

\bibitem{tian2021joint}
J.~Tian, B.~Wang, Z.~Wang, K.~Cao, J.~Li, and M.~Ozay, ``Joint adversarial example and false data injection attacks for state estimation in power systems,'' \emph{IEEE Transactions on Cybernetics}, vol.~52, no.~12, pp. 13\,699--13\,713, 2021.

\bibitem{huang2019survey}
T.~Huang, W.~Yang, J.~Wu, J.~Ma, X.~Zhang, and D.~Zhang, ``A survey on green 6g network: Architecture and technologies,'' \emph{IEEE access}, vol.~7, pp. 175\,758--175\,768, 2019.

\bibitem{group2023quafu}
B.~Q. Group, ``Quafu-rl: The cloud quantum computers based quantum reinforcement learning,'' \emph{arXiv preprint arXiv:2305.17966}, 2023.

\bibitem{liang2023unleashing}
Z.~Liang, J.~Cheng, R.~Yang, H.~Ren, Z.~Song, D.~Wu, X.~Qian, T.~Li, and Y.~Shi, ``Unleashing the potential of llms for quantum computing: A study in quantum architecture design,'' \emph{arXiv preprint arXiv:2307.08191}, 2023.

\bibitem{ren2023towards}
C.~Ren, H.~Yu, R.~Yan, M.~Xu, Y.~Shen, H.~Zhu, D.~Niyato, Z.~Y. Dong, and L.~C. Kwek, ``Towards quantum federated learning,'' \emph{arXiv preprint arXiv:2306.09912}, 2023.

\bibitem{khan2017quantum}
A.~Khan, G.~Saha, and R.~K. Pal, ``Quantum computing based inference of grns,'' in \emph{Bioinformatics and Biomedical Engineering: 5th International Work-Conference, IWBBIO 2017, Granada, Spain, April 26--28, 2017, Proceedings, Part II 5}.\hskip 1em plus 0.5em minus 0.4em\relax Springer, 2017, pp. 221--233.

\bibitem{khan2021physical}
S.~A. Khan, F.~Hu, G.~Angelatos, and H.~E. T{\"u}reci, ``Physical reservoir computing using finitely-sampled quantum systems,'' \emph{arXiv preprint arXiv:2110.13849}, 2021.

\bibitem{bausch2020recurrent}
J.~Bausch, ``Recurrent quantum neural networks,'' \emph{Advances in neural information processing systems}, vol.~33, pp. 1368--1379, 2020.

\bibitem{farhi2014quantum}
E.~Farhi, J.~Goldstone, and S.~Gutmann, ``A quantum approximate optimization algorithm,'' \emph{arXiv preprint arXiv:1411.4028}, 2014.

\bibitem{peruzzo2014variational}
A.~Peruzzo, J.~McClean, P.~Shadbolt, M.-H. Yung, X.-Q. Zhou, P.~J. Love, A.~Aspuru-Guzik, and J.~L. O’brien, ``A variational eigenvalue solver on a photonic quantum processor,'' \emph{Nature communications}, vol.~5, no.~1, p. 4213, 2014.

\bibitem{kirby2021variational}
W.~M. Kirby and P.~J. Love, ``Variational quantum eigensolvers for sparse hamiltonians,'' \emph{Physical review letters}, vol. 127, no.~11, p. 110503, 2021.

\bibitem{kuete2023universal}
N.~Kuete~Meli, F.~Mannel, and J.~Lellmann, ``A universal quantum algorithm for weighted maximum cut and ising problems,'' \emph{Quantum Information Processing}, vol.~22, no.~7, p. 279, 2023.

\bibitem{kaewpuang2023adaptive}
R.~Kaewpuang, M.~Xu, D.~Niyato, H.~Yu, Z.~Xiong, and X.~S. Shen, ``Adaptive resource allocation in quantum key distribution (qkd) for federated learning,'' in \emph{2023 International Conference on Computing, Networking and Communications (ICNC)}.\hskip 1em plus 0.5em minus 0.4em\relax IEEE, 2023, pp. 71--76.

\bibitem{ren2024icc}
C.~Ren, H.~Xu, Minrui~Yu, Z.~Xiong, Z.~Zhang, and D.~Niyato, ``Variational quantum circuit and quantum key distribution-based quantum federated learning: A case of smart grid dynamic security assessment,'' \emph{International Conference on Communications}, 2024.

\bibitem{ren2023qfdsa}
C.~Ren, R.~Yan, M.~Xu, H.~Yu, Y.~Xu, D.~Niyato, and Z.~Y. Dong, ``Qfdsa: A quantum-secured federated learning system for smart grid dynamic security assessment,'' \emph{IEEE Internet of Things Journal}, 2023.

\bibitem{gurung2023performance}
D.~Gurung, S.~R. Pokhrel, and G.~Li, ``Performance analysis and evaluation of post quantum secure blockchain federated learning,'' \emph{arXiv preprint arXiv:2306.14772}, 2023.

\bibitem{buyukates2022lightverifl}
B.~Buyukates, J.~So, H.~Mahdavifar, and S.~Avestimehr, ``Lightverifl: Lightweight and verifiable secure federated learning,'' in \emph{Workshop on Federated Learning: Recent Advances and New Challenges (in Conjunction with NeurIPS 2022)}, 2022.

\bibitem{barz2012demonstration}
S.~Barz, E.~Kashefi, A.~Broadbent, J.~F. Fitzsimons, A.~Zeilinger, and P.~Walther, ``Demonstration of blind quantum computing,'' \emph{science}, vol. 335, no. 6066, pp. 303--308, 2012.

\bibitem{qu2021secure}
G.-J. Qu and M.-M. Wang, ``Secure multi-party quantum computation based on blind quantum computation,'' \emph{International Journal of Theoretical Physics}, vol.~60, no.~8, pp. 3003--3012, 2021.

\bibitem{horodecki2009quantum}
R.~Horodecki, P.~Horodecki, M.~Horodecki, and K.~Horodecki, ``Quantum entanglement,'' \emph{Reviews of modern physics}, vol.~81, no.~2, p. 865, 2009.

\bibitem{troupe2022quantum}
J.~Troupe, S.~Haldar, I.~Agullo, and P.~Kwiat, ``Quantum clock synchronization for future nasa deep space quantum links and fundamental science,'' \emph{arXiv preprint arXiv:2209.15122}, 2022.

\bibitem{friedman2000quantum}
J.~R. Friedman, V.~Patel, W.~Chen, S.~Tolpygo, and J.~E. Lukens, ``Quantum superposition of distinct macroscopic states,'' \emph{nature}, vol. 406, no. 6791, pp. 43--46, 2000.

\bibitem{romero2011large}
O.~Romero-Isart, A.~C. Pflanzer, F.~Blaser, R.~Kaltenbaek, N.~Kiesel, M.~Aspelmeyer, and J.~I. Cirac, ``Large quantum superpositions and interference of massive nanometer-sized objects,'' \emph{Physical review letters}, vol. 107, no.~2, p. 020405, 2011.

\bibitem{park2019quantum}
D.~K. Park, I.~Sinayskiy, M.~Fingerhuth, F.~Petruccione, and J.-K.~K. Rhee, ``Quantum forking for fast weighted power summation,'' \emph{arXiv preprint arXiv:1902.07959}, 2019.

\end{thebibliography}

\end{document}